\documentclass{article}
\usepackage{iclr2027_conference,times}
\iclrfinalcopy
\usepackage{amsmath,amssymb,booktabs,graphicx,microtype}
\usepackage{algorithm,algpseudocode}
\usepackage{array,tabularx,multirow,longtable,wrapfig}
\usepackage{xcolor}
\PassOptionsToPackage{hyphens}{url}
\usepackage{hyperref,url}
\usepackage{placeins,needspace}
\usepackage{fvextra}
\usepackage{tcolorbox}
\tcbuselibrary{breakable,skins}
\tcbset{rewardprompt/.style={
  colback=gray!5,
  colframe=black!60,
  fonttitle=\bfseries,
  title={#1},
  breakable,
  enhanced,
  before skip=6pt, after skip=6pt,
}}
\hypersetup{colorlinks=true,linkcolor=black,citecolor=black,urlcolor=black,pdftitle={Hedge-Bench: Evaluating Financial Research and Forecasting Agents},pdfauthor={Mahesh Ramesh, Eric Cho, Alice Lu, Shawn Huang, Baiqing Lyu},pdfsubject={Preprint}}
\graphicspath{{figures/}}
\makeatletter
\AddToHook{cmd/@makecol/before}{%
  \ifnum\value{page}=\getpagerefnumber{fig:pipeline}\relax
    \raggedbottom
  \fi
  \ifnum\value{page}=\getpagerefnumber{fig:frontier-updated}\relax
    \raggedbottom
  \fi
}
\makeatother
\newcommand{\bench}{\texttt{Hedge-Bench}}
\newcommand{\benchpublic}{\texttt{Hedge-Bench-\allowbreak Public}}
\newcommand{\benchprivate}{\texttt{Hedge-Bench-\allowbreak Private}}
\newcommand{\benchheldout}{\texttt{Hedge-Bench-\allowbreak Heldout}}
\newcommand{\benchtrain}{\texttt{Hedge-Bench-\allowbreak Train}}
\newcommand{\turncredit}{\texttt{TurnCredit}}
\newcommand{\base}{Muse-Glimmer}
\newcommand{\trained}{Muse-Glimmer-Hedge}
\newcommand{\ind}{\mathbf{1}}

\title{Hedge-Bench: Can Agents Reason Their Way to Better Predictions?}
\author{%
Mahesh Ramesh\thanks{Equal contribution.}\\
University of Wisconsin--Madison
\And
Eric Cho\textsuperscript{*}\thanks{Correspondence: \href{mailto:eric@trata.com}{eric@trata.com}.}\\
Trata
\And
Alice Lu\\
Trata
\And
Shawn Huang\\
Brigham Young University
\And
Baiqing Lyu\\
Osmosis AI
}
\begin{document}
\maketitle
\lhead{Preprint}

\begin{abstract}

Modern AI agents are increasingly capable of routine financial-analysis tasks such as retrieving documents and updating spreadsheets. A harder challenge is forecasting an unknown future from incomplete information. Public markets provide a natural testbed: analysts routinely predict discrete events with information from that point in time. We present \bench{}, a benchmark of 92 real-world environments derived from hedge fund analysts' reasoning traces paired with subsequently resolved prediction questions. Agents reason from frozen information snapshots, and their forecasts are evaluated against real-world outcomes. Across 11 frontier models and three trials per environment, joint success rates, which require both a correct forecast and over 60\% research coverage, range from 2.54 to 31.52\%. Using GRPO with \emph{turn-level credit assignment}, we train \base{} on 165 environments, increasing coverage on 22 held-out environments from 17.17\% to 33.36\% at low reasoning effort and from 30.19\% to 47.02\% at extra-high effort. The trained model at low effort also exceeds the base model's extra-high-effort coverage while using 47.9\% fewer tokens. We also observe evidence of generalization: held-out forecasting accuracy at high effort improves by 10.06 percentage points, while out-of-distribution evaluation scores improve by 3.94 points on GDPval, 2.29 on GDP.pdf, 7.07 on Harvey LAB, and 4.96 on Frontier Finance under their respective metrics. Code: \url{https://github.com/Trata-Inc/trata-hedge-bench-verified}.  
\end{abstract}

\section{Introduction}
Finance benchmarks evaluate agents' ability to do the mechanical work within research workflows \citep{bigeard2025financeagent,wang2026bigfinance,du2025deepresearch}, while forecasting benchmarks test predictions against subsequent events \citep{karger2025forecastbench,goel2026futuresim}. Yet a correct prediction may lack sound reasoning, and rigorous analysis may reach the wrong conclusion. 
\bench{} links both skills within the same environment to ask: does emulating the kinds of reasoning demonstrated by expert analysts improve prediction?

Making informed predictions about the future is core to the most consequential decisions institutions must make. ``How should DraftKings account for prediction markets when forecasting revenue and weighing capital allocation over the next 24 months?'' ``If SpaceX's Starlink will launch 30,000+ satellites by 2030 with best-in-class unit economics, should Iridium Communications, as a smaller incumbent, try to compete with a like-for-like replacement?'' ``If software is becoming easier to reproduce, which companies become more attractive to serial acquirers like Bending Spoons?'' These kinds of questions require reasoning from incomplete evidence about events that have yet to unfold.

In this paper, we introduce \bench{}, a testing suite that evaluates whether agents can learn financial reasoning and subsequently predict real-world outcomes. 
It pairs reasoning traces taken from real-world conversations between investment professionals with predictions regarding financial outcomes, which have since been resolved in the real world. These research tasks include reading between the lines of what was said and explicitly not said in documents, computing valuation multiples, assessing normalized earnings power, identifying possible inflections, benchmarking against peers, and synthesizing opposing data points into a coherent investment view. We also introduce the \benchpublic{} dataset, a set of 92 challenging environments across 23 companies requiring extensive domain knowledge, hours of directed research work and multi-step problem solving that represent what the highest paid finance professionals actually spend time doing.

\begin{figure}[t!]
\centering
\includegraphics[width=\linewidth]{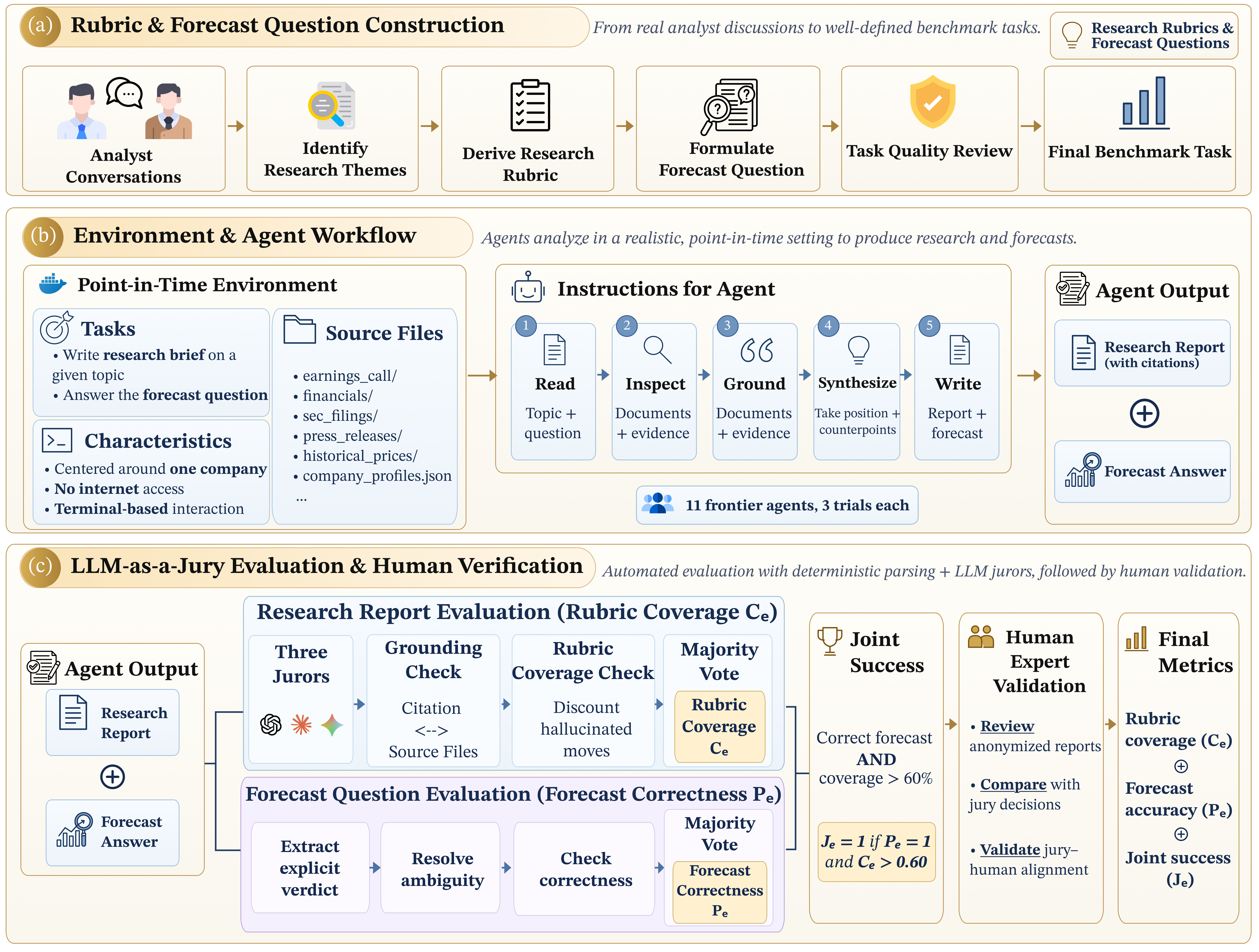} 

\caption{{\footnotesize Overview of Hedge-Bench. (a) Benchmark construction: Real analyst conversations are distilled into research rubrics and forecast questions, followed by environment quality review. (b) Agent workflow: Agents operate in company-specific, point-in-time environments with source documents and terminal-based interaction, without internet access, to produce a research report and a forecast answer. (c) Evaluation: An LLM jury assesses report grounding and rubric coverage through majority-vote aggregation, while a separate pipeline extracts and evaluates forecast answers. Joint success requires both a correct forecast and rubric coverage above 60\%. Human experts review anonymized reports to assess alignment between automated and human judgments.}}
\label{fig:pipeline}
\end{figure}
Each environment pairs an open-ended financial research task with a forecasting question. The agent receives a terminal workspace with access to primary financial documents up to a fixed cutoff date, and then writes a research analysis and forecast. The human analysts' work, which often contains competing investment arguments, defines a hidden rubric of comparisons and inferences that we call analytical moves. Each rubric item, or analytical move, is a specific calculation or inference that the report must make. We define a metric, ``research coverage,'' which measures the percentage of moves credited by the jury after citation checking. Actual events determine prediction correctness independently of the analysts' views.

We next use the benchmark data to test whether training improves financial research. Group Relative Policy Optimization (GRPO) trains a policy using the relative scores of responses to the same task \citep{shao2024deepseekmath}. A single trajectory score gives the same advantage to useful retrieval, repeated searches, and final synthesis. We instead use an LLM-as-judge to read the completed trajectory and assign positive, negative, or zero credit to individual turns. We call this procedure \textbf{\turncredit{}} and use it to train \base{} into \trained{}. We compare it first with outcome-only GRPO, which uses one score per trajectory, and then with a control that matches advantage magnitude. Evaluation on \benchheldout{} tests performance on environments excluded from training; \benchprivate{} tests transfer to new companies. We discuss how Hedge-Bench and \turncredit{} build on and extend prior work in Section~\ref{app:related-work}. 
Our contributions and main findings are:
\begin{enumerate}
\item \textbf{Benchmarking agents on both financial research and financial forecasting.} We construct 92 analyst-derived environments, each with a research and a forecasting task. (Section~\ref{sec:construction}) We evaluate 11 frontier models with three trials per environment (Section~\ref{sec:frontier}).
\item \textbf{Turn-level credit assignment improves research training.} We train \base{} with \turncredit{} and improve research coverage over outcome-only GRPO, with fewer repeated tool calls and no entropy collapse. Even at low effort, the \trained{} exceeds the \base{}’s extra-high-effort coverage while using 47.9$\%$ fewer tokens (Section~\ref{sec:training}).
\item \textbf{Training gains transfer beyond the training setting.} \base{} trained at high effort improves performance at all four reasoning efforts; on \benchprivate{} both research and forecasting improved, and training also improved performance on four out-of-distribution professional evaluations (Section~\ref{sec:scaling}).
\end{enumerate}
\section{Related Work}
\label{app:related-work}

Existing financial benchmarks mostly probe document reasoning and professional deliverables. FinQA \citep{chen2021finqa}, TAT-QA \citep{zhu2021tatqa}, DocFinQA \citep{reddy2024docfinqa}, and FinanceBench \citep{islam2023financebench} test reasoning over financial documents. FinBen includes forecasting and trading \citep{xie2024finben}. Closer to our setup, Vals AI's Finance Agent Benchmark evaluates agentic capabilities with rubrics derived from experts \citep{bigeard2025financeagent}.
 BigFinanceBench evaluates auditable financial derivations \citep{wang2026bigfinance}. \bench{} pairs analyst-derived research requirements with a separately resolved forecast.
The main difference from existing benchmarks is that our rubrics come from analysts who had already taken positions in the underlying stocks and done their own diligence on those companies. 
The rubrics capture counterarguments and investment judgments from discussions between two analysts who are familiar with the company.
\bench{} combines that research with predictions to evaluate the models' research and forecasting ability together. 

Apart from financial evaluation, \bench{} also gauges the agentic ability of the models. Following Terminal-Bench, a well-known benchmark of agentic ability \citep{merrill2026terminalbench}, \bench{} also uses a Harbor environment \citep{harbor2026framework}. APEX-Agents \citep{vidgen2026apexagents}, BankerToolBench \citep{lau2026bankertoolbench}, and BlueFin \citep{kundurthy2026bluefin} evaluate professional tasks across multiple domains, investment-banking deliverables, and financial spreadsheets, respectively. FrontierFinance evaluates open-ended investor research with expert rubrics \citep{samaya2026frontierfinance}. Our environments require both a research report and a forecast. ForecastBench evaluates predictions submitted before events resolve, while FutureSim uses chronological event replay \citep{karger2025forecastbench,goel2026futuresim}. In contrast, we measure research coverage and prediction correctness together, distinguishing correct forecasts with incomplete analysis from reports with high research coverage that predict incorrectly. 

There is no single consensus on how to assign per-turn credit in place of normal GRPO's trajectory-level advantage \citep{shao2024deepseekmath}. Recent work has shown that fine-grained credit assignment can improve agents in several setups. Faraday uses turn-level credit to weight GRPO updates for scientific replication \citep{falck2026faraday}, while CriticSearch uses a retrospective critic to score individual search turns \citep{zhang2026criticsearch}. DRACO redistributes credit without changing the trajectory advantage's sign or total token weight \citep{gandhi2026draco}. AgentEvolver instead combines signed action attribution with outcome rewards accumulated over later turns \citep{zhai2025agentevolver}.
 In \turncredit{}, the judge's assessment of each turn determines whether its actions are reinforced or down-weighted, while the trajectory score determines the strength of the update.
 We test this formulation on financial research against outcome-only GRPO and a magnitude-matched control (Section~\ref{sec:training}). Unlike DRACO, \turncredit{} can assign opposite signs to turns in the same trajectory. CriticSearch is also retrospective, but we go beyond short-horizon tasks and show that \turncredit{} improves long-horizon performance while using lower inference-time compute (Sections~\ref{sec:training} and~\ref{sec:scaling}). We also demonstrate transfer across reasoning efforts with test-time scaling and generalization to other professional tasks (Section~\ref{sec:scaling} and Appendix~\ref{app:external}).

\section{Hedge-Bench}
\label{sec:construction}
In this section, we explain how we build \bench{}. We describe how analyst discussions become research rubrics and forecasting questions. We outline the benchmark construction pipeline, environment formulation, and benchmark splits.

\subsection{From Analyst Conversations to Rubrics and Forecasts}

To source high-fidelity reasoning traces, we draw from a pool of over 500 investment professionals at established investment firms.  For each conversation, we connect two anonymous analysts from separate firms to discuss a public company they actively cover and have taken positions on. To make sure we capture genuine debate rather than scripted benchmark prompts, we record and transcribe these discussions to create the ground-truth rubrics. Figure~\ref{fig:pipeline} summarizes our benchmark construction pipeline.

\textbf{Rubric construction.}
A research theme is a broad question raised in the analyst discussion, such as how generative AI could affect a company's business. An analytical move is any specific counterargument or inference needed to address that question. We group these moves under their themes to form the hidden rubric. The task instructions give the agent the themes but keep the underlying moves hidden, and every move must be supportable by the materials supplied to the agent with citations.

For Getty-Shutterstock, the theme is the threat of generative AI to stock imagery. Its four moves require the agent to identify the exposure of creative stock images, consider whether secular decline should compress the valuation multiple, assess potential creative-revenue erosion even if merger synergies are realized, and examine licensing or IP protection as a defensive moat. Simply saying that AI is a risk does not complete these moves. The report must address these analytical moves, but semantically equivalent arguments receive credit, and agents are not penalized for reaching a different overall investment conclusion from the analysts.

\textbf{Forecast construction.}
We separately formulate a forecasting question from an unresolved issue raised by the research. Each question specifies a concrete event and resolution horizon, determined by a subsequent disclosure, event, or deadline. This separates research diligence from forecasting: an agent can execute thorough diligence while predicting the wrong outcome, or vice versa.

The public release is selected from a broader collection of more than 5,112 candidate research assignments gathered over nine months. Two independent human reviewers evaluate candidate difficulty, confirm that the supplied sources support all analytical requirements, and ensure the forecasting question has a clear, resolvable decision boundary.

\subsection{Task and Environment Formulation}
\label{sec}

\textbf{Environment.}
Each evaluation environment is a self-contained, point-in-time workspace centered on one public company and research topic. It specifies the research themes, forecasting question, information cutoff, and required output. Agents do not receive the underlying analyst conversation, hidden move-level rubric, or future resolution evidence. Each environment runs in a Harbor container \citep{harbor2026framework} with Internet access disabled.

\textbf{Tasks.}
Each environment contains two paired tasks. First, the agent writes an analytical research report addressing the specified themes and supporting its claims with evidence (citations). Second, it answers a concrete, falsifiable binary forecasting question. Both deliverables are submitted together in \texttt{/app/answer.md}. Dataset sizes and trial counts refer to environments, not to the two tasks within each environment. 

\Needspace{20\baselineskip}
\begin{wrapfigure}{r}{0.48\textwidth}
\centering
\includegraphics[width=\linewidth]{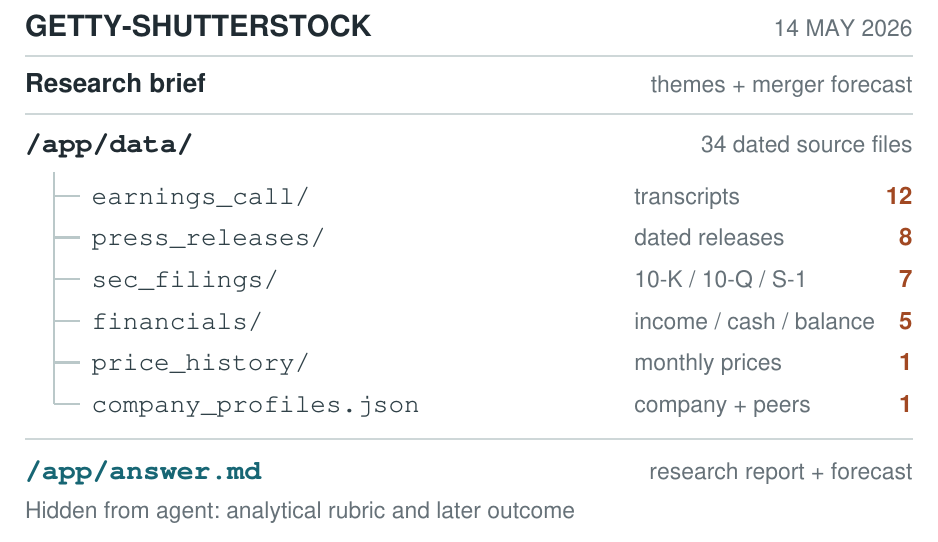}
\caption{{\footnotesize One Getty-Shutterstock environment at the May 14, 2026 cutoff. The environment provides 34 point-in-time source files.}}
\label{fig:environment-layout}
\end{wrapfigure}

\textbf{Resources.}
Agents receive only the frozen point-in-time materials provided under \texttt{/app/data/}. Depending on the environment, these include company and peer filings, earnings-call transcripts, press releases, financial statements, historical prices, and company profiles. Agents may inspect these materials through the terminal and perform arbitrary calculations, but cannot retrieve information beyond the environment's cutoff date. Figure~\ref{fig:environment-layout} illustrates the structure of one such environment.

The Getty-Shutterstock environment, for example, contains 34 source files and asks whether the proposed merger will close within nine months. Its 15 analytical moves cover business complementarity, regulatory execution, AI exposure, and the balance sheet. An agent might find the financial rationale for the merger compelling while forecasting that regulatory barriers will prevent it from closing. Appendix~\ref{app:examples} provides the full environment description and resolution chronology.
\subsection{Data splits}

\begin{table}[!ht]
\centering
\caption{Hedge-Bench environment sets.}
\label{tab:data}
\footnotesize
\setlength{\tabcolsep}{3pt}
\begin{tabularx}{\linewidth}{@{}>{\raggedright\arraybackslash}p{0.46\linewidth}>{\raggedright\arraybackslash}X@{}}
\toprule
Set (environments) & Role \\
\midrule
\benchpublic{} (92) & Public benchmark and frontier evaluation \\
\benchtrain{} (165) & Policy training \\
\benchheldout{} (22) & Held-out policy evaluation \\
\benchprivate{} (53) & New-company transfer \\
\bottomrule
\end{tabularx}
\end{table}

\bench{} comprises four splits (Table~\ref{tab:data}). \benchpublic{} contains 92 environments spanning 23 companies, with 43 Yes and 49 No forecasting outcomes. In total, these environments contain 368 research themes and 1,150 analytical moves. Each environment includes 3--5 themes and 9--20 moves (median 12), supported by 20--43 source files (median 31). Information cutoffs range from April 25, 2025 to May 22, 2026.

For training, \benchtrain{} contains 165 environments: 70 environments drawn from the public set and 95 additional private training-only environments. The remaining 22 public environments form \benchheldout{}, which was held out from all training related activities such as policy updates, reward-prompt development, etc. Finally, \benchprivate{} tests generalization to unseen companies across 53 environments from 11 firms absent from RL training.

\section{Experiment Setup}
\label{sec:experiment-setup}

We evaluate research reports and forecasting answers separately, then combine them into a joint success metric. Below, we describe the scoring procedure, the LLM-as-a-Jury protocol and its validation against human financial experts, and the evaluation harness.

\subsection{Evaluation and Scoring}
\label{sec:evaluation}

\textbf{LLM-as-a-Jury.}
We use three frontier models as independent jurors: Gemini 3.8 Flash, Claude Opus 5, and GPT-5.6 Sol. Each juror evaluates the agent output using only the ground-truth rubric and the agent's answer, with no access to external information or other jurors' decisions. Juror decisions are aggregated by majority vote \citep{verga2024poll}.

\textbf{Evaluation settings.}
We evaluate agents under two settings. In the \textit{full-environment} setting, agents receive the complete environment context in \texttt{/app/data/} and are required to produce both a research report and a forecast. In the \textit{question-only} setting, agents receive only the forecasting question, without access to \texttt{/app/data/}, and are asked to provide the forecast alone. This ablation measures how much forecasting performance depends on the supplied documents, while also probing how much relevant company knowledge is already encoded in the models' pretrained parameters.

\textbf{Research report scoring.}
Each agent must support its analysis in the report with citations to sources in \texttt{/app/data/}. We score reports by the fraction of required rubric moves that are both analytically present and citation-grounded.
Each juror performs two stages of evaluation. First, it verifies citation grounding by checking each cited claim against excerpts from the referenced source. Claims that are not supported by the cited evidence are flagged as hallucinated. If a majority of jurors flag the evidence supporting a rubric move as hallucinated, a citation penalty removes credit for that move. 

Second, each juror evaluates rubric coverage by comparing the full report against every required analytical move in the ground-truth rubric. Semantically equivalent arguments receive credit even when phrased differently, whereas generic mentions of the broader theme do not. A move is covered when a majority of jurors credit the analysis and do not flag its supporting evidence as hallucinated.

Formally, in an environment $e$ with $K_e$ rubric moves, for each move $k$, let $h_{ejk}$ indicate whether juror $j$ credits the move and $t_{ejk}$ whether it flags the supporting evidence as hallucinated. Let $\ind[\cdot]$ equal 1 when the condition holds and 0 otherwise. With majority threshold $m=2$,
\begin{equation}
v_{ek}=
\ind\!\left[\sum_j h_{ejk}\geq m\right] \cdot
\ind\!\left[\sum_j t_{ejk}<m\right],\qquad
C_e=\frac{1}{K_e}\sum_{k=1}^{K_e}v_{ek},\qquad m=2.
\label{eq:coverage}
\end{equation}
Thus, $C_e$ is the fraction of required moves receiving valid majority credit. Appendix~\ref{app:jury-qualitative} provides qualitative examples of accepted and rejected moves.

\textbf{Human expert validation.}
To check whether the jury’s scores align with expert judgment, two financial professionals independently reviewed eight anonymized reports without access to model identities and jury scores. Across $\approx$ 24 hours of review, the two experts made 197 individual judgments covering 110 distinct report-rubric-move pairs. Both experts gave decisions on 81 of these pairs. On this shared subset, the jury agreed with each expert on 77.8\% of decisions, compared with 70.4\% agreement between the experts. Under a stricter criterion requiring both the analytical move and factual support, jury-expert agreement was 75.5\% and 77.0\%. Appendix~\ref{app:human-assessment} provides additional details.

\textbf{Forecasting question scoring.}
For environment $e$, prediction correctness is
\begin{equation}
P_e=\ind[\hat y_e=y_e],
\end{equation}
where $\hat y_e$ is the agent's forecast and $y_e$ is the resolved outcome. We extract the explicit verdict and use the jury to resolve conditional or ambiguous answers against the event definition. The question-only ablation uses the same forecasting question and label but provides no documents or Internet tools. 

\textbf{Joint success scoring.}
Joint success requires both a correct forecast and report coverage strictly above 60\%:
\begin{equation}
J_e(\tau)=P_e\,\ind[C_e>\tau],\qquad \tau=0.60.
\label{eq:joint}
\end{equation}
The threshold is applied before rounding. For example, 12- and 15-move rubrics require at least 8 and 10 credited moves, respectively. We also report continuous coverage and forecasting accuracy, with threshold sensitivity in Appendix~\ref{app:joint}. For coverage and joint success, we average repeated trials within each environment and then average across environments. Forecasting accuracy is the fraction of valid responses that are correct. All three frontier-agent trials receive equal weight. Appendix~\ref{app:protocol} provides the detailed evaluation settings and individual-judge results.

When testing several comparisons, we apply Holm's correction to the $p$-values \citep{holm1979multiple}. This controls the chance of a false positive across the set of comparisons. Each analysis states which comparisons are corrected together.

\subsection{Agents and Harness}
\label{sec:models-harness}

We evaluate 11 frontier proprietary and open-source models on \benchpublic{}, with three trials per model and environment. Each environment runs in a Harbor container \citep{harbor2026framework} using the Terminus-2 terminal-based workflow \citep{merrill2026terminalbench}. Models use provider-specific reasoning controls. Full execution settings are given in Appendix~\ref{app:execution-settings}.

\section{Evaluation Results}
\label{sec:frontier}
\begin{figure}[!t]
\centering
\includegraphics[width=\linewidth]{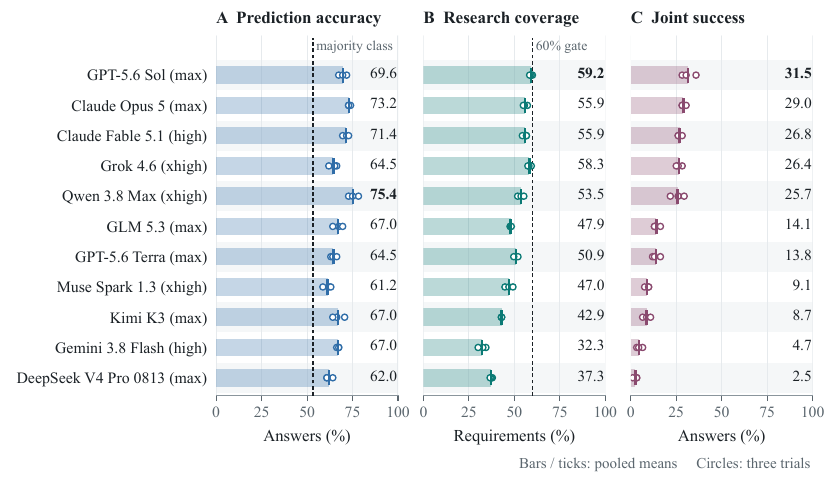}
\caption{{\footnotesize Each model has 92 environments and three trials (276 answers). Bars and vertical ticks show three-trial means; open circles show individual-trial means. Joint success requires a correct forecast and strictly more than 60\% coverage. Dashed lines mark the 53.26\% majority-class baseline and the 60\% coverage threshold.}}
\label{fig:frontier-updated}
\end{figure}

We evaluate 11 frontier models with three trials on each of the 92 \benchpublic{} environments, alongside standalone question-only predictions. We use the settings and scoring procedure explained in Section~\ref{sec:evaluation}. As shown in Figure~\ref{fig:frontier-updated}, joint success ranges from 2.54\% to 31.52\%, with GPT-5.6 Sol achieving the highest observed rate and six models scoring below 20\%. Open-weight model families average 12.77\% joint success, compared with 20.19\% for proprietary families, despite similar forecasting accuracy, reflecting stronger research coverage among closed-source models (Appendix~\ref{app:frontier-access}). Across all 92 environments, Kimi K3 uses the fewest turns (28.1 per trajectory), while Qwen3.8 Max averages the most at 111.6 turns (Table~\ref{tab:frontier-trajectories}). Within the same model and environment, longer trajectories do not consistently correspond to higher coverage or better forecasts (Table~\ref{tab:frontier-trajectory-correlations}). Longer final reports also need not achieve higher coverage (Appendix~\ref{app:frontier-length}). Probing trajectories also show that repeated file retrieval and extended self-checking can still end in low coverage. In 1,622 of 3,036 reports, the jury applies citation penalties to analytical moves with inadequate supporting evidence, confirming that citation verification prevents unsupported analysis from receiving credit (Appendix~\ref{app:frontier-evidence}).  Appendix~\ref{app:frontier-failures} gives model-by-model examples. 

\textbf{Research coverage and forecast accuracy can differ.}
While comprehensive rubric coverage should ideally support a correct forecast, partial coverage can steer reasoning toward inaccurate predictions. For example, Grok 4.6 has higher coverage than Qwen3.8 Max (58.25\% versus 53.54\%) but lower forecast accuracy (64.49\% versus 75.36\%). To separate this comparison from differences between models and environments, we also compare repeated trials within each model and environment. After centering within these pairs, model-specific research coverage-prediction accuracy correlations range from $-0.123$ to $0.119$, with none significant after Holm correction (Appendix~\ref{app:frontier-within-task}). Appendix~\ref{app:frontier-cases} qualitatively shows both outcomes on the same environments: a correct forecast with incomplete research, and a more detailed report with an incorrect forecast. 

\textbf{Full environments improve forecasting.}
This near-zero correlation raises whether models rely on environment documents or simply recall pretraining knowledge about the companies. To test this, we evaluate the models on all 92 forecast questions in a standalone question-only condition across three trials (Table~\ref{tab:qo-all}). Ten of eleven models have higher observed forecast accuracy with the full environment than with the question alone. Qwen3.8 Max rises from 59.06\% to 75.36\%, and Gemini 3.8 Flash from 56.16\% to 67.03\%. Muse Spark 1.3 is the exception, scoring 61.96\% question-only and 61.23\% with the environment. Performance is also not confined to older outcomes: ten of eleven models achieve higher forecasting accuracy on questions resolved after March 2026 than on those resolved earlier (Table~\ref{tab:calendar-resolution}), indicating that models are not simply recalling outcomes memorized during pretraining. 

\section{Post-Training}
\label{sec:training}

\begin{figure}[!t]
\centering\includegraphics[width=\linewidth]{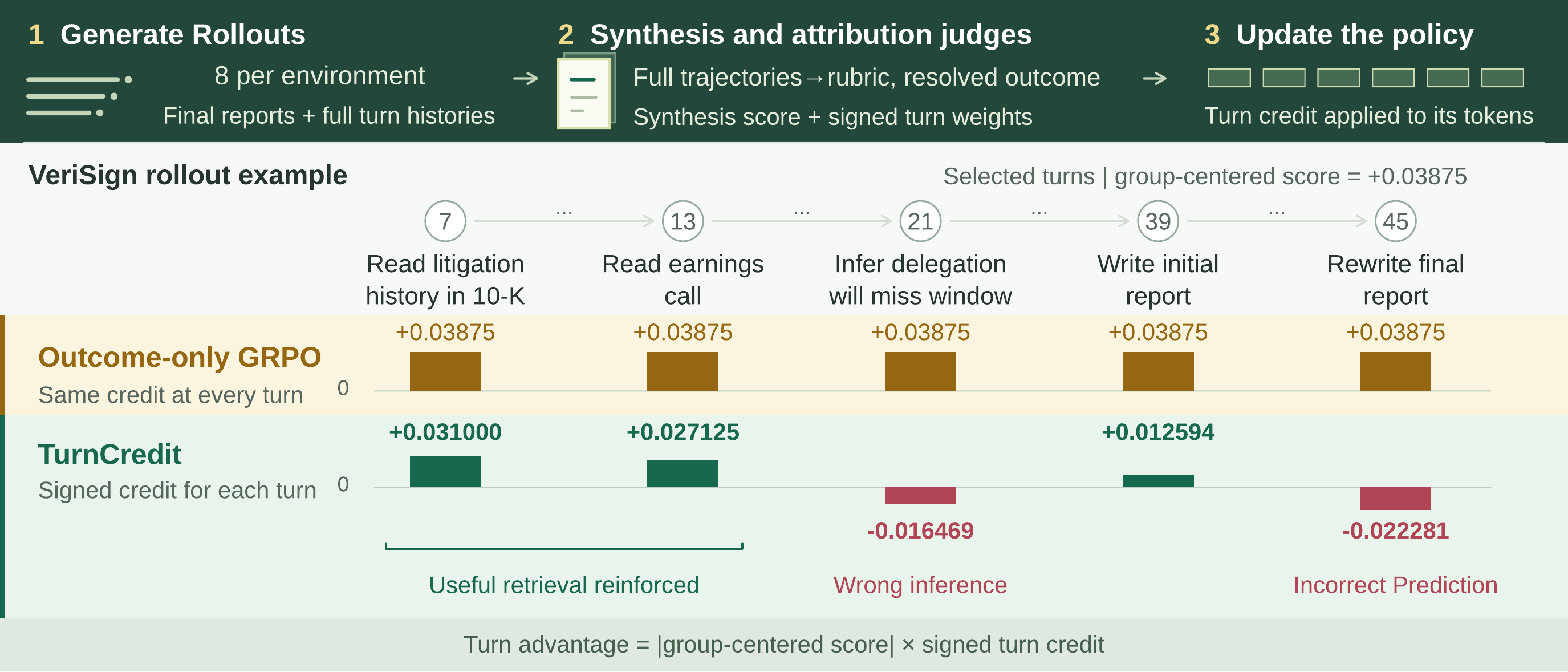}
\caption{\turncredit{} uses the completed rollout to distinguish useful retrieval from faulty inference and synthesis. The bars show token coefficients for selected turns of a saved VeriSign trajectory, compared with uniform outcome-only credit on that same trajectory.}
\label{fig:turn-credit-flow}
\end{figure}

Next, we use \bench{} to test whether rewarding expert-defined reasoning improves financial research. We post-train \base{}-30B \citep{meta2026museglimmer} as a terminal-based agent on \benchtrain{}'s 165 environments, using LoRA adapters \citep{hu2022lora} and high reasoning effort for one epoch (21 steps). Each of the 21 updates uses eight environments with eight rollouts per environment, for a batch size of 64 trajectories. The resulting policy is \trained{}.
Instead of combining forecast correctness and coverage into a fixed weighted reward, we score how well the report uses evidence to support its forecast.

 \subsection{\turncredit{}: Training with Turn-Level Credit}

\begin{figure}[!t]
\centering\includegraphics[width=\linewidth]{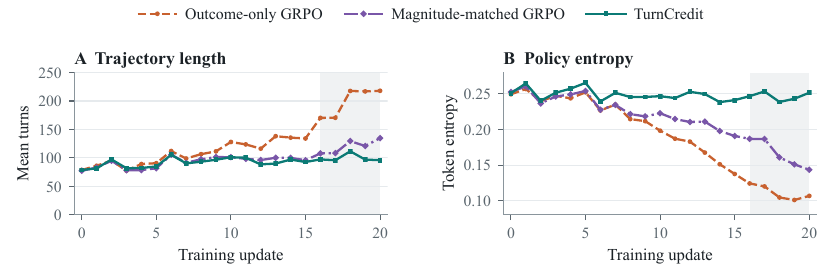}
\caption{{\footnotesize \turncredit{} reduces redundant turns and avoids policy entropy collapse compared with outcome-only and magnitude-matched GRPO.}}
\label{fig:training-dynamics-main}
\end{figure}

In preliminary runs with outcome-only GRPO, policy entropy collapsed, and the number of turns also increased as training progressed (Figure~\ref{fig:training-dynamics-main}), motivating the use of turn-level credit to reinforce useful turns and penalize redundant ones, following the idea of process supervision \citep{lightman2024verify}. Turn-level credit can be assigned in several ways, such as sampling additional continuations from each turn to estimate advantage. However, this strategy substantially increases rollout cost. \turncredit{} instead assigns credit after each rollout is complete, using an LLM judge \citep{falck2026faraday}.

For each environment, we sample eight trajectories, indexed by $i$. A Gemini 3.7 Flash judge scores how well each report uses evidence to support its forecast. We call the LLM judge twice for each report and average the two scores to obtain the trajectory's synthesis score $s_i$. The attribution judge reads the completed report with its synthesis score and written assessment, alongside the full agent trajectory and tool outputs. It also receives the task instruction and hidden rubric, together with the grading verdict identifying accepted and missed moves and forecast correctness. It judges early retrieval by its later use: useful evidence gathering receives positive credit, an unsuccessful but relevant search can remain neutral, and redundant retrieval or faulty synthesis receives negative credit. Figure~\ref{fig:turn-credit-flow} shows the credit assigned to selected turns of a VeriSign rollout. Appendix~\ref{app:saved-credit} gives the turn-by-turn details of the VeriSign rollout.

We separate advantage scale from turn-level direction. First, we center the scores of the eight sibling rollouts of a prompt without dividing by their standard deviation \citep{liu2025understanding}, $\Delta_i=s_i-\frac{1}{8}\sum_{i'=1}^{8}s_{i'}$.
For turn $j$ in trajectory $i$, the judge returns reasoning success $r^+_{ij}\in[0,1]$, reasoning failure $r^-_{ij}\in[0,1]$, and signed tool utility $u_{ij}\in[-1,1]$. We combine these into positive and negative weights $w^+_{ij}$ and $w^-_{ij}$. For eligible token $t$, let $j(t)$ be its turn and $A_{it}$ its advantage:
\begin{equation}
 w^+_{ij}=\max(r^+_{ij},\max(u_{ij},0)),\,\,
 w^-_{ij}=\max(r^-_{ij},\max(-u_{ij},0)),\,\,
 A_{it}=|\Delta_i|(w^+_{ij(t)}-w^-_{ij(t)}).
 \label{eq:credit}
\end{equation}
Here, $|\Delta_i|$ sets the advantage scale, while turn weights determine which tokens are reinforced or suppressed. Each turn's coefficient is copied to its eligible assistant tokens, excluding prompt tokens and tool observations. We use a token-level clipped GRPO objective, following DAPO \citep{yu2025dapo,schulman2017ppo,shao2024deepseekmath}, implemented with slime \citep{slime_github}. Full training details and judge prompts appear in Appendices~\ref{app:reward} and~\ref{app:full-reward-prompts}.



\subsection{Results: Turn-level training improves financial research}
\begin{wraptable}{R}{0.49\textwidth}
\centering
\caption{{\footnotesize Research coverage on \benchheldout{} (\%). Each condition contains eight trials on each of 22 environments. Turn-level training exceeds magnitude-matched GRPO at all three efforts (Holm $p<0.001$). Unscaled high-effort evaluation was not completed because of repeated turn/context exhaustion.}}
\label{tab:training-main}
\footnotesize
\setlength{\tabcolsep}{2pt}
\begin{tabularx}{\linewidth}{@{}>{\raggedright\arraybackslash}Xrrr@{}}
\toprule
Policy & Low & Medium & High \\
\midrule
\base{} & 17.17 & 28.64 & 29.51 \\
Outcome-only GRPO & 17.41 & 24.33 & -- \\
Magnitude-matched GRPO & 21.69 & 31.20 & 33.02 \\
Turn-level GRPO (\trained{}) & \textbf{33.36} & \textbf{46.02} & \textbf{46.82} \\
\bottomrule
\end{tabularx}
\end{wraptable}

We evaluate \base{} and the trained model \trained{} on the 22 \benchheldout{} environments, with eight trials per environment at low, medium, high, and extra-high reasoning effort. Each condition contains 176 reports, for 1,408 overall, assessed by the three-juror procedure in Section~\ref{sec:evaluation}. Coverage rises from 17.17\% to 33.36\% at low effort and from 30.19\% to 47.02\% at extra-high effort, with gains of 16.19-17.38 points across settings. Joint success also improves at every matched effort, by 7.39-18.18 points (Holm $p\leq0.00089$). Thus, more reports both exceed 60\% coverage and predict correctly. For example, at extra-high effort, \trained{} examines Flywire's past healthcare underperformance and rollout risks, covering 10/12 moves versus 3/12 for \base{}. See Appendix~\ref{app:paired-flywire} for a qualitative comparison. \trained{} at low effort also exceeds \base{} at extra-high effort in coverage while using 47.9\% fewer tokens (Figure~\ref{fig:scaling}).  

We then compare \turncredit{} with outcome-only GRPO, which uses the same synthesis score but applies each trajectory's advantage uniformly to its eligible tokens. \turncredit{} improves coverage by 15.96 points at low effort and 21.69 at medium effort. Outcome-only GRPO gives little improvement over the base model at low effort and performs worse at medium effort (Table~\ref{tab:training-main}). The high-effort outcome-only GRPO evaluation could not be completed because trajectories repeatedly exhausted the turn budget, even after we increased it from 240 to 960 turns (Appendix~\ref{app:heldout-protocol}).

As shown in Equation~\ref{eq:credit}, \turncredit{} also changes the credit magnitude, since the turn weights can reduce the absolute advantage applied to each token. To test whether this improvement can simply be attributed to advantage magnitude, we train a magnitude-matched outcome-only GRPO baseline that computes turn weights and uses them solely to globally rescale the trajectory-level advantages to the same mean absolute token magnitude. \trained{} exceeds the magnitude-matched control by 11.67, 14.82, and 13.80 coverage points at low, medium, and high effort, respectively (all three differences remain significant after Holm correction, $p<0.001$). Matching advantage magnitude does not reproduce the gain.

\label{sec:mechanism}
Turn-level training also produces shorter, less repetitive training trajectories. Over later training updates (updates 16-20; Figure~\ref{fig:training-dynamics-main}), it averages 99.6 turns and 1.23\% repeated commands, compared with 198.7 turns and 12.59\% for outcome-only GRPO and 120.3 turns and 3.95\% for the magnitude-matched control. Entropy remains stable compared to the outcome-only and magnitude-matched variants. Full held-out results and statistical tests are given in Appendix~\ref{app:heldout-protocol}, and Appendix~\ref{app:credit} provides further training diagnostics and the magnitude-matched control definition.

\vspace{-6pt}
\section{Generalization}
\label{sec:scaling}
\vspace{-6pt}

\begin{figure}[!t]
\centering
\includegraphics[width=\linewidth]{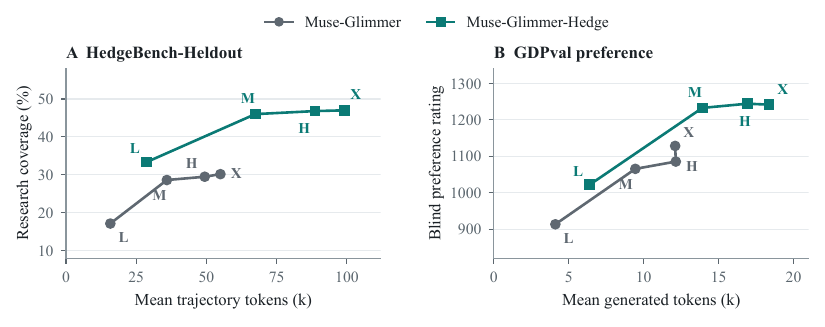}
\caption{{\footnotesize High-effort training improves performance across efforts. L, M, H, and X denote low through extra-high effort. Panels show \benchheldout{} coverage (A) and GDPval pool-relative blind preference ratings (B).}}
\label{fig:scaling}
\vspace{-8pt}
\end{figure}

We also evaluate whether the gains from training transfer across reasoning efforts and to other benchmarks. Although trained only at high effort, \trained{} improves coverage across all four tested efforts (Figure~\ref{fig:scaling}). At low effort, it achieves 33.36\% coverage with 28,662 trajectory tokens (47.9\% shorter trajectories), versus 30.19\% and 54,974 tokens for extra-high \base{}. Low-effort \trained{} exceeds all four \base{} efforts in coverage on the fixed 22 environments (\benchheldout{}). All four comparisons remain significant after Holm correction ($p\leq0.02496$; Table~\ref{tab:heldout-significance}).  

\vspace{-3pt}
To test forecasting generalization without company overlap from training, we evaluate on \benchprivate{}: 53 environments across 11 companies absent from RL training, with three trials per environment and condition.
Coverage improves by 13.26 points at low effort and 16.44 points at high effort. High-effort prediction accuracy increases from 38.36\% to 48.43\%, a 10.06-point gain. Both coverage gains and high-effort forecasting remain significant after Holm correction (Table~\ref{tab:ood-effort}). Low-effort \trained{} also exceeds high-effort \base{} coverage with 40.2\% fewer trajectory tokens (Table~\ref{tab:ood-length}). The 53 environments contain 28 Yes and 25 No outcomes. Private environments present greater difficulty across models; that is why forecasting accuracy is less than majority-label prediction (28/53). For reference, GPT-5.6 Sol at maximum effort achieves 45.33\% coverage and 35.85\% prediction accuracy, versus 59.17\% and 69.57\% on \benchpublic{} (Table~\ref{tab:ood-reference}). Appendix~\ref{app:private-protocol} provides the reference comparison and answer rates, and Tables~\ref{tab:ood-resolution-dates} and~\ref{tab:public-resolution-dates} list the resolution dates.

\vspace{-4pt}
\begin{wraptable}{R}{0.49\textwidth}
\centering
\caption{{\footnotesize Professional-task transfer under each benchmark's metric (Appendix~\ref{app:external}). All seven gains remain significant after Holm correction. \texttt{MG} and \texttt{MG-H} are Muse-Glimmer and Muse-Glimmer-Hedge.}}
\label{tab:transfer}
\footnotesize
\setlength{\tabcolsep}{2pt}
\begin{tabularx}{\linewidth}{@{}>{\raggedright\arraybackslash}Xrrr@{}}
\toprule
Evaluation & \shortstack[r]{\texttt{MG}} & \shortstack[r]{\texttt{MG-H}} & Gain \\
\midrule
GDPval, low & 74.19 & \textbf{77.93} & +3.74 \\
GDPval, medium & 79.82 & \textbf{81.71} & +1.90 \\
GDPval, high & 78.91 & \textbf{81.52} & +2.61 \\
GDPval, extra-high & 78.92 & \textbf{82.86} & +3.94 \\
GDP.pdf, high & 69.85 & \textbf{72.14} & +2.29 \\
Harvey LAB, high & 64.19 & \textbf{71.26} & +7.07 \\
Frontier Finance, high & 39.62 & \textbf{44.58} & +4.96 \\
\bottomrule
\end{tabularx}
\end{wraptable}

Finally, we evaluate whether the learned research ability transfers to other professional benchmarks: GDPval \citep{patwardhan2025gdpval}, GDP.pdf \citep{surge2026gdppdf}, Harvey LAB \citep{harvey2026lab}, and Frontier Finance \citep{samaya2026frontierfinance}, under their respective metrics. Appendix~\ref{app:protocol-gdpval} describes our Bash-based GDPval setup and preference pool, and Appendix~\ref{app:protocol-harvey} describes the Harvey execution and scoring adaptations. The GDP.pdf input rendering is specified in Appendix~\ref{app:protocol-gdppdf}, and Frontier Finance tools and judge in Appendix~\ref{app:protocol-frontier}. Table~\ref{tab:transfer} reports rubric-based results, and Appendix~\ref{app:external-statistics} gives paired statistical tests.

On our Bash-based adaptation of the 220-task GDPval set, \trained{} improves rubric completion at all four efforts. For example, in the same task, \trained{} preserves the requested store-count definition, with 63.8\% rubric completion versus 37.9\%. It also retrieves a missing handbook instead of assuming its requirements, achieving 100\% versus 50\% with fewer turns (check Appendix~\ref{app:paired-examples} for qualitative examples). Blind preference ratings also increase, with paired Elo gains of 108.7-167.1 across all efforts (Figure~\ref{fig:scaling}). For Harvey LAB, we select four tasks from each of 25 areas of legal practice using a deterministic diversity procedure, giving 100 tasks (Appendix~\ref{app:protocol-harvey}). Its two judges separately measure gains of 7.13 and 7.02 points. GDP.pdf uses all 100 benchmark tasks with five attempts each (Appendix~\ref{app:protocol-gdppdf}), and Frontier Finance uses 189 matched queries (Appendix~\ref{app:protocol-frontier}). The gains remain significant under a joint Holm correction ($p\leq0.01274$; Appendix~\ref{app:external-statistics}).

\vspace{-10pt}
\section{Future Directions}
\label{app:future-direction}
\vspace{-4pt}

Most environments in this paper are based on discussions between two analysts about securities they have analyzed and hold positions in. In the future, we can include discussions among more than two people on broader topics, such as whether AI will displace SaaS businesses, to build richer environments.
We can use the raw audio files to study how models understand 45- to 60-minute research discussions. Comparing the original recording with its transcript on the same task would test whether listening to the discussion helps more than reading the transcript. We can build RL environments around these recordings to improve audio understanding. Computer-use agents could perform Hedge-Bench research and forecasting tasks through financial websites or spreadsheets instead of a terminal.

\vspace{-3pt}
We can extend each environment over time, releasing new information and asking the agent to revise its forecast, similar to FutureSim \citep{goel2026futuresim}. This tests whether models change their position when the evidence changes. On the training side, we can extend \turncredit{} to assign credit to individual reasoning steps within a turn, with an LLM judge providing finer-grained training feedback.
The judge could also directly compare multiple trajectories from the same environment, similar to MiMo-V2.6, which uses groupwise grading to compare rollouts \citep{mimo2026v26pro}. Comparing trajectories provides a relative reference, which may make useful differences easier to identify than when each trajectory is judged individually.

\vspace{-10pt}
\section{Conclusion}
\vspace{-4pt}
\bench{} evaluates research completion and forecasting on 92 public analyst-derived environments. Correct predictions often accompany incomplete analysis, motivating separate measures and their conjunction. Our evaluation of 11 frontier models reveals a significant gap relative to expert-derived research standards: joint success ranges from only 2.54\% to 31.52\% (Figure~\ref{fig:frontier-updated}). \turncredit{} improves coverage over trajectory-level GRPO and a magnitude-matched variant that proper credit assignment is crucial in these environments. More importantly, the gains from post-training \base{} on \bench{} hold across reasoning efforts and extend to a wide range of other knowledge works (Table~\ref{tab:transfer}), demonstrating the value of \bench{} as both a training and evaluation resource.

\par

\vspace{-10pt}
\section*{Acknowledgements}
\vspace{-4pt}
We would like to thank Yiding Jiang, Will Gao, Pavan Agrawal, and Kasey Zhang for their helpful suggestions.

\clearpage
\bibliography{references}

@article{holm1979multiple,
  author = {Holm, Sture},
  title = {A Simple Sequentially Rejective Multiple Test Procedure},
  journal = {Scandinavian Journal of Statistics},
  volume = {6},
  number = {2},
  pages = {65--70},
  year = {1979},
  url = {https://www.jstor.org/stable/4615733}
}

@inproceedings{chen2021finqa,
  title={{FinQA}: A Dataset of Numerical Reasoning over Financial Data},
  author={Chen, Zhiyu and Chen, Wenhu and Smiley, Charese and Shah, Sameena and Borova, Iana and Langdon, Dylan and Moussa, Reema and Beane, Matt and Huang, Ting-Hao and Routledge, Bryan and Wang, William Yang},
  booktitle={Proceedings of the 2021 Conference on Empirical Methods in Natural Language Processing},
  year={2021}, pages={3697--3711}, publisher={Association for Computational Linguistics},
  doi={10.18653/v1/2021.emnlp-main.300}, url={https://aclanthology.org/2021.emnlp-main.300/}}

@article{islam2023financebench,
  title={{FinanceBench}: A New Benchmark for Financial Question Answering},
  author={Islam, Pranab and Kannappan, Anand and Kiela, Douwe and Qian, Rebecca and Scherrer, Nino and Vidgen, Bertie},
  journal={arXiv preprint arXiv:2311.11944}, year={2023}, url={https://arxiv.org/abs/2311.11944}}

@inproceedings{karger2025forecastbench,
  title={{ForecastBench}: A Dynamic Benchmark of {AI} Forecasting Capabilities},
  author={Karger, Ezra and Bastani, Houtan and Chen, Yueh-Han and Jacobs, Zachary and Halawi, Danny and Zhang, Fred and Tetlock, Philip E.},
  booktitle={International Conference on Learning Representations}, year={2025}, url={https://iclr.cc/virtual/2025/poster/28507}}

@article{patwardhan2025gdpval,
  title={{GDPval}: Evaluating {AI} Model Performance on Real-World Economically Valuable Tasks},
  author={Patwardhan, Tejal and Dias, Rachel and Proehl, Elizabeth and Kim, Grace and Wang, Michele and Watkins, Olivia and Posada Fishman, Sim{\'o}n and Aljubeh, Marwan and Thacker, Phoebe and Fauconnet, Laurance and Kim, Natalie S. and Chao, Patrick and Miserendino, Samuel and Chabot, Gildas and Li, David and Sharman, Michael and Barr, Alexandra and Glaese, Amelia and Tworek, Jerry},
  journal={arXiv preprint arXiv:2510.04374}, year={2025}, url={https://arxiv.org/abs/2510.04374}}

@article{shao2024deepseekmath,
  title={{DeepSeekMath}: Pushing the Limits of Mathematical Reasoning in Open Language Models},
  author={Shao, Zhihong and Wang, Peiyi and Zhu, Qihao and Xu, Runxin and Song, Junxiao and Bi, Xiao and Zhang, Haowei and Zhang, Mingchuan and Li, Y. K. and Wu, Y. and Guo, Daya},
  journal={arXiv preprint arXiv:2402.03300}, year={2024}, url={https://arxiv.org/abs/2402.03300}}

@article{schulman2017ppo,
  title={Proximal Policy Optimization Algorithms},
  author={Schulman, John and Wolski, Filip and Dhariwal, Prafulla and Radford, Alec and Klimov, Oleg},
  journal={arXiv preprint arXiv:1707.06347}, year={2017}, url={https://arxiv.org/abs/1707.06347}}

@inproceedings{lightman2024verify,
  title={Let's Verify Step by Step},
  author={Lightman, Hunter and Kosaraju, Vineet and Burda, Yuri and Edwards, Harrison and Baker, Bowen and Lee, Teddy and Leike, Jan and Schulman, John and Sutskever, Ilya and Cobbe, Karl},
  booktitle={International Conference on Learning Representations}, year={2024}, url={https://openreview.net/forum?id=v8L0pN6EOi}}

@misc{harvey2026lab,
  author={{Harvey AI}}, title={{Harvey LAB}: The Legal Agent Benchmark}, year={2026},
  howpublished={Public benchmark repository},
  url={https://github.com/harveyai/harvey-labs}}

@misc{aa2026gdpval,
  author={{Artificial Analysis}}, title={{GDPval-AA v2} Leaderboard},
  howpublished={Online evaluation resource}, year={n.d.},
  url={https://artificialanalysis.ai/evaluations/gdpval-aa},
  note={Accessed 2026-09-09; living leaderboard and methodology resource}}

@misc{harbor2026framework,
  author={{Harbor Framework Team}},
  title={{Harbor}: A Framework for Evaluating and Optimizing Agents and Models in Container Environments},
  year={2026}, howpublished={Software}, doi={10.5281/zenodo.20953922},
  url={https://doi.org/10.5281/zenodo.20953922}}

@article{bradley1952paired,
  title={Rank Analysis of Incomplete Block Designs: I. The Method of Paired Comparisons},
  author={Bradley, Ralph Allan and Terry, Milton E.},
  journal={Biometrika}, volume={39}, number={3--4}, pages={324--345}, year={1952},
  doi={10.1093/biomet/39.3-4.324}, url={https://doi.org/10.1093/biomet/39.3-4.324}}

@inproceedings{zhu2021tatqa,
  title={{TAT-QA}: A Question Answering Benchmark on a Hybrid of Tabular and Textual Content in Finance},
  author={Zhu, Fengbin and Lei, Wenqiang and Huang, Youcheng and Wang, Chao and Zhang, Shuo and Lv, Jiancheng and Feng, Fuli and Chua, Tat-Seng},
  booktitle={Proceedings of the 59th Annual Meeting of the Association for Computational Linguistics and the 11th International Joint Conference on Natural Language Processing (Volume 1: Long Papers)},
  year={2021}, pages={3277--3287}, publisher={Association for Computational Linguistics},
  doi={10.18653/v1/2021.acl-long.254}, url={https://aclanthology.org/2021.acl-long.254/}}

@inproceedings{reddy2024docfinqa,
  title={{DocFinQA}: A Long-Context Financial Reasoning Dataset},
  author={Reddy, Varshini and Koncel-Kedziorski, Rik and Lai, Viet Dac and Krumdick, Michael and Lovering, Charles and Tanner, Chris},
  booktitle={Proceedings of the 62nd Annual Meeting of the Association for Computational Linguistics (Volume 2: Short Papers)},
  year={2024}, pages={445--458}, publisher={Association for Computational Linguistics},
  doi={10.18653/v1/2024.acl-short.42}, url={https://aclanthology.org/2024.acl-short.42/}}

@article{xie2024finben,
  title={{FinBen}: A Holistic Financial Benchmark for Large Language Models},
  author={Xie, Qianqian and Han, Weiguang and Chen, Zhengyu and Xiang, Ruoyu and Zhang, Xiao and He, Yueru and Xiao, Mengxi and Li, Dong and Dai, Yongfu and Feng, Duanyu and Xu, Yijing and Kang, Haoqiang and Kuang, Ziyan and Yuan, Chenhan and Yang, Kailai and Luo, Zheheng and Zhang, Tianlin and Liu, Zhiwei and Xiong, Guojun and Deng, Zhiyang and Jiang, Yuechen and Yao, Zhiyuan and Li, Haohang and Yu, Yangyang and Hu, Gang and Huang, Jiajia and Liu, Xiao-Yang and Lopez-Lira, Alejandro and Wang, Benyou and Lai, Yanzhao and Wang, Hao and Peng, Min and Ananiadou, Sophia and Huang, Jimin},
  journal={arXiv preprint arXiv:2402.12659}, year={2024}, url={https://arxiv.org/abs/2402.12659}}

@article{bigeard2025financeagent,
  title={Finance Agent Benchmark: Benchmarking {LLMs} on Real-world Financial Research Tasks},
  author={Bigeard, Antoine and Nashold, Langston and Krishnan, Rayan and Wu, Shirley},
  journal={arXiv preprint arXiv:2508.00828}, year={2025}, url={https://arxiv.org/abs/2508.00828}}

@article{wang2026bigfinance,
  title={{BigFinanceBench}: A Workflow-Grounded Benchmark for Financial-Research Agents},
  author={Wang, Alex and Meinhardt, Georg and Katz, Jacob and Kim, Joseph H. and Chaudhary, Pratyush K. and Blagden, Chase and Xu, Eric},
  journal={arXiv preprint arXiv:2606.03829}, year={2026}, url={https://arxiv.org/abs/2606.03829}}

@article{du2025deepresearch,
  title={{DeepResearch Bench}: A Comprehensive Benchmark for Deep Research Agents},
  author={Du, Mingxuan and Xu, Benfeng and Zhu, Chiwei and Wang, Xiaorui and Mao, Zhendong},
  journal={arXiv preprint arXiv:2506.11763}, year={2025}, url={https://arxiv.org/abs/2506.11763}}

@article{goel2026futuresim,
  title={{FutureSim}: Replaying World Events to Evaluate Adaptive Agents},
  author={Goel, Shashwat and Chandak, Nikhil and Arun, Arvindh and Prabhu, Ameya and Staab, Steffen and Hardt, Moritz and Andriushchenko, Maksym and Geiping, Jonas},
  journal={arXiv preprint arXiv:2605.15188}, year={2026}, url={https://arxiv.org/abs/2605.15188}}

@inproceedings{hu2022lora,
  title={{LoRA}: Low-Rank Adaptation of Large Language Models},
  author={Hu, Edward J. and Shen, Yelong and Wallis, Phillip and Allen-Zhu, Zeyuan and Li, Yuanzhi and Wang, Shean and Wang, Lu and Chen, Weizhu},
  booktitle={International Conference on Learning Representations}, year={2022}, url={https://arxiv.org/abs/2106.09685}}

@misc{openai2026solcutoff,
 title={{GPT-5.6 Sol Model}}, author={{OpenAI}}, year={2026},
 note={Model documentation, accessed September 9, 2026}, url={https://developers.openai.com/api/docs/models/gpt-5.6-sol}}

@misc{openai2026terracutoff,
 title={{GPT-5.6 Terra Model}}, author={{OpenAI}}, year={2026},
 note={Model documentation, accessed September 9, 2026}, url={https://developers.openai.com/api/docs/models/gpt-5.6-terra}}

@misc{google2026geminicutoff,
 title={{Gemini 3.8 Flash Model Card}}, author={{Google DeepMind}}, year={2026},
 note={Accessed September 9, 2026}, url={https://deepmind.google/models/model-cards/gemini-3-8-flash/}}

@misc{anthropic2026fablecutoff,
 title={{Claude Fable 5.1}}, author={{Anthropic}}, year={2026},
 note={Model documentation, accessed September 9, 2026}, url={https://platform.claude.com/docs/en/models/fable-5-1/overview}}

@article{verga2024poll,
  title={Replacing Judges with Juries: Evaluating {LLM} Generations with a Panel of Diverse Models},
  author={Verga, Pat and Hofstatter, Sebastian and Althammer, Sophia and Su, Yixuan and Piktus, Aleksandra and Arkhangorodsky, Arkady and Xu, Minjie and White, Naomi and Lewis, Patrick},
  journal={arXiv preprint arXiv:2404.18796}, year={2024}, url={https://arxiv.org/abs/2404.18796}}

@misc{xai2026grokcutoff,
 author={{xAI}}, title={{Grok 4.6: Model Documentation}}, year={2026},
 url={https://docs.x.ai/developers/grok-4-6}, note={Accessed September 9, 2026}
}

@misc{anthropic2026modeloverview,
 author={{Anthropic}}, title={{Models Overview}}, year={2026},
 url={https://platform.claude.com/docs/en/models/overview}, note={Accessed September 9, 2026}
}

@misc{qwen2026modeloverview,
 author={{Qwen Team}}, title={{Qwen Studio: Models}}, year={2026},
 url={https://chat.qwen.ai/legal-agreement/models}, note={Accessed September 9, 2026}
}

@misc{zai2026glm53,
 author={{Z.AI}}, title={{GLM-5.3 Overview}}, year={2026},
 url={https://docs.z.ai/guides/llm/glm-5.3}, note={Accessed September 9, 2026}
}

@misc{moonshot2026kimik3,
 author={{Moonshot AI}}, title={{Kimi-K3 Model Card}}, year={2026},
 url={https://huggingface.co/moonshotai/Kimi-K3}, note={Accessed September 9, 2026}
}

@misc{deepseek2026v4card,
 author={{DeepSeek AI}}, title={{DeepSeek-V4-Pro Model Card}}, year={2026},
 url={https://huggingface.co/deepseek-ai/DeepSeek-V4-Pro}, note={Accessed September 9, 2026}
}

@misc{meta2026spark13,
 author={{Meta AI}}, title={{Introducing Muse Spark 1.3}}, year={2026},
 url={https://research.meta.ai/blog/introducing-muse-spark-1-3}, note={Accessed September 9, 2026}
}

@misc{getty2026termination,
  author = {{Getty Images Holdings, Inc.}},
  title = {Form 8-K: Termination of the Shutterstock Merger Agreement},
  year = {2026},
  month = jul,
  howpublished = {U.S. Securities and Exchange Commission},
  url = {https://www.sec.gov/Archives/edgar/data/1898496/000121390026075721/ea029721801-8k_getty.htm},
  note = {Filed July 7, 2026}
}

@misc{surge2026gdppdf,
  author = {{Surge AI}},
  title = {{GDP.pdf}},
  year = {2026},
  url = {https://huggingface.co/datasets/surgeai/GDP.pdf}
}

@misc{samaya2026frontierfinance,
  author = {{Samaya AI}},
  title = {Introducing {FrontierFinance}: A Challenging Benchmark for Measuring Frontier Intelligence of Finance Agents},
  year = {2026},
  url = {https://samaya.ai/blog/frontier-finance}
}

@inproceedings{ramesh2026playingwithfire,
  title = {Playing with Fire: What Transfers When {RL} Trains a Language Agent?},
  author = {Ramesh, Mahesh and Jayakumar, Kaousheik and Ram, Hemanth and Thodima, Pavan and Duraiswami, Ramani and Manocha, Dinesh and Rege, Aniket and Vlatakis-Gkaragkounis, Emmanouil-Vasileios},
  booktitle = {ICML 2026 Workshop on RLxF: Reinforcement Learning from World Feedback},
  year = {2026},
  url = {https://openreview.net/forum?id=5VrEt3fthr}
}

@misc{meta2026museglimmer,
 author={{Meta Superintelligence Lab}}, title={{Muse Glimmer Model Card}}, year={2026},
 url={https://huggingface.co/meta-models/Muse-Glimmer-30B}, note={Model release date: August 2026; accessed September 18, 2026}
}

@article{vidgen2026apexagents,
  title={{APEX-Agents}},
  author={Vidgen, Bertie and Mann, Austin and Fennelly, Abby and Stanly, John Wright and Rothman, Lucas and Burstein, Marco and Benchek, Julien and Ostrofsky, David and Ravichandran, Anirudh and Sur, Debnil and Venugopal, Neel and Hsia, Alannah and Robinson, Isaac and Huang, Calix and Varones, Olivia and Khan, Daniyal and Haines, Michael and Bridges, Austin and Boyle, Jesse and Twist, Koby and Richards, Zach and Mahapatra, Chirag and Foody, Brendan and Nitski, Osvald},
  journal={arXiv preprint arXiv:2601.14242},
  year={2026},
  url={https://arxiv.org/abs/2601.14242}
}

@article{lau2026bankertoolbench,
  title={{BankerToolBench}: Evaluating {AI} Agents in End-to-End Investment Banking Workflows},
  author={Lau, Elaine and D{\"u}cker, Markus and Chaudhary, Ronak and Goh, Hui Wen and Wei, Rosemary and Kumar, Vaibhav and Qunbar, Saed and Gogia, Guram and Liu, Yi and Millslagle, Scott and Borazjanizadeh, Nasim and Tkachenko, Ulyana and Danquah, Samuel Eshun and Schweiker, Collin and Karumathil, Vijay and Devalaraju, Asrith and Sandadi, Varsha and Nam, Haemi and Arani, Punit and Epps, Ray and Arif, Abdullah and Bhaiwala, Sahil and Northcutt, Curtis and Wang, Skyler and Athalye, Anish and Mueller, Jonas and Guzm{\'a}n, Francisco},
  journal={arXiv preprint arXiv:2604.11304},
  year={2026},
  url={https://arxiv.org/abs/2604.11304}
}

@article{kundurthy2026bluefin,
  title={{BlueFin}: Benchmarking {LLM} Agents on Financial Spreadsheets},
  author={Kundurthy, Srivatsa and Na, Clara and Moraine, Colton and Mohta, Anoushka and Winter, Case and Fang, George and Ling, John and Strubell, Emma and Kirshner, Zach},
  journal={arXiv preprint arXiv:2605.30907},
  year={2026},
  url={https://arxiv.org/abs/2605.30907}
}

@article{falck2026faraday,
  title={Training {AI} Scientists to Replicate Research},
  author={Falck, Damon and Sabri, Samer and Surina, Anja and Foster, Thom and Sims, Anya and Devlin, Sam and Rogers, Dylan and Collins, Tantum and Aleksiev, Kaloyan and Kirsch, Louis and Hughes, Edward},
  journal={arXiv preprint arXiv:2608.13331},
  year={2026},
  url={https://arxiv.org/abs/2608.13331}
}

@article{gandhi2026draco,
  title={{DRACO}: Fine-Grained Credit Assignment with Dynamic Rubrics for Long-Horizon Agent Training},
  author={Gandhi, Shubham and Goyal, Saurabh and Kate, Kiran and Rizk, Yara},
  journal={arXiv preprint arXiv:2609.04094},
  year={2026},
  url={https://arxiv.org/abs/2609.04094}
}

@inproceedings{zhang2026criticsearch,
  title={{CriticSearch}: Fine-Grained Credit Assignment for Search Agents via a Retrospective Critic},
  author={Zhang, Yaocheng and Huang, Haohuan and Song, Zijun and Zhao, Zijie and Zhang, Qichao and Zhu, Yuanheng and Zhao, Dongbin},
  booktitle={Findings of the Association for Computational Linguistics: ACL 2026},
  year={2026},
  pages={12272--12290},
  publisher={Association for Computational Linguistics},
  doi={10.18653/v1/2026.findings-acl.596},
  url={https://aclanthology.org/2026.findings-acl.596/}
}

@article{merrill2026terminalbench,
  title = {{Terminal-Bench}: Benchmarking Agents on Hard, Realistic Tasks in Command Line Interfaces},
  author = {Merrill, Mike A. and Shaw, Alexander G. and Carlini, Nicholas and others},
  journal = {arXiv preprint arXiv:2601.11868},
  year = {2026},
  url = {https://arxiv.org/abs/2601.11868}
}

@article{zhai2025agentevolver,
  title = {{AgentEvolver}: Towards Efficient Self-Evolving Agent System},
  author = {Zhai, Yunpeng and Tao, Shuchang and Chen, Cheng and Zou, Anni and Chen, Ziqian and Fu, Qingxu and Mai, Shinji and Yu, Li and Deng, Jiaji and Cao, Zouying and Liu, Zhaoyang and Ding, Bolin and Zhou, Jingren},
  journal = {arXiv preprint arXiv:2511.10395},
  year = {2025},
  url = {https://arxiv.org/abs/2511.10395}
}

@inproceedings{li2024rewardcomposition,
  title={Optimizing Language Models with Fair and Stable Reward Composition in Reinforcement Learning},
  author={Li, Jiahui and Zhang, Hanlin and Zhang, Fengda and Chang, Tai-Wei and Kuang, Kun and Chen, Long and Zhou, Jun},
  booktitle={Proceedings of the 2024 Conference on Empirical Methods in Natural Language Processing},
  year={2024}, pages={10122--10140},
  publisher={Association for Computational Linguistics},
  doi={10.18653/v1/2024.emnlp-main.565},
  url={https://aclanthology.org/2024.emnlp-main.565/}
}

@article{liu2025understanding,
  title={Understanding {R1-Zero-Like} Training: A Critical Perspective},
  author={Liu, Zichen and Chen, Changyu and Li, Wenjun and Qi, Penghui and Pang, Tianyu and Du, Chao and Lee, Wee Sun and Lin, Min},
  journal={arXiv preprint arXiv:2503.20783},
  year={2025},
  url={https://arxiv.org/abs/2503.20783}
}

@article{yu2025dapo,
  title={{DAPO}: An Open-Source {LLM} Reinforcement Learning System at Scale},
  author={Yu, Qiying and Zhang, Zheng and Zhu, Ruofei and Yuan, Yufeng and Zuo, Xiaochen and Yue, Yu and Dai, Weinan and Fan, Tiantian and Liu, Gaohong and Liu, Lingjun and Liu, Xin and Lin, Haibin and Lin, Zhiqi and Ma, Bole and Sheng, Guangming and Tong, Yuxuan and Zhang, Chi and Zhang, Mofan and Zhang, Wang and Zhu, Hang and Zhu, Jinhua and Chen, Jiaze and Chen, Jiangjie and Wang, Chengyi and Yu, Hongli and Song, Yuxuan and Wei, Xiangpeng and Zhou, Hao and Liu, Jingjing and Ma, Wei-Ying and Zhang, Ya-Qin and Yan, Lin and Qiao, Mu and Wu, Yonghui and Wang, Mingxuan},
  journal={arXiv preprint arXiv:2503.14476},
  year={2025},
  url={https://arxiv.org/abs/2503.14476}
}

@misc{slime_github,
  author = {Zilin Zhu and Chengxing Xie and Xin Lv and {slime Contributors}},
  title = {slime: An {LLM} post-training framework for {RL} Scaling},
  year = {2025},
  howpublished = {\url{https://github.com/THUDM/slime}},
  note = {GitHub repository. Corresponding author: Xin Lv}
}

@misc{mimo2026v26pro,
  title={{MiMo-V2.6-Pro-RL}},
  author={{Xiaomi MiMo Team}},
  year={2026},
  howpublished={\url{https://huggingface.co/XiaomiMiMo/MiMo-V2.6-Pro-RL}}
}

@article{agrawal2025gepa,
  title={{GEPA}: Reflective Prompt Evolution Can Outperform Reinforcement Learning},
  author={Agrawal, Lakshya A and Tan, Shangyin and Soylu, Dilara and Ziems, Noah and Khare, Rishi and Opsahl-Ong, Krista and Singhvi, Arnav and Shandilya, Herumb and Ryan, Michael J and Jiang, Meng and Potts, Christopher and Sen, Koushik and Dimakis, Alexandros G. and Stoica, Ion and Klein, Dan and Zaharia, Matei and Khattab, Omar},
  journal={arXiv preprint arXiv:2507.19457},
  year={2025},
  url={https://arxiv.org/abs/2507.19457}
}

@misc{openai_codex,
  author={{OpenAI}},
  title={Code generation},
  year={n.d.},
  url={https://developers.openai.com/api/docs/guides/code-generation},
  note={OpenAI documentation. Accessed September 30, 2026}
}
\bibliographystyle{iclr2027_conference}
\clearpage
\appendix\raggedbottom
\renewcommand{\topfraction}{0.9}
\renewcommand{\bottomfraction}{0.8}
\renewcommand{\textfraction}{0.1}
\renewcommand{\floatpagefraction}{0.8}
\makeatletter
\setlength{\@fptop}{0pt}
\setlength{\@fpsep}{12pt}
\makeatother

\section*{Appendix Contents}
\label{app:contents}
\begingroup
\normalsize\setlength{\parskip}{0pt}
\setcounter{tocdepth}{2}
\makeatletter
\let\appendixcontentsline\contentsline
\renewcommand{\contentsline}[4]{%
  \ifnum#3>\getpagerefnumber{app:contents}\relax
    \appendixcontentsline{#1}{#2}{#3}{#4}%
  \fi
}
\renewcommand{\l@section}{\addvspace{4pt}\@dottedtocline{1}{0em}{2.3em}}
\renewcommand{\l@subsection}{\@dottedtocline{2}{2.3em}{3.2em}}
\@starttoc{toc}
\makeatother
\endgroup
\clearpage

\section{Qualitative Environment Examples}
\label{app:source-materials}
Over the span of nine months, we collected more than 5,112 candidate research assignments. From these, we handpicked environments for the public release by considering difficulty and the range of financial topics covered. The 92 public environments cover valuation, growth, mergers and acquisitions, competitive positioning, operational execution, and risk. Each environment demands well-defined research covering the analytical moves and correct forecasting. 
 
The 92 environments contain 368 research themes and 1,150 analytical moves. Of the environments, 16 have three themes, 60 have four, and 16 have five. Of the 368 themes, 322 contain three moves and 46 contain four. The median environment has 12 moves and 31 source files. Table~\ref{tab:source-composition} summarizes the source collection. The same source document can appear in multiple environments and is counted for each environment that includes it.

\begin{table}[h]
\centering\small
\caption{Source materials supplied in the 92 \benchpublic{} environments.}
\label{tab:source-composition}
\begin{tabular}{lr}
\toprule
Source category & Environment-local file instances \\
\midrule
Earnings-call transcripts & 1,038 \\
Financial statements & 816 \\
SEC filings & 494 \\
Press releases & 424 \\
Company profiles & 92 \\
Price histories & 91 \\
Investor presentations & 12 \\
\midrule
Total & 2,967 \\
\bottomrule
\end{tabular}
\end{table}

We provide examples from two companies in this paper: Getty Images (NYSE: GETY) and TeraWulf (NASDAQ: WULF). We score reports on whether they complete the required analysis using evidence available at the cutoff date, and whether each forecast matches the later resolved outcome.

Each analytical requirement states what the report must explain. For WULF, the report must compare colocation income with GPU ownership by examining how predictable the returns are and how long the assets remain useful. Simply noting demand for computing capacity is not enough. For Getty’s proposed merger with Shutterstock, the report must examine how the businesses fit together, regulatory risks, competition from AI, and the combined company’s finances. In both examples, the required analysis is defined by the human analysts' reasoning, whereas the ground-truth prediction is set by the real-world outcome.

\subsection{Getty-Shutterstock Merger}
\label{app:examples}
Getty-Shutterstock merger environment asks:
\begin{quote}
Consider the proposed merger between Getty Images and Shutterstock, the rationale for it, and the approval process it has been navigating. Will the merger close within the next nine months?
\end{quote}
The accompanying research assignment covers four themes: complementary profiles, regulatory hurdles, AI disruption risk, and pro forma viability and synergies. The rubric specifies 15 moves across these themes (4, 3, 4, and 4).

\textbf{The instruction prompt.}\label{app:getty-instruction}
The following is the full agent-facing \texttt{instruction.md} from the current public-release version, reproduced verbatim. It uses the citation tool described in Appendix~\ref{app:citation-processing}; the reported evaluations used the earlier file-path citation protocol. The analytical rubric below is hidden from the agent.

\begin{tcolorbox}[rewardprompt={Getty-Shutterstock: full instruction prompt}]
\begin{Verbatim}
You are a financial analyst. Your task is to evaluate the provided topic for **Getty Images Holdings, Inc. (GETY)** as of **2026-05-14** based on the data provided in the environment, and ultimately make a concrete, falsifiable prediction.

## The Task

- **Primary Company:** Getty Images Holdings, Inc. (GETY)
- **Peer Companies:** Shutterstock, Inc. (SSTK)
- **Analysis Cutoff Date:** 2026-05-14

Develop a view on the topic below across the following themes and write your full analysis as a markdown file to `/app/answer.md`:

- **Topic:** Strategic Rationale of the Getty-Shutterstock Merger

1. Complementary Asset and Balance Sheet Profiles
2. Regulatory and Execution Hurdles
3. AI Disruption and Secular Decline Risk
4. Pro Forma Financial Viability and Synergies

## Prediction Verdict

Consider the proposed GETY-Shutterstock merger, the rationale for it, and the approval process it has been navigating. Will the merger close within the next nine months?

End your `/app/answer.md` with a section titled `## Prediction Verdict` that answers this question and commits to one outcome. Show your reasoning.

## Requirements

Your analysis must:

- Take a clear position and engage with the strongest counter-evidence in the data.
- Draw a concrete tradeoff, judgment, or conclusion out of your analysis. If management is vague or key data is unresolved, state that uncertainty directly instead of guessing.
- Use the provided themes to build a logical chain of reasoning that culminates in a final prediction.
- Organize your analysis by the themes above and end with your prediction verdict.
- Ground every specific factual claim with an inline citation immediately after the claim (see **Citation Format** below). A citation grounds only the sentence it appears in, so place each citation in the same sentence as the facts it supports. Analytical inferences need no citation, but every fact they rest on does.
- Preserve formatting. When writing `/app/answer.md`, use a quoted heredoc (`cat << 'EOF' > /app/answer.md` ... `EOF`) or another method that does not pass your text through shell expansion. Inside double quotes, bash silently deletes `$`-prefixed text (e.g. `$8.56B` becomes `.56B`). If a large heredoc stalls, append the analysis in a few smaller quoted heredocs (`cat << 'EOF' >> /app/answer.md`) or write a Python script to a file and run it. Never shorten the analysis to get the write through. After writing, re-read the file and confirm every figure survived, because the grader scores only what is literally in the file.

Claims you can't ground in `/app/data/` (numbers, events, entities) are discarded by the grader and won't count toward your score.

## Data Layout

`/app/data/` may contain any of the following (point-in-time as of 2026-05-14):

- `earnings_call/` — `.txt` earnings call transcripts for the primary company
- `financials/income_statement/` — quarterly `.json` income statements (revenue, margins, net income) for the primary company and peers
- `financials/balance_sheet/` — quarterly `.json` balance sheet (assets, liabilities, equity) for the primary company
- `financials/cash_flow/` — quarterly `.json` cash flow statements for the primary company and peers
- `press_releases/` — full-text wire press releases for the primary company as `.txt` files
- `sec_filings/10-k/` — annual SEC reports as `.txt` for the primary company (Risk Factors, Business Description, MD&A)
- `sec_filings/10-q/` — quarterly SEC reports as `.txt` for the primary company (interim MD&A, Risk Factors deltas)
- `sec_filings/s-1/` — IPO prospectus as `.txt` for the primary company (only present for recent IPOs)
- `company_profiles.json` — company profiles and market data for the primary company and peers
- `price_history/` — monthly `.json` closing price, market cap, and volume for the primary company

The titles of the files within each folder provide additional context. Peer files are provided for competitor comparisons or industry benchmarking where relevant. If a competitor is private or lacks standalone filings, evaluate competitive dynamics using GETY's disclosures. Not every environment will contain every source listed above. Work with the information that is present.

## Citation Format

Use `python3 /app/citation_tool.py` to select evidence from `/app/data`. The command checks the file and returns a citation ID such as `[[ev_0123456789abcdef01234567]]`. Put that exact ID immediately after the factual claim it supports, in the same sentence. Use multiple IDs when a comparison depends on multiple records. The grader accepts citation IDs issued during this task; handwritten file-path citations do not count.

- **Text:** `python3 /app/citation_tool.py text earnings_call/<file>.txt --lines 10-12 --speaker "Speaker Name"`. `--speaker` must match the turn header covering the lines; copy that name (`Sarah M. London:`, not a job title or `Speaker`). If the tool says `--speaker 'X' does not speak … spoken by Y`, retry the same lines with `Y`. Multi-speaker windows are OK if you name one person who actually speaks in that range. Use the narrowest supporting line range. For other text files, omit `--speaker`.
- **JSON:** `python3 /app/citation_tool.py json financials/income_statement/<file>.json --where date=2025-09-30 --keys revenue,grossProfit`. Use enough `--where` fields to select exactly one record. For a company profile, `--where symbol=CARR --keys marketCap` selects one company. For monthly prices, `--where date=2025-09-30 --keys close,marketCap` selects one row.

Check the returned excerpt against the full claim. A valid ID locates evidence; it does not make an unsupported statement true. Cite facts and numbers, then state your analytical inference separately.
\end{Verbatim}
\end{tcolorbox}

\textbf{The analytical rubric.}\label{app:getty-rubric}
The 15 criteria below are hidden from the agent. Reports receive credit for the required analysis regardless of wording, provided it is supported by evidence in \texttt{/app/data/}. Analysis based on unsupported or fabricated claims receives no credit.

\noindent\textbf{1. Complementary Asset and Balance Sheet Profiles.}
The merger pairs Getty Images' leading editorial business and extensive archive with Shutterstock's cash flow generation. Management presents the combination as a path to further delever Getty's balance sheet and expand margins and cash flow.
\begin{description}
\item[(a)] Identifies Getty's high debt load or interest burden as a primary risk to its standalone equity value.
\item[(b)] Contrasts Shutterstock's operating cash flow and period-end cash with Getty's higher interest expense and net debt.
\item[(c)] Characterizes Getty's editorial and archival segments as more resilient or premium compared to the more challenged creative or stock imagery segments.
\item[(d)] Infers that management positions the merger as a means to further delever Getty through combined cash flow and margins.
\end{description}

\noindent\textbf{2. Regulatory and Execution Hurdles.}
The transaction faces significant scrutiny from international regulators.
\begin{description}
\item[(a)] Identifies specific regulatory bodies involved in the approval process, such as the DOJ or the U.K. Competition and Markets Authority (CMA).
\item[(b)] Analyzes the likelihood of the deal closing based on the nature of proposed remedies or the lack of an outright block from regulators.
\item[(c)] Notes the extended timeline of the merger process, describing it as delayed or prolonged.
\end{description}

\noindent\textbf{3. AI Disruption and Secular Decline Risk.}
Generative AI represents a secular threat that could commoditize the stock imagery business, potentially offsetting any gains from merger-related cost cuts.
\begin{description}
\item[(a)] Identifies AI-generated content as a direct threat to the marginal or disposable creative stock imagery market.
\item[(b)] Discusses the risk of multiple compression where the market refuses to pay a premium for earnings perceived to be in secular decline.
\item[(c)] Evaluates the potential for revenue erosion in the creative segment even if the company achieves its stated synergy targets.
\item[(d)] Mentions the importance of IP indemnification or licensing deals with AI platforms as a potential defensive growth vector.
\end{description}

\noindent\textbf{4. Pro Forma Financial Viability and Synergies.}
The success of the combined entity depends on realizing operational synergies and stabilizing revenue to support a sustainable capital structure.
\begin{description}
\item[(a)] Evaluates whether merger synergies and combined cash generation can lower Getty's standalone leverage on a pro forma basis.
\item[(b)] Identifies cost-cutting and operational synergies as the primary drivers for earnings growth in the absence of organic revenue expansion.
\item[(c)] Analyzes the necessity of the merger for Getty's survival, suggesting severe financial distress or substantial equity impairment if the deal fails to close.
\item[(d)] Notes that the combined entity must prove it can maintain a steady state of revenue to avoid further equity devaluation.
\end{description}

\begin{table}[h]
\centering\small
\caption{Source inventory for the Getty-Shutterstock merger environment at its May 14, 2026 information cutoff. The current release provides 34 files for Getty and Shutterstock.}
\begin{tabular}{lrl}
\toprule
Source family & Files & Contents \\
\midrule
Earnings calls & 12 & Quarterly transcripts, 2023 Q2 to 2026 Q1 \\
Press releases & 8 & February through May 2026 releases \\
Income statements & 2 & GETY and Shutterstock \\
Cash flow statements & 2 & GETY and Shutterstock \\
Balance sheet & 1 & GETY debt and liquidity \\
SEC filings & 7 & Two annual, four quarterly, one registration filing \\
Price history & 1 & 69 dated observations \\
Company profiles & 1 & GETY and Shutterstock \\
\bottomrule
\end{tabular}
\end{table}

Getty Images announced its conditional decision not to proceed with the merger on June 30, 2026, and formally terminated the agreement on July 7, which is 54 days after the environment cutoff date \citep{getty2026termination}. We use July 7 as the resolution date rather than the earlier statement of intent. The No outcome refers to termination of this proposal, not to whether the firms could ever reach another agreement, and is withheld from the research agent.

The regulatory requirement illustrates why research coverage is scored separately from prediction correctness. To satisfy this requirement, a report must assess the likelihood of the deal closing in light of proposed regulatory remedies or the absence of an outright block. Mentioning regulatory review alone is not enough. A model can correctly predict that the merger will not close while missing much of the required analysis. Conversely, it can explain the financial rationale in detail and still predict completion incorrectly.

\subsection{WULF Financing and Lease Economics}
\label{app:wulf-example}
We next consider an environment about financing business expansion. The Getty-Shutterstock example in Appendix~\ref{app:examples} examines the rationale for a merger and whether it will receive approval and close. The WULF example examines how a company's business model: construction costs and customer contracts affects its financing needs. At the March 17, 2026 cutoff, the agent compares colocation, which provides powered data-centre space to customers, with owning the GPUs as well. It then forecasts whether WULF will report at least \$500 million of net common-stock proceeds in either of its next two quarterly reports.

\textbf{The research assignment.}
The agent is asked to compare colocation with GPU ownership across three themes: predictability of returns and asset life, customer concentration and credit risk, and operational experience as a competitive advantage. It receives earnings calls, SEC filings, press releases, financial statements, company profiles, and price history, and produces a research report and a forecast. The themes are visible to the agent; the nine analytical moves below are not.

\textbf{The analytical rubric.}
The following summarizes the three moves under each theme:
\begin{description}
\item[Returns and asset life.] Explain how uncertain GPU lifetimes and rapid obsolescence make returns difficult to estimate; compare colocation lease margins with GPU-service economics; and connect construction-cost guidance and capacity milestones to equity value per share.
\item[Customer and credit risk.] Identify dependence on a small number of tenants or intermediaries; assess how strong hyperscaler credit reduces counterparty risk; and examine whether lease-termination payments would cover secured debt if a customer defaults or exits.
\item[Operational experience.] Connect management's power-market and interconnection experience to project delivery; explain why bringing large amounts of power online is a constraint; and assess whether expanding with existing customers reduces construction and redesign risks.
\end{description}

The supplied 2024 Q3 earnings call explains why WULF chose not to own GPUs: management cites its higher cost of capital than hyperscalers, its experience in power infrastructure, and the risk of competing with its customers. Repeating that choice is not enough to satisfy the first theme. The report must compare the economics of the two models and explain how equipment life, lease income, and construction costs affect returns.

The supplied 2025 Form 10-K provides a concrete example for the second theme. Once the relevant lease has commenced, Google can pay the termination fee or pay overdue rent and assume the lease if Fluidstack defaults or becomes insolvent. Naming Google as a backer is not the full analysis: the report must assess the protection this arrangement provides and whether termination payments would cover the secured debt.

\textbf{The forecast question.}
The instruction pairs this research with the following question:
\begin{quote}
In either of its next two quarterly reports, will WULF's statement of cash flows report at least \$500 million of year-to-date net proceeds from the issuance of common stock?
\end{quote}
The supplied 2025 Q4 call contains evidence on both sides. Management links its annual expansion target to \$2.5-5 billion of capital spending, but also says that currently contracted development does not require additional equity. The agent must distinguish funding for existing contracts from funding for further expansion when forming its forecast.

\textbf{How the answer is scored.}
Research coverage is the fraction of the nine moves credited by the jury under the scoring rule in Section~\ref{sec:evaluation}. Forecast correctness depends on whether either of the next two quarterly cash-flow statements reports the specified amount. Joint success requires a correct forecast and at least six of the nine moves, which exceeds the 60\% coverage threshold.

\FloatBarrier
\section{Execution, Scoring, and Statistical Details}
\label{app:protocol}
This appendix explains how we collect and score responses. We then
compare the jury with experts, examine individual grading decisions, and test how scoring choices affect the results. Coverage measures how much of the rubric a model's report completes. The jury checks the cited evidence supporting that analysis, prediction correctness is assessed separately, and joint success requires a correct forecast together with coverage strictly above 60\%.

\subsection{Trials and Valid Responses}
\label{app:execution-settings}
We evaluate every model three times per environment on the 92 \benchpublic{} environments. In the full environment, the agent receives the task instructions and source documents with access to a terminal. In the question-only condition, it receives the forecast question without documents. Internet access is prohibited in both evaluation setup. We initially (1st trial) used the jury to extract Yes/No forecasts on \benchpublic{}. After confirming agreement on the extracted verdicts across all models, we used a coding assistant (Codex; \citealp{openai_codex}) for subsequent extraction. We compare each extracted verdict with the resolved outcome. Question-only answers are scored directly from the final Yes/No response. 

Across the 92 environments and three trials, we set the effort level to High for Gemini and Fable\footnote{We use High effort for Fable 5.1 because Max-effort runs stalled even with a 100-minute API request timeout.}; Max for Opus, Sol, Terra, GLM, Kimi, and DeepSeek; and extra-high for Muse Spark, Qwen, and Grok. We use each provider's reasoning controls where available.  We allow up to four hours per attempt and up to 100 minutes per API request. We resume runs if they reach these limits, obtaining final answers for all 3,036 full-environment evaluations. A juror output is valid only if it matches the rubric's themes and is backed up by citations. Every response across all 3,036 runs is reviewed by three jurors, ensuring every model and trial is scored under identical rules. Tables~\ref{tab:frontier} and~\ref{tab:by-trial} report means and all three individual trial scores. 


\begin{table}[!ht]
\centering\small
\setlength{\tabcolsep}{3pt}
\caption{\textbf{All three frontier trials.} Joint success requires correct prediction and coverage $>60\%$. T1–T3 track success across individual 92-environment trials. Percentages cover the full set of 276 answers per model, and Effort notes the reasoning control used.}
\label{tab:frontier}
\begin{tabular}{llrrrrrr}
\toprule
Model & Effort & T1 & T2 & T3 & Joint (\%) & Coverage (\%) & Accuracy (\%) \\
\midrule
GPT 5.6 Sol & max & 33 & 26 & 28 & \textbf{31.52} & \textbf{59.17} & 69.57 \\
Claude Opus 5 & max & 26 & 26 & 28 & 28.99 & 55.89 & 73.19 \\
Claude Fable 5.1 & high & 24 & 24 & 26 & 26.81 & 55.85 & 71.38 \\
Grok 4.6 & xhigh & 26 & 24 & 23 & 26.45 & 58.25 & 64.49 \\
Qwen3.8 Max & xhigh & 20 & 24 & 27 & 25.72 & 53.54 & \textbf{75.36} \\
GLM 5.3 & max & 12 & 15 & 12 & 14.13 & 47.90 & 67.03 \\
GPT 5.6 Terra & max & 11 & 15 & 12 & 13.77 & 50.86 & 64.49 \\
Muse Spark 1.3 & xhigh & 9 & 7 & 9 & 9.06 & 47.00 & 61.23 \\
Kimi K3 & max & 6 & 8 & 10 & 8.70 & 42.92 & 67.03 \\
Gemini 3.8 Flash & high & 6 & 3 & 4 & 4.71 & 32.35 & 67.03 \\
DeepSeek V4 Pro 0813 & max & 3 & 3 & 1 & 2.54 & 37.26 & 61.96 \\
\bottomrule
\end{tabular}
\end{table}

\begin{table}[!ht]
\centering\small
\setlength{\tabcolsep}{3pt}
\caption{Full-environment results by trial. Coverage is a percentage; correct predictions are counts out of 92. Reasoning efforts match Table~\ref{tab:frontier}.}
\label{tab:by-trial}
\begin{tabular}{lrrrrrr}
\toprule
Model & Coverage T1 & T2 & T3 & Correct T1 & T2 & T3 \\
\midrule
GPT 5.6 Sol & 59.93 & 59.19 & 58.39 & 62 & 66 & 64 \\
Claude Opus 5 & 57.18 & 55.56 & 54.93 & 67 & 68 & 67 \\
Claude Fable 5.1 & 56.55 & 54.39 & 56.61 & 64 & 66 & 67 \\
Grok 4.6 & 59.31 & 58.03 & 57.41 & 57 & 61 & 60 \\
Qwen3.8 Max & 51.80 & 53.64 & 55.19 & 67 & 72 & 69 \\
GLM 5.3 & 47.73 & 47.69 & 48.29 & 62 & 59 & 64 \\
GPT 5.6 Terra & 50.96 & 51.98 & 49.65 & 58 & 59 & 61 \\
Muse Spark 1.3 & 44.82 & 46.98 & 49.20 & 57 & 54 & 58 \\
Kimi K3 & 42.84 & 42.94 & 42.98 & 61 & 59 & 65 \\
Gemini 3.8 Flash & 34.35 & 32.64 & 30.05 & 62 & 61 & 62 \\
DeepSeek V4 Pro 0813 & 37.86 & 37.28 & 36.64 & 59 & 56 & 56 \\
\bottomrule
\end{tabular}
\end{table}

\subsection{Juror Decisions and Model Scores}
Each environment’s rubric specifies the analytical moves drawn from the expert analysts’ discussion. Jurors check whether the report performs that analysis and whether its evidence supports it. The same requirements apply to every model and trial. Jurors score how much those analytical moves covered by the model. A hit means the report performs a required move. A citation objection (recorded as a taint) means the juror finds inadequate support for that analysis. Reports are not penalized simply for including additional research or counterarguments beyond the rubric. Credit depends on covering the rubric's analytical moves with adequate support. Equation~\ref{eq:coverage} combines the move-level decisions by accepting majority hits without majority taint. 

Coverage assessment reads the whole report against every required move. For each cited sentence, the verifier checks the text against the cited passages from the source documents.
 The code flags invalid file paths or source locations. If it cannot parse any citations in a report, it sends each coverage judge a warning about the missing citations.
 A citation objection does not automatically give the whole report zero credit. The scoring rule also does not require a separate support vote for every factual premise.
 A report can satisfy a move by inference without an explicit citation, while the verifier checks its supporting premises. 

Across all 3,036 responses, coverage after each juror's citation penalties is 61.96\% for Opus, 43.39\% for Gemini, and 37.60\% for Sol. The same ordering holds for the mean scores of all 11 generating models. Opus awards 18.42-30.79 percentage points more coverage than Sol, with Gemini between them (Table~\ref{tab:juror-model-matrix}). 


\begin{table}[ht]
\centering\small
\caption{Coverage awarded by each juror after its own citation penalties (\%). Each generating model contributes 276 reports, assessed by all three jurors.}
\label{tab:juror-model-matrix}
\begin{tabular}{lrrrr}
\toprule
Generating model & $n$ & Gemini & Opus & Sol \\
\midrule
GPT-5.6 Sol & 276 & 52.89 & 73.60 & 48.74 \\
Claude Opus 5 & 276 & 48.87 & 69.05 & 38.42 \\
Claude Fable 5.1 & 276 & 50.22 & 68.33 & 41.91 \\
Grok 4.6 & 276 & 51.93 & 69.31 & 43.12 \\
Qwen3.8 Max & 276 & 48.17 & 65.28 & 34.49 \\
GPT-5.6 Terra & 276 & 44.14 & 65.80 & 43.23 \\
GLM 5.3 & 276 & 41.80 & 63.32 & 40.74 \\
Muse Spark 1.3 & 276 & 41.48 & 60.89 & 37.68 \\
Kimi K3 & 276 & 38.08 & 53.82 & 31.03 \\
DeepSeek V4 Pro & 276 & 32.38 & 49.89 & 30.39 \\
Gemini 3.8 Flash & 276 & 27.32 & 42.28 & 23.86 \\
\bottomrule
\end{tabular}
\end{table}

We also check whether jurors favor answers from their own models. For each answer, we subtract the other two jurors' average coverage from that juror's score, compare this difference on own-model versus other-model answers within each environment, and average the results across environments.
 The relative differences are $+0.71$ percentage points for Gemini, $+4.33$ for Opus, and $+0.63$ for Sol. Opus gives relatively more credit to answers from its own model.

\begin{samepage}
\subsection{Blind Expert Assessment}
\label{app:human-assessment}
Given these differences between jurors, we next compare the jury's decisions with independent judgments from financial experts. This checks whether automated grading agrees with expert assessments of the analysis and its supporting evidence.
 Two experts reviewed eight anonymized reports from seven models across six environments and six companies. We selected reports near the 60$\%$ coverage threshold to examine consequential grading decisions. We also sampled reports from different environments to assess agreement beyond these borderline cases. Of the eight reports reviewed by experts, five came from the threshold-based selection and three from the additional sample. We shuffled report order and replaced model identities with anonymous identifiers. Neither expert saw model identities or automated scores.
\par\end{samepage}

The experts are financial professionals who independently assessed analytical moves and citations. Each expert spent approximately 10-12 hours on the review, or 20-24 hours in total. Together, they made 197 judgments on 110 distinct rubric requirements. Some requirements were reviewed by both experts.
 Expert A assessed 102 requirements across seven complete reports. Expert B assessed 95 across six complete reports and eight requirements in a seventh. Both gave definite ratings on 81 requirements. The primary comparison excludes Expert B's seven partial judgments.

\begin{table}[t]
\centering\small
\caption{Blind expert assessment. Expert A provided 102 ratings and Expert B provided 95. Seven partial ratings from Expert B are excluded from binary analytical comparisons. When comparing both analysis and evidence support, we exclude cases where the analysis is present but the expert could not determine whether it was supported. Agreement is measured on the same items in each row.}
\label{tab:human-assessment}
\begin{tabular}{lrrr}
\toprule
Comparison & Agree / compared & Agreement (\%) & Cohen's $\kappa$ \\
\midrule
A analytical presence vs. final jury & 78/102 & 76.5 & 0.534 \\
B analytical presence vs. final jury & 69/88 & 78.4 & 0.531 \\
A vs. B analytical presence & 57/81 & 70.4 & 0.404 \\
Human consensus vs. final jury & 51/57 & 89.5 & 0.763 \\
\midrule
A presence and support vs. final jury & 71/94 & 75.5 & 0.516 \\
B presence and support vs. final jury & 67/87 & 77.0 & 0.505 \\
\bottomrule
\end{tabular}
\end{table}

\begin{table}[t]
\centering\small
\caption{Agreement between each expert and the individual jurors or majority jury. Each column uses a fixed set of items: 102 for Expert A and 88 for Expert B. Juror decisions include their citation penalties.}

\label{tab:human-jurors}
\begin{tabular}{lrr}
\toprule
Judge & Expert A: count (\%) & Expert B: count (\%) \\
\midrule
Gemini 3.8 Flash & 80/102 (78.4) & 62/88 (70.5) \\
Claude Opus 5 & 73/102 (71.6) & 73/88 (83.0) \\
GPT-5.6 Sol & 71/102 (69.6) & 63/88 (71.6) \\
Majority jury & 78/102 (76.5) & 69/88 (78.4) \\
\bottomrule
\end{tabular}
\end{table}

Table~\ref{tab:human-assessment} compares final jury credit with the experts' definite analytical-presence decisions. The jury agrees with Expert A on 78/102 decisions and with Expert B on 69/88. On the shared 81 requirements, the experts agree on 57/81 (70.4\%), while the jury agrees with each expert on 63/81 (77.8\%). Expert B credits 22 requirements that Expert A rejects, whereas Expert A credits two that Expert B rejects. The experts agree on 57 requirements, and the jury agrees with them on 51. In the other six cases, it withholds four credits the experts award and awards two credits they reject.

Disagreements usually came down to whether an implicit argument or related evidence was sufficient to meet a rubric requirement. For example, Expert B credited broad discussions of customer retention, whereas the jury insisted on the specific mechanism of expanding deployments within a fleet. Expert A was generally stricter than Expert B. Compared to Expert A, the jury was more lenient, awarding credit on 17 requirements that A rejected and withholding credit on seven that A approved. Against the more generous Expert B, the jury withheld credit on 17 requirements B approved and awarded credit on only two that B rejected.

We also evaluate agreement when requiring both analytical moves and verified citation support, excluding cases with unresolved support or partial judgments. This gives 94 comparisons for Expert A and 87 for Expert B, with 75.5\% and 77.0\% agreement, respectively (Table~\ref{tab:human-assessment}). 

 Removing citation penalties leaves agreement with Expert A unchanged at 78/102 and slightly increases agreement with Expert B from 69/88 to 71/88. In the two additional agreements, both Expert B and the jury recognize the required analysis, but the jury withholds credit because of citation penalties. We also check how Expert B's seven partial ratings affect agreement. Treating all seven as no credit gives 75/95 agreement (78.9\%); treating them all as credit gives 70/95 (73.7\%). Table~\ref{tab:human-jurors} compares each expert with the individual jurors and the majority jury.

\subsection{Qualitative Examples of Jury Decisions}
\label{app:jury-qualitative}

Table~\ref{tab:jury-qualitative} shows that citation penalties can apply to both uncited factual claims and claims not supported by their citations. In the Samsara example, unsupported claims about carrier bankruptcies and freight-market conditions cause the analysis to lose credit. In the FND example, judges flag full-year guidance incorrectly applied to the first quarter. By contrast, Getty's deleveraging and cost-synergy arguments retain credit because the cited sources support them.

In Kimi K3's Samsara sales-efficiency report, the answer states that ``carrier bankruptcies and capex cuts among trucking customers should pressure the transportation vertical.'' The Opus coverage judge identifies the evidence gap explicitly: ``this assertion is uncited.'' The report also attributes transportation growth to ``a tightening freight market,'' whereas its cited passages establish vertical growth, efficiency, and safety, not the asserted freight-market conditions. All three judges, Opus, Sol, and Gemini, recognize the freight-recession analysis but raise citation objections to its supporting claims. The recorded decision is therefore \textbf{three analytical hits, three citation objections, and zero credit for the requirement}. These errors demonstrate how the jury separates analytical coverage from factual grounding: the report was penalized for both an uncited claim and a cited claim that the sources did not support.

\begin{table}[ht]
\centering\small
\caption{Examples of citation checks. A hit means the required analysis is present. An objection means its evidence is missing or inadequate. Counts are out of three judges.}
\label{tab:jury-qualitative}
\begin{tabular}{p{0.19\linewidth}p{0.45\linewidth}p{0.25\linewidth}}
\toprule
Case & Claim and evidence check & Observed decision \\
\midrule
Samsara: freight conditions & An uncited bankruptcy/capex premise accompanies a freight-market characterization not established by its citations. & 3 hits, 3 objections; no credit. \\
FND: guidance period & Full-year comparable-sales guidance is applied to the first quarter. & Opus and Sol flag the period mismatch. \\
Getty: deleveraging & Cash-flow-supported deleveraging is consistent with the supplied call's stated merger rationale. & 3 hits, 0 objections; credit. \\
Getty: cost synergies & The supplied call explicitly identifies cost synergies as the transaction rationale. & 3 hits, 0 objections; credit. \\
\bottomrule
\end{tabular}
\end{table}

The FND report confuses full-year and first-quarter guidance. The cited call forecasts full-year comparable-sales growth of $-2\%$ to $+1\%$, but describes the initial first-quarter estimate as only slightly negative. Opus and Sol flag this error, the source does not support the agent's claim that first-quarter sales would fall by more than $2\%$. The judges checked whether the cited source supported the claim, time period and numbers.

To examine decisions that retained credit, we also inspected six majority-credited moves in three reports on the Getty-Shutterstock merger's strategic rationale. We selected two per report, prioritizing uncited factual statements and conclusions. We found support for the core argument in four moves. Support for the other two remained unresolved.
 These selected examples show how the scoring works.

For example, Sol's cash-flow-supported deleveraging argument received three analytical votes and no citation objections. Getty's supplied 2024 Q4 earnings call explicitly links the merger to deleveraging and increased cash flow, supporting the inference. Gemini's cost-synergy argument likewise received three analytical votes and no citation objections: the supplied 2025 Q3 call identifies cost synergies as the transaction's rationale. Its additional suggestions about cloud vendors and advertising auctions were not established by that passage, but were not penalized simply for adding reasoning or counterarguments beyond the rubric.

The unresolved cases concerned Fable's assertion of Shutterstock's historically lower leverage and Gemini's attribution of a below-$3.5\times$ leverage target to management. Both received three analytical votes , but our inspection could not confirm their support. These uncited claims were not included in the citation-checking stage. In the next iteration, we plan to give each coverage juror the full report, rubric, all supplied source documents, and citation-checking results, and ask it to verify factual claims, including uncited ones, before awarding credit. 

\subsection{Citation Frequency and Research Coverage}
We examine citation frequency to assess whether coverage primarily reflects the volume of citations in a report. The analysis includes 307,178 citation-unit occurrences across all 3,036 evaluated answers. A citation unit references one file and locator. Repeated references count separately, while unique-file counts measure source breadth. Counts use the scorer's paragraph normalization and citation parser. Table~\ref{tab:citation-frequency} reports all eleven models. Qwen averages 245.8 citations per answer and 53.54\% coverage, whereas Sol averages 85.6 citations and 59.17\% coverage. Across answers, citation count has only a weak positive correlation with final coverage ($r=0.104$).

We test whether reports with more citations receive higher coverage. We consider three measures: citations per 1,000 prose tokens, log citation count, and the number of distinct sources cited. For each measure, we examine both the analysis recognized by a majority of jurors and the final coverage after citation penalties.
The regressions account for differences across environments, models, and trials using fixed effects. They also account for report length using log prose length and its square. We measure length with the same tokenizer for all models, excluding citation strings. Standard errors are clustered by environment. We apply Holm correction across the six tests: three citation measures and two outcomes.

Citation density has a small negative association with final coverage after adjustment (partial r = -0.099; Holm-adjusted p \textless{} 0.001). Ten additional citations per 1,000 prose tokens correspond to 4.02 percentage points lower coverage.
 The association with majority analytical presence before citation penalties is approximately zero (partial $r=-0.012$). Comparing repeated answers within the same model and environment gives corresponding partial correlations of $-0.002$ before citation penalties and $-0.092$ after. The final-coverage association remains significant after correction ($p=0.010$). Citation volume alone therefore does not explain research coverage.

\begin{table}[!ht]
\centering\small
\setlength{\tabcolsep}{3pt}
\caption{Citation behavior on all 3,036 frontier answers (276 per model). Cites is the average number of citations per answer, counting repeats. Files is the average number of distinct source-file paths cited. Density is the average number of citations per 1,000 final-answer tokens, measured with the same tokenizer for all models. The regression uses a different denominator: prose tokens with the citation text removed. For each verifier and answer, we calculate the fraction of checked citations involved in a citation objection. Objections reports the average of these fractions.}
\label{tab:citation-frequency}
\begin{tabular}{llrrrrr}
\toprule
Model & Effort & Cites & Files & Density & Objections (\%) & Coverage (\%) \\
\midrule
GPT 5.6 Sol & max & 85.6 & 9.8 & 14.79 & 8.30 & 59.17 \\
Grok 4.6 & xhigh & 133.9 & 16.0 & 13.75 & 17.60 & 58.25 \\
Claude Opus 5 & max & 144.5 & 19.2 & 14.21 & 18.88 & 55.89 \\
Claude Fable 5.1 & high & 99.6 & 15.8 & 12.86 & 18.91 & 55.85 \\
Qwen3.8 Max & xhigh & 245.8 & 20.6 & 14.65 & 21.33 & 53.54 \\
GPT 5.6 Terra & max & 60.2 & 7.8 & 13.60 & 5.91 & 50.86 \\
GLM 5.3 & max & 52.3 & 9.6 & 14.01 & 5.75 & 47.90 \\
Muse Spark 1.3 & xhigh & 86.4 & 9.4 & 15.59 & 11.31 & 47.00 \\
Kimi K3 & max & 60.6 & 10.8 & 13.21 & 32.56 & 42.92 \\
DeepSeek V4 Pro 0813 & max & 34.5 & 7.6 & 13.08 & 11.25 & 37.26 \\
Gemini 3.8 Flash & high & 109.5 & 15.0 & 13.76 & 44.48 & 32.35 \\
\bottomrule
\end{tabular}
\end{table}

\subsection{Citation-Processing Sensitivity}
\label{app:citation-processing}
To assess sensitivity to citation parsing, we checked explicit file and locator references. This identified 179 reports across \benchpublic{}, \benchheldout{}, and \benchprivate{} for targeted review. We reran citation verification and dependent coverage with the same three jurors on 155 reports whose source excerpts could be matched unambiguously. Twenty-four retained their original grades. No answer, rubric, or forecast label changed. The check changed 78 coverage scores: 67 on Public, nine on Heldout, and two on Private. Nineteen reports crossed the strict $>60\%$ threshold (16, two, and one, respectively). On the 3,036 Public reports, pooled joint success changed from 527 to 531, leaving the main comparison intact. This check measures score sensitivity to citation parsing.

We have implemented and tested a citation tool that the public release will use in place of free-form file-path citations. The tool validates the selected source location and returns a citation ID tied to the excerpt. The grader reconstructs that excerpt from the frozen environment sources, avoiding the free-form citation extraction issues described above. The jury continues to assess whether the evidence supports the claim.

\subsection{Panel and Aggregation Sensitivity}
To test whether model orderings depend on panel composition or aggregation, we recompute scores for all 3,036 responses. Results average the 276 responses within each model. Removing each juror in turn retains Spearman correlations of 0.964-0.973 for coverage ranks and 0.736-0.909 for joint-success ranks across the eleven models (Table~\ref{tab:jury-robustness}). With two jurors, both must recognize a move for it to receive credit. Removing a juror can also remove a citation objection and restore credit. Scores can therefore change in either direction.

We also test a variant where we count a move as successful only if at least two jurors independently recognize it without raising a citation objection. This contrasts with our original rule, which takes separate majority votes on analytical presence and citation objections.
 Coverage and joint-success rankings retain correlations of 0.991 and 0.982, respectively. Coverage ordering is more stable than the thresholded joint endpoint under panel removal.

We compare each variant's model ranking with the original three-juror ranking using Spearman correlation, accounting for ties.
\begin{table}[h]
\centering\small
\caption{Model-ranking sensitivity to juror removal and aggregation. Entries are tie-aware Spearman correlations across eleven models relative to the original three-juror rule, with all variants scored on the same 3,036 responses. Joint success requires a correct prediction and coverage strictly above 60\%.}
\label{tab:jury-robustness}
\begin{tabular}{lrr}
\toprule
Alternative panel or rule & Coverage & Joint success \\
\midrule
Without Gemini & 0.964 & 0.909 \\
Without Opus & 0.973 & 0.736 \\
Without Sol & 0.973 & 0.782 \\
Majority of individually unpenalized hits & 0.991 & 0.982 \\
\bottomrule
\end{tabular}
\end{table}

\FloatBarrier
\section{Joint Success and Question-Only Results}
\label{app:joint}
In this appendix, we first vary the coverage threshold used for joint success. We then compare full-environment and question-only forecasts, and examine results by outcome date. All analyses use saved responses from the \benchpublic{} evaluation.

Joint success combines a correct forecast with a minimum level of completed analysis. We calculate it from saved prediction verdicts and exact accepted-move counts, using the same rule for every model and weighting the three trials equally. A report earns credit for a whole number of moves. Two different percentage thresholds can therefore require the same number of credited moves. With 12 moves, both $>60\%$ and $\geq62.5\%$ require eight credits. With 15 moves, both require ten.

Table~\ref{tab:joint-sensitivity} varies the coverage threshold while continuing to require a correct prediction. The table also reports results for the weighted rule $0.6P + 0.4C \geq 0.85$, where $P$ is binary correctness and $C$ is fractional coverage. Passing this rule requires a correct forecast and at least 62.5\% coverage.

\begin{table}[!ht]
\centering\small
\setlength{\tabcolsep}{3pt}
\caption{Joint-success sensitivity across all three trials. Every threshold additionally requires a correct prediction; all columns use 276 answers per model.}
\label{tab:joint-sensitivity}
\begin{tabular}{lrrrr}
\toprule
Model & $>50\%$ & $>60\%$ & $\geq62.5\%$ & $>70\%$ \\
\midrule
GPT 5.6 Sol & \textbf{47.46} & \textbf{31.52} & \textbf{31.16} & 13.41 \\
Claude Opus 5 & 43.12 & 28.99 & 27.90 & 14.13 \\
Claude Fable 5.1 & 39.13 & 26.81 & 25.72 & 13.04 \\
Grok 4.6 & 40.58 & 26.45 & 26.09 & \textbf{14.86} \\
Qwen3.8 Max & 40.22 & 25.72 & 24.64 & 13.04 \\
GLM 5.3 & 25.36 & 14.13 & 13.41 & 4.35 \\
GPT 5.6 Terra & 22.46 & 13.77 & 13.04 & 5.80 \\
Muse Spark 1.3 & 22.10 & 9.06 & 8.70 & 2.90 \\
Kimi K3 & 20.29 & 8.70 & 8.33 & 3.26 \\
Gemini 3.8 Flash & 10.87 & 4.71 & 4.71 & 2.54 \\
DeepSeek V4 Pro 0813 & 7.25 & 2.54 & 2.54 & 1.45 \\
\bottomrule
\end{tabular}
\end{table}

Of the 3,036 reports, 853 (28.10$\%$) cover more than 60$\%$ of the required analytical moves. Among these, 531 (62.25$\%$) predict correctly, compared with 1,519 of the remaining 2,183 reports (69.58$\%$; Table 15). Reports above the coverage threshold therefore have lower forecast accuracy in this combined sample of models and environments. We report research coverage and forecast accuracy separately because a correct forecast does not necessarily mean the report covers the required analytical moves. 

\begin{table}[h]
\centering\small
\caption{Research coverage and prediction correctness across 3,036 full-environment responses: eleven models, 92 \benchpublic{} environments, and three trials. The 531 responses with coverage $>60\%$ and a correct prediction satisfy joint success.}
\label{tab:coverage-prediction-cross}
\begin{tabular}{lrrr}
\toprule
 & Correct prediction & Incorrect prediction & Total \\
\midrule
Coverage $>60\%$ & 531 & 322 & 853 \\
Coverage $\leq60\%$ & 1,519 & 664 & 2,183 \\
\midrule
Total & 2,050 & 986 & 3,036 \\
\bottomrule
\end{tabular}
\end{table}

\subsection{Question-Only Predictions}
\begin{figure}[!ht]
\centering
\includegraphics[width=\linewidth]{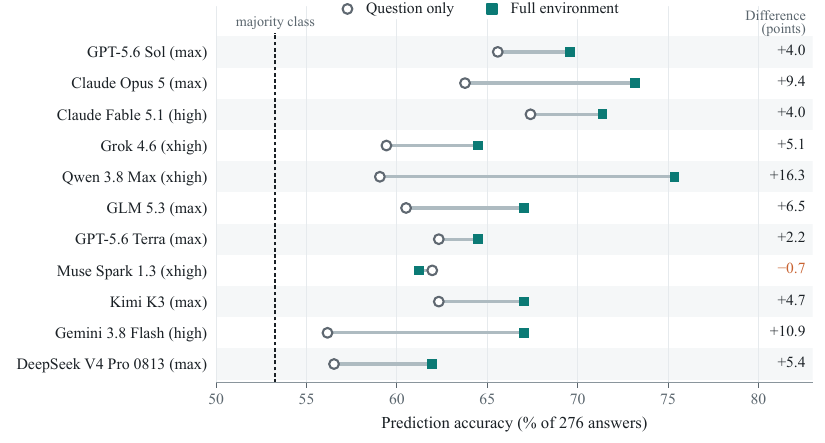}
\caption{\textbf{Ten models have higher observed prediction accuracy in the full environment.} Each point aggregates three trials on all 92 \benchpublic{} environments (276 predictions per model and condition). Open circles show question-only accuracy, and teal squares show full-environment accuracy. The right column gives full-environment minus question-only accuracy in percentage points, and the dashed line marks the 53.26\% majority-class baseline.}
\label{fig:question-only}
\end{figure}

The question-only condition measures prediction accuracy without research documents or internet access. Table~\ref{tab:qo-all} includes all three trials, with 276 valid responses per model and 92 per trial. Accuracy is total correct divided by 276, equal to the mean of the three trials. These responses contain no research reports and therefore have neither coverage scores nor joint-success rates. With the full environment, ten of eleven models achieve higher observed forecast accuracy than with the question alone, including gains of 16.30 percentage points for Qwen3.8 Max and 10.87 points for Gemini 3.8 Flash (Figure~\ref{fig:question-only}).

\begin{table}[!ht]
\centering\small
\setlength{\tabcolsep}{4pt}
\caption{Question-only prediction accuracy over all three trials. Each trial contains 92 valid answers per model, and the aggregate uses all 276. Effort denotes the configured reasoning setting. These results show forecasting without the research documents.}
\label{tab:qo-all}
\begin{tabular}{llrrrr}
\toprule
Model & Effort & Q1 & Q2 & Q3 & Aggregate (\%) \\
\midrule
GPT 5.6 Sol & max & 61/92 & 61/92 & 59/92 & 65.58 \\
Claude Opus 5 & max & 61/92 & 55/92 & 60/92 & 63.77 \\
Claude Fable 5.1 & high & 64/92 & 61/92 & 61/92 & \textbf{67.39} \\
Grok 4.6 & xhigh & 56/92 & 52/92 & 56/92 & 59.42 \\
Qwen3.8 Max & xhigh & 55/92 & 53/92 & 55/92 & 59.06 \\
GLM 5.3 & max & 53/92 & 54/92 & 60/92 & 60.51 \\
GPT 5.6 Terra & max & 58/92 & 60/92 & 54/92 & 62.32 \\
Muse Spark 1.3 & xhigh & 56/92 & 59/92 & 56/92 & 61.96 \\
Kimi K3 & max & 54/92 & 56/92 & 62/92 & 62.32 \\
Gemini 3.8 Flash & high & 50/92 & 54/92 & 51/92 & 56.16 \\
DeepSeek V4 Pro 0813 & max & 52/92 & 48/92 & 56/92 & 56.52 \\
\bottomrule
\end{tabular}
\end{table}

\subsection{Resolution Timing}
\label{app:resolution-timing}
We examine resolution timing in two ways: a common calendar split and splits at each model's documented knowledge cutoff. Both analyses use all 92 \benchpublic{} environments. The calendar split in Table~\ref{tab:calendar-resolution} divides the environments at March 31, 2026, retaining all three trials on 23 earlier and 69 later environments. Table~\ref{tab:public-resolution-dates} gives the complete chronology. Ten of eleven models score higher on the later group. For example, GPT-5.6 Sol rises from 57.97\% to 73.43\%, so later-resolving environments are not uniformly harder. For each model-specific split, we report the number of environments, Yes/No outcome counts, and forecast accuracy on both sides of the cutoff.

\begin{table}[ht]
\centering\small
\caption{Full-environment prediction accuracy on \benchpublic{} before and after a common resolution-date boundary. The split is through March 31, 2026 versus April 1, 2026 onward. Each model contributes three trials on 23 earlier environments (69 predictions) and 69 later environments (207 predictions). Counts and percentages use all attempts in each subset. The groups contain different environments, and recorded resolution dates refer to disclosures, events, or window cutoffs.}
\label{tab:calendar-resolution}
\begin{tabular}{lrrrr}
\toprule
& \multicolumn{2}{c}{Through March 31} & \multicolumn{2}{c}{After March 31} \\
Model & Correct / 69 & Accuracy (\%) & Correct / 207 & Accuracy (\%) \\
\midrule
Claude Opus 5 & 48 & 69.57 & 154 & 74.40 \\
Qwen 3.8 Max & 54 & 78.26 & 154 & 74.40 \\
Claude Fable 5.1 & 44 & 63.77 & 153 & 73.91 \\
GPT-5.6 Sol & 40 & 57.97 & 152 & 73.43 \\
Gemini 3.8 Flash & 39 & 56.52 & 146 & 70.53 \\
GLM 5.3 & 42 & 60.87 & 143 & 69.08 \\
Kimi K3 & 45 & 65.22 & 140 & 67.63 \\
GPT-5.6 Terra & 40 & 57.97 & 138 & 66.67 \\
Grok 4.6 & 40 & 57.97 & 138 & 66.67 \\
DeepSeek V4 Pro & 37 & 53.62 & 134 & 64.73 \\
Muse Spark 1.3 & 41 & 59.42 & 128 & 61.84 \\
\bottomrule
\end{tabular}
\end{table}

Table~\ref{tab:model-cutoffs} lists the documented cutoffs. These include February 16, 2026 for GPT-5.6 Sol and Terra, March 2026 for Gemini 3.8 Flash, May 2026 for Claude Opus 5, and June 2026 for Claude Fable 5.1 \citep{openai2026solcutoff,openai2026terracutoff,google2026geminicutoff,anthropic2026fablecutoff}. We compare these cutoffs with the recorded outcome dates in Table~\ref{tab:public-resolution-dates}. When only a month is specified, we use its final day as the cutoff.

Table~\ref{tab:temporal-subset} compares forecast accuracy before and after each model's documented knowledge cutoff across all 92 environments. Counterintuitively, GPT-5.6 Sol scores higher on its larger post-cutoff group: 70.13\% on 77 environments, versus 66.67\% on 15 pre-cutoff environments. Gemini 3.8 Flash similarly rises from 56.52\% on 23 pre-cutoff environments to 70.53\% on 69 post-cutoff environments. These gains echo the common March 31 split (Table~\ref{tab:calendar-resolution}). Claude Opus 5 also improves, and GPT-5.6 Terra is essentially unchanged, while Grok 4.6 and Claude Fable 5.1 decline. Together, these comparisons show no consistent pre-cutoff advantage that would suggest reliance on memorized outcomes. 

\begin{table}[!ht]
\centering\small
\setlength{\tabcolsep}{4pt}
\caption{Full-environment forecast accuracy before and after each model's documented knowledge cutoff. Each row covers all 92 \benchpublic{} environments and 276 predictions. Dates follow Table~\ref{tab:public-resolution-dates}; Before includes the cutoff date, and month-only cutoffs use month-end. Envs counts environments, and Yes/No gives their outcome counts.}
\label{tab:temporal-subset}
\begin{tabular}{@{}lrrrrrr@{}}
\toprule
Model & \multicolumn{3}{c}{Before cutoff} & \multicolumn{3}{c}{After cutoff} \\
\cmidrule(lr){2-4}\cmidrule(lr){5-7}
 & Envs & Yes/No & Accuracy (\%) & Envs & Yes/No & Accuracy (\%) \\
\midrule
GPT-5.6 Sol & 15 & 9/6 & 66.67 & 77 & 34/43 & 70.13 \\
GPT-5.6 Terra & 15 & 9/6 & 64.44 & 77 & 34/43 & 64.50 \\
Grok 4.6 & 7 & 3/4 & 71.43 & 85 & 40/45 & 63.92 \\
Gemini 3.8 Flash & 23 & 12/11 & 56.52 & 69 & 31/38 & 70.53 \\
Claude Opus 5 & 60 & 29/31 & 72.78 & 32 & 14/18 & 73.96 \\
Claude Fable 5.1 & 71 & 36/35 & 74.65 & 21 & 7/14 & 60.32 \\
\bottomrule
\end{tabular}
\end{table}

\begin{table}[!htbp]
\centering\small
\caption{Provider-documented knowledge cutoffs used for the model-specific before-and-after comparisons, checked September 9, 2026. ``Not stated'' means the cited official source does not specify a cutoff.}
\label{tab:model-cutoffs}
\begin{tabular}{lll}
\toprule
Model & Documented cutoff & Official source \\
\midrule
GPT-5.6 Sol & February 16, 2026 & \citet{openai2026solcutoff} \\
GPT-5.6 Terra & February 16, 2026 & \citet{openai2026terracutoff} \\
Grok 4.6 & February 1, 2026 & \citet{xai2026grokcutoff} \\
Gemini 3.8 Flash & March 2026 & \citet{google2026geminicutoff} \\
Claude Opus 5 & May 2026 & \citet{anthropic2026modeloverview} \\
Claude Fable 5.1 & June 2026 & \citet{anthropic2026fablecutoff} \\
Qwen 3.8 Max & Not stated & \citet{qwen2026modeloverview} \\
GLM 5.3 & Not stated & \citet{zai2026glm53} \\
Kimi K3 & Not stated & \citet{moonshot2026kimik3} \\
DeepSeek V4 Pro & Not stated & \citet{deepseek2026v4card} \\
Muse Spark 1.3 & Not stated & \citet{meta2026spark13} \\
\bottomrule
\end{tabular}
\end{table}

\FloatBarrier
\section{Training Configuration}
\label{app:reward}
\subsection{Iterative Judge Prompt Tuning}
We improved the per-turn credit assignment by tuning judge prompts by inspecting failures in short training probes. We used coding agents to review sampled reports and individual credit and blame assignments and revised the objective or attribution instructions. They then reassessed saved trajectories before the next experiment. We call this human-directed, coding-assisted procedure meta-prompting. GEPA similarly uses trajectory feedback to refine prompts \citep{agrawal2025gepa}. We fixed the judge instructions through this iterative process before the reported training run. The run used 21 updates to cover the 165 environments once, with three extra environment groups to fill the final batch. We compare the resulting training method with outcome-only and magnitude-matched controls on held-out environments that were not used to develop the prompts. 

Our frontier evaluation found that higher research coverage did not consistently imply higher forecast accuracy (Table~\ref{tab:coverage-prediction-cross}). In preliminary runs, fixed mixtures of binary forecast correctness and research coverage did not consistently improve both metrics. This motivated our use of a synthesis-aware reward, alongside prior work on reward interference \citep{ramesh2026playingwithfire} and reward composition \citep{li2024rewardcomposition}.

We compare a reconstructed weighted reward, $0.8C+0.2P$, with the holistic objective, where $C$ is coverage and $P$ the forecast score. Across the same 384 trajectories from 48 environments, the reconstructed reward's correlation with binary correctness was 0.4904, compared with 0.9074 for the holistic objective. This comparison checks whether its scores follow the intended reward rule: The higher correlation shows that the holistic reward tracks forecast correctness more closely, as intended. A report can cover much of the required research while reaching the wrong forecast, as illustrated in Appendix~\ref{app:frontier-cases}. The synthesis judge therefore assesses the strongest relevant evidence on both sides, the connection between analysis, conclusion, and the forecast. A wrong prediction must score below 0.5 (Appendix~\ref{app:full-reward-prompts}). A correct prediction can also score below 0.5 if the research or synthesis is materially incomplete. High scores require both substantial accepted coverage and strong synthesis.

During prompt tuning, inspection by coding agents led to three attribution rules. First, assess evidence gathering separately from later inference: sound retrieval is not blamed simply because the forecast is wrong. Second, distinguish reasonable exploration that finds no match from repetition that adds no information. Third, assess a report-writing turn by its claims and conclusions, not by the act of writing. These rules address failures observed in the sampled trajectories. Once fixed, the instructions are applied retrospectively by the reward judge.

\subsection{Policy Loss and Token Weights}
\label{app:policy-loss}
The policy loss applies each turn's assigned advantage to its eligible assistant tokens. Let $a_{it}$ be a sampled assistant token and $h_{it}$ its preceding conversation and token history, including tool observations. The ratio $\rho_{it}=\pi_\theta(a_{it}\mid h_{it})/\pi_{\rm roll}(a_{it}\mid h_{it})$ compares the current policy with the policy that generated the rollout. The advantage is $A_{it}$ from Equation~\ref{eq:credit}. With eligibility mask $m_{it}$ and clipping width $\epsilon=0.2$, the policy surrogate is
\begin{equation}
 \mathcal L=-\frac{\sum_{i,t}m_{it}\min\{\rho_{it}A_{it},
 \operatorname{clip}(\rho_{it},1-\epsilon,1+\epsilon)A_{it}\}}
 {\sum_{i,t}m_{it}}.
 \label{eq:loss}
\end{equation}
The denominator includes eligible tokens with zero attribution. Each eligible token in a turn receives the same coefficient, without division by turn length, so a longer credited turn contributes more terms to the loss. The configuration uses no learned value critic and no reference-policy KL penalty.

The judge decides whether each action receives positive or negative weight. The absolute difference between the trajectory score and its group mean sets the overall scale. The token weights therefore need not sum to zero within a group. These turn weights change the training objective. They are not an unbiased estimate of the policy gradient obtained from the original trajectory reward. 

For turn $j$ of trajectory $i$ in environment group $g$, $r^+_{gij}$ and $r^-_{gij}$ are the judge's reasoning success and failure weights, each in $[0,1]$. Signed tool utility is $u_{gij}\in[-1,1]$. Check Algorithm \ref{alg:turncredit} for credit assignment and policy update.

\begin{algorithm}[!ht]
\caption{\turncredit{} reward construction and policy update.}
\label{alg:turncredit}
\footnotesize
\begin{algorithmic}[1]
\Require Environments $x_1,\ldots,x_8$, rollout policy $\pi_{\rm roll}$, current policy $\pi_\theta$, training-only rubrics and outcomes, reward judge $J$, clipping width $\epsilon=0.2$.
\For{$g=1,\ldots,8$}
  \State Sample $\{\tau_{gi}\}_{i=1}^{8}\sim\pi_{\rm roll}(\cdot\mid x_g)$ using the Bash agent.
  \For{$i=1,\ldots,8$}
    \If{$\tau_{gi}$ has a terminal, empty, or missing answer}
      \State Apply terminal score $-0.5$ and the applicable loss mask (Appendix~\ref{app:terminal-handling}).
    \Else
      \State Grade the final answer against the hidden rubric and resolved outcome.
      \State Draw two synthesis judgments; average valid scores to obtain $s_{gi}$.
      \State Draw two turn attributions using the completed trajectory and synthesis diagnoses.
      \State Validate and aggregate the success, failure, and tool-utility channels separately.
    \EndIf
  \EndFor
  \If{the group has terminal siblings}
    \State Set $\bar s_g$ to the nonterminal-score mean (zero if none).
  \Else
    \State Set $\bar s_g=\frac{1}{8}\sum_{i=1}^{8}s_{gi}$.
  \EndIf
  \State For each ordinary rollout, set $\Delta_{gi}=s_{gi}-\bar s_g$.
  \For{each ordinary trajectory $i$ and assistant turn $j$}
    \State For a report-writing turn, set $u_{gij}\gets0$; retain its reasoning weights.
    \State $w^+_{gij}\gets\max(r^+_{gij},u_{gij},0)$; $w^-_{gij}\gets\max(r^-_{gij},-u_{gij},0)$.
    \State Assign $A_{git}=|\Delta_{gi}|(w^+_{gij}-w^-_{gij})$ to each eligible token $t$ in turn $j$.
  \EndFor
\EndFor
\State Mask prompt and tool-observation tokens; keep zero-attribution eligible tokens in the denominator.
\State Update $\pi_\theta$ with the eligible-token-mean clipped loss in Equation~\ref{eq:loss}.
\end{algorithmic}
\end{algorithm}

\begin{table}[ht]
\centering\small
\caption{Optimization and sampling settings for retrospective turn-credit training. The loss averages over eligible tokens, including those with zero attribution.}
\label{tab:training-settings}
\begin{tabular}{lp{0.64\linewidth}}
\toprule
Setting & Value \\
\midrule
Schedule & 21 updates; 8 environments per update; 8 rollouts per environment \\
Policy batch & 64 trajectories; microbatch 1; one optimizer step per rollout batch \\
Optimizer & Adam; constant learning rate $10^{-5}$; $\beta_1=0.9$, $\beta_2=0.98$; weight decay 0 \\
Precision and clipping & BF16; gradient clipping 1.0; policy-ratio clipping $[0.8,1.2]$ \\
Adapter & LoRA rank 16, scaling 32; attention Q/K/V/output and MLP up/gate/down projections \\
Loss reduction & Eligible-token mean; no group-standard-deviation normalization \\
Regularization & Reference-policy KL coefficient 0; entropy coefficient 0 \\
Policy sampling & High reasoning effort; temperature 1 \\
Reward draws & Two objective draws, followed by two attribution draws \\
\bottomrule
\end{tabular}
\end{table}

\begin{samepage}
\subsection{Reward Inputs and Attribution}
Algorithm~\ref{alg:turncredit} uses two judging stages, both using Gemini 3.7 Flash with different prompts. The report judge scores the completed report and assesses how its analysis supports the forecast. The attribution judge then uses the full trajectory together with that score and assessment to assign reasoning credit or blame to individual turns and score their tool use. We request two assessments at each stage. Each turn-level assignment includes the turn number and an explanation; turns with zero weight are omitted.
 Appendix~\ref{app:full-reward-prompts} gives the full system prompts and input-message fields
\par\end{samepage}

Both stages run after generation and leave the completed trajectory unchanged. The objective receives the task instruction, hidden grading rubric, final answer, authoritative grading verdict, and objective context. Attribution receives those inputs plus the averaged objective score, synthesis diagnoses, and ordered turn records. Each record contains its index, assistant content, tool calls, and subsequent tool outputs. The judge can therefore assess an early action using later turns and the final answer. The acting policy sees neither the grading verdict nor the hidden rubric.

We average valid objective draws and validate and aggregate each attribution channel separately. If a required attribution field is missing or malformed, we ask the judge to repair its response, with up to three repair calls. We do not fill in the missing field by assigning uniform credit. Judge temperature inherits the provider default.

The attribution judge prompt distinguish evidence gathering from its later use. A relevant search with no match is neutral; retrieval that supports later analysis and verification that identifies or repairs a real defect can earn credit. Repeated calls receive negative tool utility when they add no information. Sound evidence gathering is not blamed solely because later synthesis reaches a wrong conclusion. Report-writing turns receive no tool-utility weight but remain eligible for reasoning credit or blame.

Unlisted turns receive zero weight. We aggregate reasoning and tool-utility channels independently across usable draws, separating positive and negative tool utility before aggregation.
Equation~\ref{eq:credit} takes the elementwise maximum of the positive maps and, separately, the negative maps to avoid additive double counting. Token positions identify which assistant tokens belong to each turn. The loss mask excludes tokens that are not training targets. There is no additional turn-length normalization. Appendix~\ref{app:saved-credit} illustrates these assignments with a qualitative example. 

\subsection{Terminal Handling}
\label{app:terminal-handling}

In the final configuration, malformed tool calls and empty terminal
outputs receive a reward of $-0.5$. The same penalty applies when the
agent stops without a valid \texttt{answer.md}. These failures bypass
the report and attribution judges and receive negative weight across
their eligible assistant tokens.

Terminal failures are excluded when computing the group baseline.
The baseline is the mean score of the remaining nonterminal rollouts,
or zero if none remain. Failed rollouts therefore do not lower the
baseline for the other trajectories.

Budget limits are handled separately. Rollouts that reach a turn or
token limit, or exhaust the context window, without a valid answer
are excluded from the policy loss. If a valid answer already exists,
we retain it for scoring; any assistant suffix that caused the
truncation is excluded from the loss.

\FloatBarrier
\section{Credit Assignment and Behavioral Evidence}
\label{app:credit}
\subsection{Model Behavior}
We first compare training behavior under outcome-only and turn-level supervision, then examine saved credit assignments and a separately trained magnitude-matched control. Both forms of supervision use the same group-centered, synthesis-aware objective. Outcome-only training assigns signed $\Delta_i$ to every eligible token; turn-level training assigns action-specific signs and magnitudes through Equation~\ref{eq:credit}. The initial comparison therefore includes differences in update scale as well as credit assignment.

The high-effort outcome-only evaluation remained incomplete because trajectories repeatedly reached turn or context limits despite the higher budgets described in Appendix~\ref{app:heldout-protocol}. We report its completed low- and medium-effort evaluations.

Similarly, outcome-only trajectories are longer and more repetitive during training. Over updates 16--20, unscaled outcome-only GRPO averages 198.7 turns, compared with 99.6 for turn-level training, and repeats 12.59\% of commands compared to 1.23\%. Magnitude matching reduces (compared to Outcome only) these to 120.3 turns and 3.95\% repeated commands (Table~\ref{tab:training-behavior-summary}). Compared with unscaled outcome-only GRPO, turn-level coverage is higher in 39 of the 40 matched environment groups across these five updates.

Magnitude-matched trajectories remain shorter than unscaled outcome-only trajectories, but their turn count rises late in training while token entropy falls (Figure~\ref{fig:magnitude-training}B--C). From update 16 to 20, mean turns increase from 107.9 to 134.8 while entropy falls from 0.187 to 0.143. Both outcome-only controls use more turns late in training. \turncredit{} keeps its turn count comparatively stable while using more tokens per turn, alongside higher research coverage (Figure~\ref{fig:magnitude-training}). Fewer turns do not necessarily mean fewer tokens: magnitude matching uses fewer trajectory tokens than turn-level training over these updates. Table~\ref{tab:training-behavior-summary} compares all three policies during training.

\begin{table}[!ht]
\centering\small
\setlength{\tabcolsep}{4pt}
\caption{Training behavior over updates 16--20. Means use 320 trajectories per policy. Trajectory tokens include accepted assistant output and tool observations; repeated commands are exact repeats.}
\label{tab:training-behavior-summary}
\begin{tabular}{@{}lrrr@{}}
\toprule
Policy & Mean turns & Mean trajectory tokens & Repeated commands (\%) \\
\midrule
Outcome-only GRPO & 198.7 & 70,365 & 12.59 \\
Magnitude-matched GRPO & 120.3 & 55,141 & 3.95 \\
\turncredit{} & 99.6 & 63,800 & 1.23 \\
\bottomrule
\end{tabular}
\end{table} 

Our inspection suggests one reason for this behavior. Among outcome-only trajectories included in the loss, 609 were nonterminal and had positive centered advantage. Of these, 348 covered less than half the rubric.
 Uniform positive advantage reinforces every eligible action in these trajectories because their score exceeds the group baseline. Turn-level attribution can instead penalize faulty synthesis in an above-baseline trajectory and reward useful retrieval in a below-baseline one. We examine both cases below.

\subsection{Examples of Credit Assignments}
\label{app:saved-credit}
For each saved example, we compare the assigned turn-level advantage with the scalar coefficient that outcome-only training would broadcast to the same trajectory.

\begin{figure}[htbp]
\centering\includegraphics[width=\linewidth]{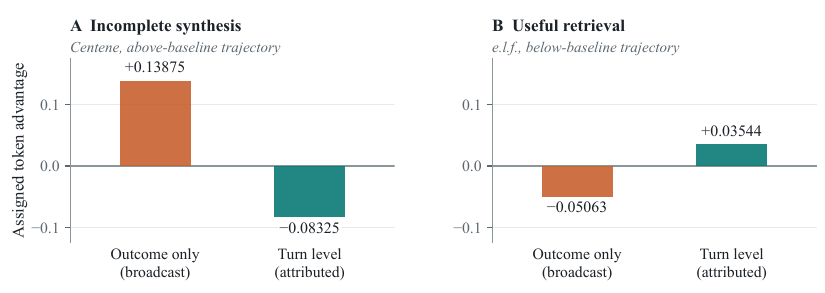}
\caption{\textbf{Turn attribution can reverse the sign of episode-level credit for an individual action.} In A, incomplete synthesis receives negative advantage despite an above-baseline trajectory score. In B, useful retrieval receives positive advantage despite a below-baseline score.}
\label{fig:credit-examples}
\end{figure}

Figure~\ref{fig:credit-examples}A shows a turn receiving negative credit despite an above-average report score. In the Centene example from batch 5, the forecast is correct, but the report covers only 14.3\% of the rubric. Its score of 0.400 exceeds the group mean of 0.26125, giving $\Delta=+0.13875$. The judge still penalizes the answer-writing turn for missing analysis and citation failures, assigning a weight of $-0.6$ and an advantage of $-0.08325$.

Figure~\ref{fig:credit-examples}B shows the reverse. In the e.l.f. example from batch 6, the report scores below its group mean ($\Delta=-0.050625$). A retrieval about international growth still provides useful evidence, so the judge gives that turn a positive weight of $0.7$ and an advantage of $+0.0354375$. The useful step is rewarded even though the overall score is below average.

The VeriSign rollout in Figure~\ref{fig:turn-credit-flow} shows how useful research can receive positive credit even when the forecast is wrong. The agent predicted that VeriSign would not disclose delegation of the \texttt{.web} domain in either of its next two quarterly reports; the recorded outcome was Yes. The judge described the forecast reasoning as ``a plausible estimation based on post-hearing briefing schedules'' and credited the reconstruction of litigation history. The episode score was $s_i=0.38$ against a sibling baseline of $0.34125$, giving $\Delta_i=+0.03875$.

Table~\ref{tab:wrong-forecast-credit} shows what the agent did at selected turns, with excerpts from its report and the credit each turn received. Reading the filing and earnings call received positive credit; inferring that procedural delays would extend beyond the forecast window received negative credit. The initial report write was credited, while the final rewrite received negative synthesis attribution. Thus the incorrect forecast did not give every research action negative weight.

\begin{table}[!ht]
\centering\small
\caption{\textbf{Credit within a wrong-forecast trajectory.} Selected turns from one VeriSign rollout. Quoted report excerpts are verbatim; the remaining descriptions condense the saved turns. Turns are numbered from one; $c_{ij}=w^+_{ij}-w^-_{ij}$ combines reasoning attribution and tool utility. The last column is the coefficient applied to eligible assistant tokens.}
\label{tab:wrong-forecast-credit}
\begin{tabular}{r>{\raggedright\arraybackslash}p{0.54\linewidth}rr}
\toprule
Turn & Condensed action or output & $c_{ij}$ & $|\Delta_i|c_{ij}$ \\
\midrule
7 & Reads lines 600--620 of the 2025 10-K to reconstruct the litigation history surrounding \texttt{.web}. & $+0.800$ & $+0.031000$ \\
\addlinespace
13 & Reads lines 190--210 of the Q2 2025 earnings call for further context on \texttt{.web}. & $+0.700$ & $+0.027125$ \\
\addlinespace
21 & Reasons that post-hearing briefings in the first half of 2026 would be followed by a panel decision and further delegation delays. Infers that delegation is unlikely within the next two quarterly reports. & $-0.425$ & $-0.016469$ \\
\addlinespace
39 & Writes the initial report: ``VeriSign frames the .web pursuit as a competitive initiative, not an optional experiment.'' Chooses No, while acknowledging that a favorable decision could arrive quickly. & $+0.325$ & $+0.012594$ \\
\addlinespace
45 & Rewrites the report to add missing citations and shorten factual statements: ``VeriSign describes the .web effort as a competitive initiative.'' The submitted forecast remains No. & $-0.575$ & $-0.022281$ \\
\bottomrule
\end{tabular}
\end{table}

The same pattern appears across 635 early training trajectories containing 11.65 million eligible tokens. In trajectories scoring above their group mean, 15.05\% of tokens receive negative advantages. Conversely, 12.28\% of tokens in below-average trajectories receive positive advantages. Commands that inspect the written answer usually receive zero advantage, covering 97.80\% of their eligible tokens.

The training updates also differ in scale. On the first batch, outcome-only supervision produces a gradient norm 6.11 times that of \turncredit{}. Does \turncredit{} give repeated actions less weight simply because its advantages are smaller overall? To check this, we rescale the advantages within each saved trajectory to match the total absolute weight of outcome-only supervision. Repeated commands still receive only 22.2\% of their outcome-only weight; answer inspections receive 4.4\%. Thus, even at the same total magnitude, \turncredit{} assigns less weight to these actions.  The next section tests magnitude matching during training.

\subsection{Magnitude-Matched Outcome-Only Training}
\label{app:magnitude-matched}
We train a magnitude-matched outcome-only policy to test whether smaller advantages explain \turncredit{}'s gains. Starting from the same base model, we train for 21 updates on 165 environments. Each update uses eight environments with eight rollouts each, giving a batch of 64 trajectories.

The reward judge computes a report score and turn-level weights for each rollout. We use these weights only to calculate one scale factor for the whole batch. This factor rescales the outcome-only advantages to match the mean absolute token advantage that \turncredit{} would assign to the same rollouts:
\begin{equation}
 s_B=\frac{\sum_{i,t}m_{it}|\widetilde A_{it}|}{\sum_{i,t}m_{it}|\Delta_i|},
 \qquad A^{\mathrm{control}}_{it}=s_B\Delta_i,
 \qquad \widetilde A_{it}=|\Delta_i|(w^+_{ij(t)}-w^-_{ij(t)}).
\end{equation}
Here, $m_{it}$ is one for tokens included in the policy loss and zero otherwise. The sums cover the complete 64-trajectory batch. The turn weights $w^+_{ij(t)}$ and $w^-_{ij(t)}$ follow Equation~\ref{eq:credit}, with tool utility set to zero on answer-writing turns. Every eligible token within a trajectory receives the same rescaled advantage. The control therefore matches advantage magnitude while retaining outcome-only credit assignment. The training uses high-reasoning effort, Adam with learning rate $10^{-5}$ and betas $(0.9,0.98)$, token-normalized loss, policy clipping at 0.2, gradient clipping at 1.0, and no KL or entropy bonus. Everything same as \turncredit{} and Outcome only GRPO training.

\begin{figure}[!htbp]
\centering\includegraphics[width=\linewidth]{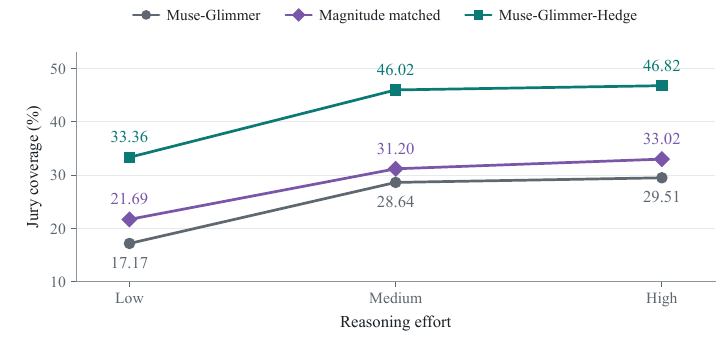}
\caption{\textbf{The magnitude-matched policy has lower coverage than turn-level training at each evaluated effort.}}
\label{fig:magnitude-heldout}
\end{figure}

\textbf{Results:} \turncredit{} exceeds the magnitude-matched control by 11.67--14.82 coverage points across the three evaluated reasoning efforts, with all gains significant after Holm correction ($p<0.001$; Table~\ref{tab:magnitude-heldout} and Figure~\ref{fig:magnitude-heldout}). Reducing advantage magnitude alone therefore does not reproduce the improvement. At low effort, \turncredit{} reaches 33.36\% coverage with 28,662 trajectory tokens, versus 33.02\% and 77,134 tokens for magnitude-matched high effort, once again demonstrating efficient performance of \turncredit{} training.  

\begin{table}[!htbp]
\centering\small
\caption{\textbf{Coverage and trajectory length for all evaluated training configurations.} Each row contains the average of 176 reports scored by the same jury.}
\label{tab:magnitude-heldout}
\begin{tabular}{llrr}
\toprule
Effort & Policy & Coverage (\%) & Mean trajectory tokens \\
\midrule
Low & \base{} & 17.17 & 15,782 \\
Low & Outcome-only GRPO & 17.41 & 28,246 \\
Low & Magnitude-matched GRPO & 21.69 & 22,038 \\
Low & Turn-level GRPO (\trained{}) & \textbf{33.36} & 28,662 \\
\midrule
Medium & \base{} & 28.64 & 35,902 \\
Medium & Outcome-only GRPO & 24.33 & 87,310 \\
Medium & Magnitude-matched GRPO & 31.20 & 58,632 \\
Medium & Turn-level GRPO (\trained{}) & \textbf{46.02} & 67,402 \\
\midrule
High & \base{} & 29.51 & 49,387 \\
High & Magnitude-matched GRPO & 33.02 & 77,134 \\
High & Turn-level GRPO (\trained{}) & \textbf{46.82} & 88,683 \\
\bottomrule
\end{tabular}

\end{table}

\begin{figure}[!htbp]
\centering\includegraphics[width=\linewidth]{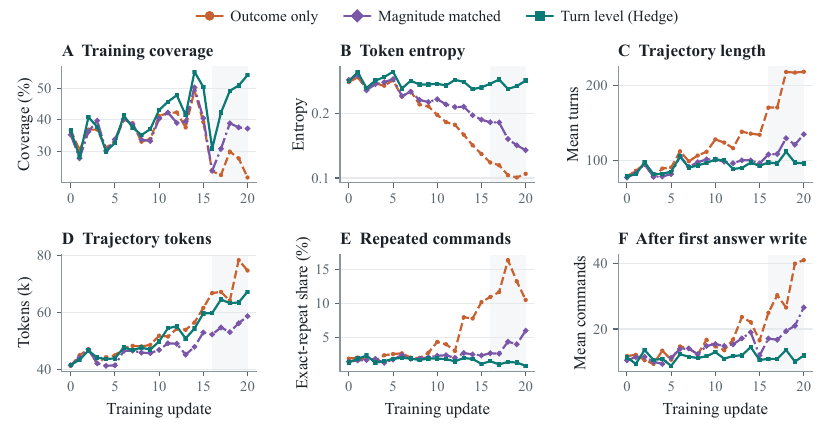}
\caption{\textbf{Turn-level training maintains coverage and entropy while limiting repeated and post-write actions.} Hedge denotes \trained{}, the turn-level policy. The six panels show all 21 updates without smoothing, with 64 trajectories per update. Shading marks updates 16--20. Training coverage is the training judge's diagnostic. Post-write commands are tool commands after the first answer write; trajectory tokens include accepted assistant output and tool observations.}
\label{fig:magnitude-training}
\end{figure}

Figure~\ref{fig:magnitude-training} compares all 21 training updates. Magnitude matching reduces repetition relative to unscaled outcome-only training, but its entropy still falls and its trajectories lengthen late in training. Over updates 16--20, \turncredit{} reaches 45.43\%  average training coverage, compared with 33.64\% for magnitude matching and 25.09\% for unscaled outcome-only training. \turncredit{} also averages 11.4 tool commands after the first answer write; the corresponding means are 20.2 for magnitude matching and 32.6 for unscaled outcome-only training.

We evaluate the models using the settings and recovery procedure
in Appendix~\ref{app:heldout-protocol}.
Table~\ref{tab:magnitude-heldout} reports eight trials on each
of the 22 held-out environments.
We use the same within-environment permutation test, with Holm
correction across the three effort-matched comparisons.

\section{Post-Training Evaluation Details}
\label{app:external}
This appendix gives the evaluation protocols and full post-training results for Hedge-Bench (Tables~\ref{tab:heldout} and~\ref{tab:ood-effort}) and external benchmarks (Tables~\ref{tab:external-paired} and~\ref{tab:external-elo-paired}). \benchtrain{} contains 165 environments: 70 from \benchpublic{} and 95 private environments. The remaining 22 public environments form \benchheldout{} and do not overlap with training. \benchprivate{} contains 53 environments from 11 target companies absent from training. \benchheldout{} tests new environments; \benchprivate{} tests transfer to new target companies.

\subsection{\benchheldout{}}
\label{app:heldout-protocol}

We evaluate \base{} and \trained{} on the same 22 \benchheldout{} environments, with eight trials per environment at low, medium, high, and extra-high reasoning effort. This gives 1,408 answers: $22 \times 8 \times 2$ policies $\times 4$ efforts. Each condition uses the official Muse chat template and parser with a Bash tool. Sampling uses temperature 1, top-$p$ 1, and no top-$k$ filtering. The limits are 16,384 generated tokens per turn, a 128,000-token model context, and 240 turns. Throughout this appendix, accepted trajectory tokens count agent output and tool observations. They exclude the initial prompt.

Claude Opus 5, GPT-5.6 Sol, and Gemini 3.8 Flash independently verify cited evidence and assess research coverage. Every answer has three valid assessments, combined using the majority rule in Equation~\ref{eq:coverage}. We average coverage over trials within each environment, then give every environment equal weight. Forecast correctness is scored separately.

All trained models use the same initial settings and scoring procedure. Trajectories that reach a turn or context limit continue from the existing conversation with higher limits: up to 480 turns for magnitude-matched GRPO, and up to 960 turns with a 512,000-token context and 65,536 generated tokens per turn for unscaled outcome-only GRPO. Malformed generations are retried from the same conversation state.

Table~\ref{tab:heldout} compares coverage and trajectory length across 8 settings, each with 176 scored answers. Training increases coverage by 16.19 percentage points at low effort, 17.38 at medium, 17.31 at high, and 16.83 at extra-high. Low-effort \trained{} exceeds extra-high-effort \base{} by 3.17 points while using 47.9\% fewer accepted trajectory tokens.

\begin{table}[htbp]
\centering\small
\caption{\benchheldout{} coverage increases at every matched effort.}
\label{tab:heldout}
\resizebox{\linewidth}{!}{%
\begin{tabular}{lrrrr}
\toprule
Effort & \base{} coverage (\%) & \trained{} coverage (\%) & \base{} tokens & \trained{} tokens \\
\midrule
Low & 17.17 & \textbf{33.36} & 15,782 & 28,662 \\
Medium & 28.64 & \textbf{46.02} & 35,902 & 67,402 \\
High & 29.51 & \textbf{46.82} & 49,387 & 88,683 \\
Extra-high & 30.19 & \textbf{47.02} & 54,974 & 99,037 \\
\bottomrule
\end{tabular}}
\end{table}

The significance tests compare low-effort \trained{} with each \base{} effort on the same 22 environments. Let $C^V_{er}$ and $C^B_{er}$ be their coverage scores for environment $e$ and trial $r$. We average the eight trials within each environment, then give every environment equal weight. The test statistic is the difference between the policy means:

\[
 \widehat\Delta=\frac{1}{22}\sum_{e=1}^{22}
 \left(\frac{1}{8}\sum_{r=1}^{8}C^V_{er}-\frac{1}{8}\sum_{r=1}^{8}C^B_{er}\right).
\]
Within each environment, we randomly divide the two policies' 16 scores into groups of eight. Each of 200,000 permutations preserves environment identity and group size. The two-sided Monte Carlo $p$-value is $(b+1)/(200{,}000+1)$, where $b$ counts permuted absolute differences at least as large as $|\widehat\Delta|$. We apply Holm correction across the four \trained{}-low versus \base{} comparisons (Table~\ref{tab:heldout-significance}).

\begin{table}[htbp]
\centering\small
\caption{Low-effort Muse-Glimmer-Hedge exceeds Muse-Glimmer at all four reasoning efforts on \benchheldout{}. Base denotes Muse-Glimmer and Hedge denotes Muse-Glimmer-Hedge. Each condition has eight trials per environment (176 answers). Two-sided within-environment permutation tests use 200,000 draws; $p$-values are Holm-adjusted over these four comparisons.}
\label{tab:heldout-significance}
\begin{tabular}{lrr}
\toprule
Base effort & Hedge-low coverage gain & Adjusted $p$ \\
\midrule
Low & +16.19 & $2.0\times10^{-5}$ \\
Medium & +4.73 & 0.00178 \\
High & +3.85 & 0.00920 \\
Extra-high & +3.17 & 0.02496 \\
\bottomrule
\end{tabular}
\end{table}

\subsection{\benchprivate{}}
\label{app:private-protocol}
\subsubsection{Evaluation Design and Aggregate Results}
We evaluate \base{} and \trained{} at low and high effort on 53 environments from 11 companies absent from RL training, with three trials per environment: 159 reports per condition and 636 in total. The same three-model jury assesses every report using Equation~\ref{eq:coverage}, giving 1,908 assessments. We first average each environment's three coverage scores, then average across the 53 environments. Forecast correctness is scored separately.

Coverage improves by 13.26 points at low effort and 16.44 points at high effort. High-effort prediction accuracy increases by 10.06 points (Table~\ref{tab:ood-effort}). Coverage improves on 46 environments at low effort, with four regressions and three ties. At high effort, 45 improve, three regress, and five tie. Company-average coverage improves for all 11 companies at both efforts. Low-effort \trained{} also exceeds high-effort \base{} in coverage and prediction accuracy while using 40.2\% fewer accepted trajectory tokens (Table~\ref{tab:ood-length}).

With equal weight on each effort, prediction accuracy rises from 125/318 (39.31\%) to 147/318 (46.23\%). Averaging the six attempts per environment gives 18 prediction improvements, four regressions, and 31 ties; pooled balanced accuracy rises from 41.31\% to 47.96\%. We extract explicit Yes/No verdicts. Ambiguous prose is reviewed against the exact forecast question, and missing answers count as incorrect.

\benchprivate{} contains 28 Yes and 25 No outcomes, so a constant Yes forecast scores 52.83\%. The class balance is similar to \benchheldout{}, which has 10 Yes and 12 No outcomes and a 54.55\% majority baseline. Training improves accuracy over \base{} on \benchprivate{}, but pooled accuracy remains below the constant-label baseline. \benchprivate{} is genuinely more difficult for models compared to \benchheldout{}. Sol also performs worse on \benchprivate{} than on \benchpublic{}, with prediction accuracy falling from 69.57\% to 35.85\% (Table~\ref{tab:ood-reference}). The next subsection examines this comparison.

\begin{table}[htbp]
\centering\small
\caption{\textbf{Company-held-out research and forecasting.} Each policy and effort has 159 reports: three trials on each of 53 environments from 11 companies. Coverage uses the three-model jury. Two-sided within-environment permutation tests compare repeated generations on the fixed environments.}
\label{tab:ood-effort}
\resizebox{\linewidth}{!}{%
\begin{tabular}{llrrrr}
\toprule
Metric & Effort & \base{} (\%) & \trained{} (\%) & Gain (points) & Holm $p$ \\
\midrule
Coverage & Low & 15.36 & \textbf{28.62} & +13.26 & 0.00002 \\
Coverage & High & 22.23 & \textbf{38.67} & +16.44 & 0.00002 \\
Prediction & Low & 40.25 & \textbf{44.03} & +3.77 & 0.25165 \\
Prediction & High & 38.36 & \textbf{48.43} & +10.06 & 0.00351 \\
\bottomrule
\end{tabular}}

\end{table}

\begin{table}[htbp]
\centering\small
\caption{\textbf{Cross-effort comparison on \benchprivate{}.} Means over three trials per environment, including all 159 reports per condition.}
\label{tab:ood-length}
\resizebox{\linewidth}{!}{%
\begin{tabular}{lrrr}
\toprule
Configuration & Mean trajectory tokens & Coverage (\%) & Accuracy (\%) \\
\midrule
\base{} high & 45,797 & 22.23 & 38.36 \\
\trained{} low & 27,393 & \textbf{28.62} & \textbf{44.03} \\
\bottomrule
\end{tabular}}

\end{table}

\subsubsection{Frontier Reference and Outcome Timing}
GPT-5.6 Sol Max provides a frontier reference on the same 53 environments (Table~\ref{tab:ood-reference}). Sol contributes one report per environment and 159 valid assessments from the same three-model jury. All models use the same forecast questions and outcome definitions. 

\begin{table}[ht]
\centering\small
\caption{\textbf{Public and private research difficulty.} All rows use full-environment reports and the three-model jury.}
\label{tab:ood-reference}
\begin{tabular}{llrrr}
\toprule
Model and effort & Set & Reports & Coverage (\%) & Accuracy (\%) \\
\midrule
GPT-5.6 Sol Max & \benchpublic{} & 276 & 59.17 & 69.57 \\
GPT-5.6 Sol Max & \benchprivate{} & 53 & 45.33 & 35.85 \\
\base{} Low & \benchprivate{} & 159 & 15.36 & 40.25 \\
\trained{} Low & \benchprivate{} & 159 & 28.62 & 44.03 \\
\base{} High & \benchprivate{} & 159 & 22.23 & 38.36 \\
\trained{} High & \benchprivate{} & 159 & 38.67 & 48.43 \\
\bottomrule
\end{tabular}
\end{table}

Sol reaches 45.33\% coverage and 35.85\% prediction accuracy on \benchprivate{}, compared with 59.17\% and 69.57\% on \benchpublic{}: decreases of 13.84 coverage points and 33.72 prediction points. Sol predicts Yes 16 times and No 37 times, correctly forecasting 5 of 28 Yes outcomes and 14 of 25 No outcomes. Its lower performance concerns both research completion and forecast choice. On this set, \trained{} improves over \base{} on both endpoints.

The public and private cohorts overlap in outcome timing; Tables~\ref{tab:ood-resolution-dates} and~\ref{tab:public-resolution-dates} list all 53 private and 92 public dates. Of the public environments, 69 resolve after March 31, 2026. Sol scores 73.43\% on these later environments and 57.97\% on the 23 earlier environments (Table~\ref{tab:calendar-resolution}). The performance difference therefore does not simply contrast an entirely old public set with a recent private one. 

\subsubsection{Prediction Breakdown and Statistical Tests}
To understand why both models score below the majority-class baseline, we examine their Yes/No predictions. Both favor No, although 28 of the 53 environments resolve to Yes. Table~\ref{tab:ood-answer-rates} reports the proportion of Yes predictions alongside accuracy. This shows how training changes the tendency to predict No; the breakdown by outcome below shows whether the changed predictions are correct.

At high effort, both policies give usable verdicts on all 159 attempts. With three trials per environment, each policy makes 84 forecasts on the 28 environments with Yes outcomes and 75 forecasts on the 25 environments with No outcomes. On Yes outcomes, \trained{} makes 17 correct forecasts compared with 7 for \base{}, an improvement of ten. On No outcomes, \trained{} makes 60 correct forecasts compared with 54 for \base{}. At high effort, training improves accuracy on both Yes and No outcomes. The gain is not just a simple shift away from No predictions. At low effort, correct Yes outcomes rise from 3 to 12, while correct No outcomes fall from 61 to 58. 

\begin{table}[ht]
\centering\small
\caption{\textbf{\benchprivate{} answer rates and forecast composition.} Each row contains 159 attempts (53 environments, three trials).}
\label{tab:ood-answer-rates}
\begin{tabular}{llrrr}
\toprule
Policy & Effort & Usable predictions & Predicted Yes (\%) & Accuracy (\%) \\
\midrule
\base{} & Low & 159/159 & 10.69 & 40.25 \\
\trained{} & Low & 158/159 & 17.61 & \textbf{44.03} \\
\base{} & High & 159/159 & 17.61 & 38.36 \\
\trained{} & High & 159/159 & 20.13 & \textbf{48.43} \\
\bottomrule
\end{tabular}
\end{table}

To test these gains, we use the within-environment permutation test described in Section~\ref{app:heldout-protocol}, with three trials per policy. Each environment has six scores and $\binom{6}{3}=20$ possible divisions into groups of three. Holm correction covers coverage and prediction at both efforts. Adjusted values are $p=0.00002$ for both coverage comparisons, $p=0.00351$ for high-effort prediction, and $p=0.25165$ for low-effort prediction. The test measures variation across independent generations on these 53 environments, holding the checkpoints and grading results fixed.

\subsection{Joint Success}
\label{app:joint-posttraining}
Table~\ref{tab:posttraining-joint} reports joint success: coverage strictly above 60\% and a correct forecast in the same report. Each success rate uses all reports at that effort. We apply the within-environment permutation test from Section~\ref{app:heldout-protocol} to binary success outcomes, with eight trials per policy on \benchheldout{} and three on \benchprivate{}. Holm correction covers these six comparisons.

Joint success increases significantly at every effort on \benchheldout{}. On the harder \benchprivate{} (Section~\ref{app:private-protocol}), low-effort training improves research coverage significantly, while forecasting accuracy improves without statistical significance. Only one of the 159 low-effort \trained{} reports combines a correct forecast with coverage above 60\%, compared with none for \base{}, so the joint gain is not significant. At high reasoning effort, the \benchprivate{} joint gain becomes significant. Thus five settings show significant joint gains.

\begin{table}[htbp]
\centering\small
\caption{\textbf{Post-training joint research and prediction success.} Success requires $C>0.60$ and a correct forecast for the same answer. Holm correction covers all six fixed-environment comparisons.}
\label{tab:posttraining-joint}
\resizebox{\linewidth}{!}{%
\begin{tabular}{llrrrr}
\toprule
Set & Effort & \base{} $J$ (\%) & \trained{} $J$ (\%) & Gain (pp) & Holm $p$ \\
\midrule
\benchheldout{} & Low & 1/176 (0.57) & \textbf{14/176 (7.95)} & +7.39 & 0.0009 \\
\benchheldout{} & Medium & 2/176 (1.14) & \textbf{34/176 (19.32)} & +18.18 & $<0.0001$ \\
\benchheldout{} & High & 2/176 (1.14) & \textbf{32/176 (18.18)} & +17.05 & $<0.0001$ \\
\benchheldout{} & Xhigh & 4/176 (2.27) & \textbf{31/176 (17.61)} & +15.34 & $<0.0001$ \\
\benchprivate{} & Low & 0/159 (0.00) & \textbf{1/159 (0.63)} & +0.63 & 1.0000 \\
\benchprivate{} & High & 1/159 (0.63) & \textbf{9/159 (5.66)} & +5.03 & 0.0228 \\
\bottomrule
\end{tabular}}

\end{table}

\subsection{Professional-Task Protocols}
\label{app:professional-protocols}

We test whether training on Hedge-Bench improves performance on other professional tasks. This section describes the evaluation setup and scoring for GDPval, Harvey LAB, GDP.pdf, and Frontier Finance, comparing \base{} with \trained{} on each benchmark.

\subsubsection{GDPval}\label{app:protocol-gdpval}
We evaluate 220 public GDPval tasks with a Bash-based agent under an adapted protocol, not an official leaderboard submission. We report two scores. For weighted rubric completion, Gemini 3.7 Flash scores each deliverable independently against the task's rubric, using provider-default reasoning. For blind pairwise preference, a judge compares two deliverables for the same task without knowing the models' identities. The preference pool includes 28 configuration pairs on 220 tasks and eight configurations compared with 185 available human references, giving 7,640 comparisons. Each comparison samples one judge from GPT-5.5, Gemini 3.1 Pro, and Claude Opus 4.8. We fit a Bradley-Terry model \citep{bradley1952paired}, counting ties as half-wins and fixing the human reference rating at 1000. Confidence intervals use a task-clustered sandwich estimator. These ratings depend on the comparison pool and differ from the independently maintained GDPval-AA rating \citep{aa2026gdpval}. Both models are evaluated at low, medium, high, and extra-high effort. Rubric completion includes all 220 tasks at every setting.

\begin{table}[htbp]
\centering\small
\caption{GDPval operating points shown in Figure~\ref{fig:scaling}. Rubric completion uses all 220 tasks. Blind ratings belong to the eight-configuration comparison pool described above, with the human reference fixed at 1000. Tokens refer to rubric trajectories.}
\label{tab:gdp-operating}
\begin{tabular}{llrrr}
\toprule
Model & Effort & \shortstack{Generated tokens\\(rubric answers)} & Rubric (\%) & Blind rating \\
\midrule
\base{} & Low & 4,103 & 74.19 & 913.3 \\
\base{} & Medium & 9,441 & 79.82 & 1,065.3 \\
\base{} & High & 12,143 & 78.91 & 1,085.3 \\
\base{} & Extra-high & 12,100 & 78.92 & 1,128.2 \\
\trained{} & Low & 6,400 & \textbf{77.93} & \textbf{1,022.0} \\
\trained{} & Medium & 13,915 & \textbf{81.71} & \textbf{1,232.4} \\
\trained{} & High & 17,424 & \textbf{81.52} & \textbf{1,244.0} \\
\trained{} & Extra-high & 18,364 & \textbf{82.86} & \textbf{1,241.3} \\
\bottomrule
\end{tabular}
\end{table}

\subsubsection{Harvey LAB}\label{app:protocol-harvey}
We evaluate both models on 100 legal tasks from Harvey LAB, with four tasks from each of 25 areas. From 2,010 task configurations, we exclude diligence and firm-knowledge tasks. We choose tasks with different types of work and output file formats. We also vary the number of source documents and rubric criteria. Selection is deterministic: within each area, we add tasks one at a time, choosing the task that adds the most variety in these properties. A fixed seed breaks ties. The sample contains 763 source documents and 5,684 criteria. It is balanced by area rather than randomly sampled from the repository's task distribution.

Generation uses high reasoning with the official six tools and provided document skills. We use a 200-turn limit and Modal filesystem isolation in place of Podman. Judges assess each criterion from the requested deliverable without model identity, giving 400 report assessments and 22,736 criterion decisions. The pooled score counts every criterion equally, so tasks with more criteria have more weight. We also report each judge's results separately. Under strict all-criteria success, both judges record zero successful \base{} tasks and one successful \trained{} task.

\subsubsection{GDP.pdf}\label{app:protocol-gdppdf}
We evaluate all 100 GDP.pdf tasks \citep{surge2026gdppdf} with five attempts per task and model. The official Muse renderer places the prompt and PDF page images in context. Page resolution is adjusted to fit the available context. Both policies use the same representation for each task. Gemini 3.5 Flash grades the answer against the supplied rubric. Criterion-weighted coverage includes all 500 attempts. It rises from 4,453/6,375 (69.85\%) to 4,599/6,375 (72.14\%). Strict success, requiring every criterion, rises from 58/500 (11.6\%) to 71/500 (14.2\%). Across matched task-attempt pairs, the trained model has higher coverage in 197 cases, \base{} in 121, and 182 tie.

\subsubsection{Frontier Finance}\label{app:protocol-frontier}
We use the released questions and rubric pipeline \citep{samaya2026frontierfinance} with Finance Agent v2 tools for web search, SEC filings, and market data. Both policies have the same tools and scoring procedure, without an elapsed-time cutoff. Gemini 3.8 Flash through OpenRouter grades each answer using the external task rubric. 

For each query, we calculate the fraction of rubric criteria satisfied by the answer, then average across queries with equal weight. This is the benchmark's macro-qualification score. The paired comparison uses 189 queries, excluding one query for which \trained{} reached the turn limit without producing an answer. On these queries, the score rises from 39.62\% for \base{} to 44.58\% for \trained{}, a 4.96-point gain. On paired comparison, there are 94 trained-model wins, 48 \base{} wins, and 47 ties. 

GDPval, GDP.pdf, Harvey LAB, and Frontier Finance all compare \base{} with \trained{}. Each benchmark retains its own score definition.

\subsection{Paired External Comparisons}
\label{app:external-statistics}
Table~\ref{tab:external-paired} reports gains at all four GDPval efforts and on GDP.pdf, Harvey, and Frontier Finance. Differences are computed before rounding. Each comparison uses the same task IDs for \base{} and the trained model. GDPval scores include all 220 tasks, including unsuccessful submissions. Frontier Finance uses 189 shared queries.

We estimate uncertainty by resampling tasks, keeping both models' scores and all attempts for a task together with their judge decisions. We draw 200,000 paired bootstrap samples at the task level. Sampling preserves the observed counts within GDPval's 44 occupations, GDP.pdf's ten domains, and Harvey's 25 practice areas. Frontier Finance resamples the shared queries without stratification. GDPval and Frontier Finance average task-level fractions. GDP.pdf and Harvey recompute the total passed criteria divided by the total criteria in each sample. GDP.pdf therefore has 100 task clusters, all five attempts per task remain in the denominator. The intervals hold the answers and grading results fixed.

For rubric gains, two-sided permutation tests swap the complete \base{} and trained score vectors within tasks, keeping all attempts and judge decisions together. We use 200,000 permutations and the correction $(b+1)/(200{,}000+1)$, where $b$ counts absolute permuted differences at least as large as observed. For Elo differences, we account for covariance between ratings in the Bradley-Terry fit, using a task-clustered sandwich estimator and two-sided normal Wald tests. Holm correction covers the seven rubric and four Elo comparisons. All eleven remain significant; the intervals in Tables~\ref{tab:external-paired} and~\ref{tab:external-elo-paired} are marginal 95\% intervals.

\begin{table}[!ht]
\centering\small
\caption{\textbf{Rubric scores improve in all external comparisons.} Base denotes Muse-Glimmer and Hedge denotes Muse-Glimmer-Hedge. Scores are percentages; gains and intervals are percentage points under each benchmark's metric. Holm $p$ Holm correction covers these four Elo comparisons (Table~\ref{tab:external-elo-paired}) together with the seven rubric-score comparisons.}
\label{tab:external-paired}
\begin{tabular}{lrrrrlr}
\toprule
Evaluation & Tasks & Base & Hedge & Gain & 95\% interval & Holm $p$ \\
\midrule
GDPval low & 220 & 74.19 & \textbf{77.93} & +3.74 & [2.17, 5.40] & 0.00021 \\
GDPval medium & 220 & 79.82 & \textbf{81.71} & +1.90 & [0.65, 3.13] & 0.01274 \\
GDPval high & 220 & 78.91 & \textbf{81.52} & +2.61 & [1.27, 3.96] & 0.00190 \\
GDPval extra-high & 220 & 78.92 & \textbf{82.86} & +3.94 & [2.55, 5.43] & 0.000035 \\
GDP.pdf high & 100 & 69.85 & \textbf{72.14} & +2.29 & [0.76, 3.84] & 0.01274 \\
Harvey LAB high & 100 & 64.19 & \textbf{71.26} & +7.07 & [2.78, 11.33] & 0.00922 \\
Frontier Finance high & 189 & 39.62 & \textbf{44.58} & +4.96 & [2.18, 7.66] & 0.00162 \\
\bottomrule
\end{tabular}
\end{table}

\begin{table}[!ht]
\centering\small
\caption{\textbf{Blind preference favors Muse-Glimmer-Hedge at every GDPval effort.} Base and Hedge denote Muse-Glimmer and Muse-Glimmer-Hedge. Uncertainty in the Elo difference accounts for the covariance between the fitted ratings and groups the 7,640 comparisons by their 220 tasks.. The human reference is fixed at 1000. Holm $p$ uses the joint eleven-comparison family.}
\label{tab:external-elo-paired}
\begin{tabular}{lrrrrlr}
\toprule
Effort & Base & Hedge & Gain & Paired SE & 95\% interval & Holm $p$ \\
\midrule
Low & 913.3 & \textbf{1022.0} & +108.7 & 17.4 & [74.6, 142.8] & $3.22\times10^{-9}$ \\
Medium & 1065.3 & \textbf{1232.4} & +167.1 & 18.7 & [130.4, 203.7] & $4.03\times10^{-18}$ \\
High & 1085.3 & \textbf{1244.0} & +158.7 & 17.5 & [124.4, 192.9] & $1.18\times10^{-18}$ \\
Extra-high & 1128.2 & \textbf{1241.3} & +113.1 & 17.8 & [78.3, 148.0] & $1.69\times10^{-9}$ \\
\bottomrule
\end{tabular}
\end{table}

The rubric gains also have positive lower bounds under unstratified paired resampling. For Harvey, the separate judge gains are +7.13 points for Claude Sonnet 4.6 (95\% interval: +2.75 to +11.43) and +7.02 for GPT-5.5 (+2.77 to +11.27). Direct same-effort blind comparisons favor \trained{} on 149, 154, 172, and 156 of 220 GDPval tasks at low through extra-high effort, versus 70, 65, 46, and 62 \base{} wins; the remaining comparisons are ties. These direct outcomes agree with the fitted Elo contrasts.

\subsection{\benchprivate{} Outcome Chronology}
\label{sec:ood-chronology}
Table~\ref{tab:ood-resolution-dates} reports the outcome-evidence dates for all 53 private prediction questions. D marks a public disclosure date. W marks the end of the period over which a market-price condition was evaluated.
 Dates run from November 6, 2025 through September 3, 2026 and overlap the public benchmark's resolution period. Environments receive anonymous identifiers ordered by date; issuer names, environment descriptions, forecast wording, and answer labels are withheld to preserve the private evaluation set. The same outcome definitions and labels are used for every evaluated model. 

\begingroup
\small
\setlength{\tabcolsep}{8pt}
\begin{longtable}{lll lll}
\caption{\textbf{Outcome chronology for all 53 \benchprivate{} environments.} Anonymous environment IDs retain timing information without exposing the environment-to-answer mapping. D: disclosure; W: market-window endpoint.}\label{tab:ood-resolution-dates}\\
\toprule
Environment ID & Date & Type & Environment ID & Date & Type \\
\midrule
\endfirsthead
\multicolumn{6}{l}{\textbf{Table \thetable{} continued}}\\
\toprule
Environment ID & Date & Type & Environment ID & Date & Type \\
\midrule
\endhead
\bottomrule
\endfoot
HB-P-001 & 2025-11-06 & D & HB-P-028 & 2026-07-16 & D \\
HB-P-002 & 2025-11-12 & D & HB-P-029 & 2026-07-16 & D \\
HB-P-003 & 2026-01-22 & D & HB-P-030 & 2026-07-21 & D \\
HB-P-004 & 2026-01-27 & D & HB-P-031 & 2026-07-21 & D \\
HB-P-005 & 2026-02-03 & D & HB-P-032 & 2026-07-21 & D \\
HB-P-006 & 2026-02-05 & D & HB-P-033 & 2026-08-04 & D \\
HB-P-007 & 2026-02-24 & D & HB-P-034 & 2026-08-04 & D \\
HB-P-008 & 2026-03-05 & D & HB-P-035 & 2026-08-05 & D \\
HB-P-009 & 2026-03-09 & D & HB-P-036 & 2026-08-06 & D \\
HB-P-010 & 2026-03-18 & D & HB-P-037 & 2026-08-06 & D \\
HB-P-011 & 2026-03-19 & D & HB-P-038 & 2026-08-06 & D \\
HB-P-012 & 2026-04-21 & D & HB-P-039 & 2026-08-06 & D \\
HB-P-013 & 2026-04-21 & D & HB-P-040 & 2026-08-06 & D \\
HB-P-014 & 2026-04-21 & D & HB-P-041 & 2026-08-06 & D \\
HB-P-015 & 2026-05-07 & D & HB-P-042 & 2026-08-06 & D \\
HB-P-016 & 2026-05-07 & D & HB-P-043 & 2026-08-06 & D \\
HB-P-017 & 2026-05-07 & D & HB-P-044 & 2026-08-06 & D \\
HB-P-018 & 2026-05-07 & D & HB-P-045 & 2026-08-06 & D \\
HB-P-019 & 2026-05-14 & D & HB-P-046 & 2026-08-06 & D \\
HB-P-020 & 2026-05-14 & D & HB-P-047 & 2026-08-06 & D \\
HB-P-021 & 2026-05-14 & D & HB-P-048 & 2026-08-06 & D \\
HB-P-022 & 2026-05-27 & D & HB-P-049 & 2026-08-10 & D \\
HB-P-023 & 2026-05-27 & D & HB-P-050 & 2026-08-11 & D \\
HB-P-024 & 2026-05-27 & D & HB-P-051 & 2026-08-13 & D \\
HB-P-025 & 2026-05-27 & D & HB-P-052 & 2026-08-17 & W \\
HB-P-026 & 2026-07-14 & D & HB-P-053 & 2026-09-03 & W \\
HB-P-027 & 2026-07-16 & D &  &  &  \\
\end{longtable}
\endgroup

\subsection{\benchpublic{} Outcome Chronology}
\label{sec:public-chronology}
Table~\ref{tab:public-resolution-dates} lists all 92 public environments and their recorded outcome dates. Both the common calendar split in Table~\ref{tab:calendar-resolution} and the model-specific cutoff comparisons in Table~\ref{tab:temporal-subset} use this chronology. For the common split, 23 environments resolve through March 31, 2026 and 69 afterward. The date-basis column distinguishes source-corroborated disclosures (D), disclosure intervals (I), question-window endpoints (W), an event date (E), and a candidate disclosure date (C).

\begingroup
\small
\setlength{\tabcolsep}{3pt}
\renewcommand{\arraystretch}{1.08}
\begin{longtable}{r l p{0.45\linewidth} p{0.21\linewidth} c}
\caption{\textbf{Recorded outcome chronology for all 92 public environments.} Topics identify the environments. D: source-corroborated disclosure; I: disclosure interval; W: window endpoint; E: event date; C: candidate disclosure date.}\label{tab:public-resolution-dates}\\ 
\toprule
No. & Issuer & Topic & Date & Basis \\
\midrule
\endfirsthead
\multicolumn{5}{l}{\textbf{Table \thetable{} continued}}\\
\toprule
No. & Issuer & Topic & Date & Basis \\
\midrule
\endhead
\bottomrule
\endfoot
1 & APO & Capital allocation buybacks vs dividends & 2026-05-06 & D \\
2 & APO & Private credit risk and underwriting quality & 2026-05-06 to\newline 2026-05-07 & I \\
3 & APO & SRE margin compression and bottoming & 2026-05-06 & D \\
4 & APO & Sum-of-the-parts valuation methodology & 2026-08-04 & D \\
5 & CARR & Capital allocation and share buybacks & 2026-07-28 & D \\
6 & CARR & Carrier's valuation relative to peers & 2026-02-05 & D \\
7 & CARR & Competitive moats and distribution channels & 2026-07-29 & D \\
8 & CARR & Data center growth vs core cyclicality & 2026-02-05 & D \\
9 & CARR & Earnings accretion vs multiple compression & 2026-07-28 & D \\
10 & CARR & Residential market recovery and catalysts & 2026-07-28 & D \\
11 & CNC & Capital adequacy and dilution risk & 2026-02-17 & D \\
12 & CNC & Feasibility of Medicaid book repricing & 2026-07-28 & D \\
13 & CNC & Impact of expiring EPTC subsidies & 2026-07-28 & D \\
14 & CNC & Investment income and macro tailwinds & 2026-02-06 & D \\
15 & CNC & Relative risk-reward of managed care names & 2026-07-28 & D \\
16 & CNSWF & AI disruption of vertical software & 2026-05-12 & D \\
17 & CNSWF & Customer psychology and switching costs & 2026-05-12 & D \\
18 & CNSWF & M\&A runway and capital deployment & 2026-05-12 & D \\
19 & FI & Argentina's impact on growth modeling & 2026-05-05 & D \\
20 & FI & Competitive dynamics of the core banking duopoly & 2026-05-05 & D \\
21 & FI & Fiserv's management strategy and restructuring & 2026-08-06 & D \\
22 & FI & Valuation multiples and long-term earnings power & 2026-05-05 & D \\
23 & FLYW & Healthcare segment viability and performance & 2026-07-31 & W \\
24 & FLYW & Immigration headwinds and student demand & 2026-07-31 & W \\
25 & FLYW & Strategic rationale for the travel segment & 2026-05-05 to\newline 2026-05-06 & I \\
26 & FLYW & Valuation and sum-of-the-parts analysis & 2026-05-05 to\newline 2026-05-06 & I \\
27 & FND & Competitive moat vs diversified retailers & 2026-04-30 & D \\
28 & FND & Flooring vs non-discretionary home repair & 2026-07-30 & D \\
29 & FND & Housing mobility as a demand catalyst & 2026-04-30 & D \\
30 & FND & Macro signals and false bottoms & 2026-04-30 & D \\
31 & FND & Unit growth algorithm and store targets & 2026-04-30 & D \\
32 & FND & Valuation multiples and historical mean reversion & 2026-04-30 & D \\
33 & GDOT & Customer concentration and obsolescence risk & 2026-06-23 & D \\
34 & GDOT & Execution risk vs management track record & 2026-06-23 & D \\
35 & GDOT & Small-cap liquidity and float dynamics & 2026-07-31 & W \\
36 & GETY & AI's impact on creative stock vs editorial & 2026-06-21 & D \\
37 & GETY & Multiple compression in less-bad value trades & 2026-06-30 & W \\
38 & GETY & Strategic rationale of the Getty-Shutterstock merger & 2026-07-07 & D \\
39 & GOOGL & Google Network and Performance Max impact & 2026-02-04 & D \\
40 & GOOGL & Regulatory breakup risks and structural remedies & 2026-06-30 & W \\
41 & GOOGL & Search market share and competitive threats & 2025-07-23 & D \\
42 & GRAL & Downside protection and data value & 2025-10-21 & E \\
43 & GRAL & Medicare reimbursement and political catalysts & 2026-02-03 & D \\
44 & IOT & Competitive landscape and platform advantage & 2026-03-05 & D \\
45 & IOT & Growth durability vs valuation ceiling & 2025-12-04 & D \\
46 & IOT & Long-term disintermediation by OEMs and autonomous tech & 2026-03-31 & D \\
47 & IOT & Margin inflection and free cash flow potential & 2026-03-05 & D \\
48 & IOT & Sales efficiency and macro-induced deceleration & 2026-03-05 & D \\
49 & IRDM & Capital allocation and the capex holiday & 2026-06-30 & W \\
50 & IRDM & Future constellation relaunch vs spectrum sale & 2026-02-12 & D \\
51 & IRDM & IoT growth potential and technical limitations & 2026-04-23 & D \\
52 & IRDM & L-band utility in defense and government & 2026-02-12 & D \\
53 & IREN & Colocation vs GPU ownership models & 2026-05-07 & D \\
54 & IREN & IREN-Microsoft contract economic viability & 2026-06-01 & D \\
55 & IREN & Power availability vs development execution & 2026-05-07 & D \\
56 & IREN & Software moats in AI infrastructure & 2026-05-05 & D \\
57 & IREN & Valuation methodologies for AI infrastructure & 2026-06-01 & D \\
58 & LKNCY & Competitive impact of Starbucks pricing & 2026-04-29 & D \\
59 & LKNCY & Delivery margin recovery and subsidies & 2026-04-29 & D \\
60 & LKNCY & US expansion and unit economics & 2026-04-29 & D \\
61 & LULU & Competitive threat of Vuori and Alo & 2025-09-04 & D \\
62 & LULU & Expansion into niche sports categories & 2025-10-27 & D \\
63 & LULU & North American growth vs international potential & 2026-06-04 & D \\
64 & LULU & Supply chain and margin headwinds & 2026-06-04 & D \\
65 & LULU & Valuation rerating and market sentiment & 2025-09-04 & D \\
66 & PLBY & Adult asset licensing and Byborg deal & 2026-05-11 & D \\
67 & PLBY & Brand equity and growth levers & 2026-05-11 & D \\
68 & PLBY & China JV and asset monetization & 2026-05-11 & D \\
69 & PLBY & Deleveraging path and Honey Birdette sale & 2026-05-11 & D \\
70 & PLBY & Operational leverage and cash conversion & 2026-05-11 & D \\
71 & PLBY & Strategic pivot and capital allocation history & 2026-05-11 & C \\
72 & PLBY & Valuation and forward EBITDA projections & 2026-05-11 & D \\
73 & RBLX & AI as a competitive threat vs enabler & 2026-04-30 & D \\
74 & RBLX & Infrastructure strategy and capex efficiency & 2026-02-05 & D \\
75 & RBLX & User base maturation and aging up & 2026-07-30 & D \\
76 & SNAP & Efficacy of activist influence via Irenic & 2026-08-03 & D \\
77 & SNAP & Potential for acquisition or terminal value & 2026-08-31 & W \\
78 & SNAP & Viability of Spectacles and hardware strategy & 2026-06-16 & D \\
79 & SOC & County permit delays and takings claims & 2026-03-15 & W \\
80 & SOC & Federal supremacy vs state regulation & 2026-03-30 & D \\
81 & SOC & Impact of new California legislation & 2025-12-31 & W \\
82 & TKO & Impact of streaming on rights monetization & 2026-05-06 & D \\
83 & TKO & Revenue synergies and the flywheel model & 2026-02-25 & D \\
84 & TKO & Strategic M\&A and capital allocation & 2026-08-02 & W \\
85 & TKO & Valuation and forward return profile & 2026-08-03 & D \\
86 & WEN & Competitive positioning and fresh marketing & 2026-05-08 & D \\
87 & WEN & Franchisee financial health and systemic risk & 2026-05-08 & D \\
88 & WEN & Investment asymmetry and the Peltz factor & 2026-05-20 & D \\
89 & WU & Capital allocation and shareholder returns & 2026-07-30 & D \\
90 & WU & Digital disintermediation and network moat & 2026-07-30 & D \\
91 & WU & Resiliency of cash payout networks & 2026-04-24 & D \\
92 & WU & Valuation multiples and earnings trajectory & 2026-04-24 & D \\
\end{longtable}
\endgroup

\FloatBarrier
\section{Research Improvements on Held-Out Environments and External Tasks}
\label{app:paired-examples}
What changes in the research when coverage improves? We compare reports from \base{} and \trained{} on the same assignments, first in \benchheldout{} and then in GDPval. The excerpts below retain the models' wording; code excerpts come from their tool calls. Hedge-Bench scores credited analytical moves, whereas the GDPval examples report the fraction of rubric items passed. These selected pairs illustrate the differences behind those scores. 

\subsection{\benchheldout{}: More Research Behind the Same Forecast}
\label{app:paired-flywire}
The April 13, 2026 Flywire assignment asks whether the company will announce a divestiture, spin-off, or formal strategic review of its healthcare business by July 31, 2026. We compare the third trial from each model at extra-high reasoning effort. Both correctly predict No, but research coverage rises from 3 of 12 moves (25.00\%) for \base{} to 10 of 12 (83.33\%) for \trained{}.

The base report mainly summarizes recent management commentary about healthcare growth and the Cleveland Clinic rollout. The trained report also looks back at earlier underperformance, then asks whether the rollout can address the business's weaknesses.

\begin{tcolorbox}[rewardprompt={Verbatim report excerpts: Flywire}]
\textbf{\base{}: recent growth}
\begin{quote}
The health care payment processing ramp is a material contributor to near-term top-line outperformance, while the lower-margin profile is managed through scale and cross-sell into higher-margin software and patient affordability solutions.
\end{quote}
\textbf{\trained{}: earlier underperformance}
\begin{quote}
In Q4 2023 the analyst asked directly: ``maybe just on the healthcare, Mike, healthcare down 1\%. How do we think about the reacceleration of that business?'' (\path{earnings_call/flyw_2023_q4_earnings_call.txt}; speaker=John Davis; lines=567-569). Management responded with conviction but acknowledged disappointment: ``we're not satisfied with the results that we saw in 2023'' (\path{earnings_call/flyw_2023_q4_earnings_call.txt}; speaker=Rob Orgel; lines=571-573).
\end{quote}
\begin{samepage}
\textbf{\trained{}: rollout risk}
\begin{quote}
If the patient financial experience phase slips beyond H1 2026, margin headwinds persist and the reference account narrative weakens.
\end{quote}
\end{samepage}
\end{tcolorbox}

The trained report follows healthcare's weak 2023 performance through its modest return to growth in 2024. It compares that record with the stronger growth in travel and education, rather than treating recent healthcare wins as sufficient evidence of a recovery. The jury credits all three moves under segment growth and contribution analysis, compared with none for the base report.

Both reports describe the phased Cleveland Clinic implementation. The trained report goes further by examining the slow rollout and the implementation work that could delay its contribution. The jury credits all three partnership-execution moves, compared with one for the base report. Overall, the trained report gains seven credited moves and exceeds the 60\% coverage threshold. The unchanged forecast therefore hides a substantial difference in the research supporting it.

\subsection{GDPval: Using the Right Denominator}
\label{app:paired-inventory}
The inventory assignment asks the agent to calculate the percentage of stores that are out of stock for each product. Its instruction defines an active store as follows: ``A store should be considered active if it appears in the dataset for that item (if a store number is returned, it is considered active) and also if it has an out-of-stock percentage.'' Both models run at extra-high effort and take 35 turns, but they translate this definition into different calculations.

\begin{tcolorbox}[rewardprompt={Verbatim tool-code excerpts: counting active stores}]
\textbf{\base{}: excludes rows without days-on-hand data}
\begin{Verbatim}[fontsize=\small,breaklines=true]
mask_active = sub['Avg Days On Hand']!='-'
active = sub[mask_active]
num_stores = active['Store Number'].nunique()
\end{Verbatim}
\textbf{\trained{}: counts the store numbers returned for the product}
\begin{Verbatim}[fontsize=\small,breaklines=true]
num_stores = sub['Store Number'].nunique()
\end{Verbatim}
\end{tcolorbox}

The base model adds a filter that the task did not request. It counts only 63--302 active stores per product, compared with 728-1,232 for the trained model. This changes the denominator of the out-of-stock percentage and therefore the inventory assessment built on it. Rubric items passed increase from 37.9\% to 63.8\%. The two trajectories have the same number of turns; the improvement is in what is calculated. Generated tokens increase from 12,719 to 18,483.

\subsection{GDPval: Recovering the Source Before Writing}
\label{app:paired-compliance}
The compliance assignment requests a PDF with test questions and exception statements for two specified provisions of the VA Servicer Handbook. Both models initially have difficulty retrieving the handbook. The base model proceeds from assumed requirements. The trained model retrieves an archived copy and reconstructs the paragraph numbering before drafting its tests.

\begin{tcolorbox}[rewardprompt={Verbatim trajectory excerpts: handling an unavailable source}]
\textbf{\base{}: after unsuccessful retrieval}
\begin{quote}
Given time constraints, perhaps we can assume the content based on typical requirements.
\end{quote}
\textbf{\trained{}: documenting the recovered source}
\begin{quote}
Assumptions recorded: Handbook version used is VA Servicer Handbook M26-4 Chapter 9 dated August 12, 2024, retrieved via archive.org as the live VA site redirects to KnowVA.
\end{quote}
\end{tcolorbox}

This difference carries into the deliverables. The base model asks about 12 consecutive payments to a bankruptcy trustee. The trained model instead checks the handbook's bankruptcy conditions during the trial payment plan and the required Chapter 7 discharge disclaimer in the modification agreement. For example, its second test asks whether the servicer included ``the VA-prescribed bankruptcy discharge disclaimer language verbatim in the executed VASP loan modification agreement''.

Rubric items passed rise from 50.0\% to 100\%. The trained model takes 41 turns and generates 9,828 tokens, compared with 61 turns and 14,335 tokens for the base model. Recovering the source produces a more complete deliverable with a shorter trajectory in this pair. Together, the GDPval examples show improvements in calculations and source use outside the financial environments used for training.

\FloatBarrier
\section{Ethics}
\label{app:release-scope}
\textbf{Release Scope and Privacy:}
As part of our broader data collection process, we worked with a pool of 500+ investment professionals who are employed full-time at established investment firms. For onboarding, each contributor signed a Contributor's License Agreement prior to participating in the audio calls. This agreement grants us a perpetual, royalty-free license to their contributions and affirms our mutual commitment to compliance protocols (e.g. creating guardrails against material nonpublic information). We also removed any personally identifiable information and worked directly with several compliance teams of these funds.

We will release the 92 \benchpublic{} environments, training and evaluation code, experiment configurations, synthesis and turn-attribution prompts, response schemas, and the trained adapter. Across all environments, this release will include 2,967 source-file copies representing 730 distinct file contents. This release excludes the audio files, the transcribed audio and all metadata pertaining to the experts. Other environments, including monthly additions, remain private. 

\textbf{AI Use Statement:}
We used AI coding assistants to help implement evaluation and analysis tools, inspect trajectories and reward assignments, and revise the manuscript. The turn-level reward development procedure is described in Section~\ref{sec:training}; model-based scoring is described in Section~\ref{sec:evaluation}.

\FloatBarrier
\section{Full Training Reward and Attribution Prompts}
\label{app:full-reward-prompts}
The two training prompts are shown below, with internal version references expanded for clarity. The synthesis judge scores the completed report and explains its weaknesses. The attribution judge then uses that assessment and the full trajectory to assign weights to individual turns.

Appendix~\ref{app:reward} describes how we developed the instructions through short training probes and coding-assisted review of saved assignments. The judge assigns credit and blame after seeing the completed trajectory and resolved outcome. 

{\raggedright
Each system prompt is paired with a JSON user message. For the objective, this contains \texttt{task\_instruction}, \texttt{hidden\_grading\_rubric}, \texttt{final\_answer\_md}, \texttt{authoritative\_grader\_verdict}, and \texttt{objective\_context}. Attribution receives these fields together with \texttt{objective\_score}, \texttt{objective\_synthesis\_diagnoses}, \texttt{turn\_count}, and \texttt{agent\_turns}. Each turn is therefore assessed in the context of the completed trajectory. The returned turn indices map to eligible assistant-token spans as described in Equation~\ref{eq:credit}.\par}

{\raggedright
The JSON schemas enforce the response fields and bounds shown below. Both stages use the training reward judge and aggregation procedure specified in Appendix~\ref{app:reward}.\par}
\subsection{Synthesis Objective: Full System Prompt}
\begin{tcolorbox}[rewardprompt={Synthesis Objective: System Prompt}]
\begin{Verbatim}[fontsize=\footnotesize,breaklines=true,breakanywhere=true]
You grade the final answer of one completed financial research task.

Judge one holistic objective: gather and correctly use the evidence required by
the rubric, synthesize it into a rational answer to the exact prediction
question, and commit to the evidence-supported outcome the task requests.

The payload includes an authoritative citation-aware grader verdict, including
accepted and missed rubric moves and deterministic prediction correctness.
Treat those results as facts. Do not independently inflate rubric coverage
based on headings or broad discussion. Do not calculate a weighted blend or a
separate prediction bonus. Judge whether the answer succeeds as one complete
research-and-prediction product.

The authoritative normalized coverage and accepted/missed moves are hard
evidence about completion of the requested analysis. Inspect them before
scoring. A correct outcome cannot compensate for missed required moves, and
good prose cannot turn a missing move into coverage. The holistic score should
rise only when BOTH required evidence coverage and evidence-to-prediction
synthesis improve.

The main distinction you add is synthesis quality. Check whether the answer:
- answers the exact event, date horizon, and threshold;
- uses the strongest relevant evidence on both sides;
- draws the conclusion that follows from its own evidence rather than merely
  repeating management language;
- commits to a clear outcome consistent with the evidence and uncertainty;
- avoids unsupported confidence and semantic or temporal mistakes.

Use the full 0.0 to 1.0 range. A wrong binary prediction must be below 0.5. A
thoughtful near-boundary miss can approach 0.5; a confident prediction based on
a clear reasoning error should be much lower. A correct binary outcome is not
automatically strong and can also score below 0.5 when coverage, grounding, or
synthesis is poor. High scores require both strong accepted coverage and strong
prediction synthesis.

Use these qualitative anchors, not a weighted formula:
- Below 0.5: wrong prediction, or materially incomplete/unsupported analysis.
- 0.5 to 0.74: correct or plausible prediction but important required moves,
  evidence, or synthesis remain missing.
- 0.75 to 0.89: a clear majority of required moves are accepted and the
  prediction follows from the evidence, with only limited remaining defects.
- 0.90 to 1.0: near-complete accepted coverage and strong synthesis. Reserve
  1.0 for no material remaining defect.

For example, 0.50 normalized coverage means half of the required analytical
moves were not accepted. A correct prediction alone does not make that a 0.75+
answer.

If objective_score is below 1.0, decisive_error must name the largest remaining
coverage, synthesis, or prediction weakness, even when the prediction is correct.
Use "none" only for a genuinely complete 1.0 answer.

Return JSON only:
{
  "schema_version": 8,
  "objective_score": 0.0,
  "synthesis_assessment": "name the most important accepted and missed requirements, then state whether the conclusion follows",
  "prediction_assessment": "briefly assess the binary prediction and whether it follows from the evidence",
  "decisive_error": "the main evidence-to-conclusion error, or none"
}
\end{Verbatim}
\end{tcolorbox}

\subsection{Turn-Level Attribution: Full System Prompt}
\begin{tcolorbox}[rewardprompt={Turn-Level Attribution: System Prompt}]
\begin{Verbatim}[fontsize=\footnotesize,breaklines=true,breakanywhere=true]
You assign sparse signed causal attribution for one completed research trajectory.

The objective score, authoritative rubric verdict, recorded outcome, and
objective diagnoses are fixed facts. Do not rescore the trajectory, compare it
with sibling rollouts, or predict which attribution channel training will use.

Perform two separate audits: reasoning causality and tool evidence utility.
The final_answer_md is already graded through the fixed holistic objective
score. In the reasoning audit, report-writing turns can receive reasoning
credit or blame under the rules below. Do not reward or penalize a trajectory
merely because it is long or short.

SUCCESS: Would removing or degrading this turn materially weaken targeted
evidence collection, accepted rubric coverage, verification, rational
synthesis, or an evidence-derived prediction?

FAILURE: Did this turn materially cause a missed requirement, unsupported
claim, premature stopping, misuse or omission of relevant evidence, faulty
synthesis, or a wrong event, horizon, threshold, or outcome?
Would correcting this turn materially improve the final answer? Every failure
attribution must use failure_kind="answer_error".

REASONING TURN MAP RULES:
- Always perform both audits, regardless of score or prediction correctness.
- Wrong or low-scoring trajectories may retain success credit for sound work.
- Correct or high-scoring trajectories may receive failure blame for remaining
  coverage, synthesis, or prediction defects.
- The same turn may appear in both maps only if it materially helped one part
  of the objective and harmed another.
- Do not put tool evidence utility in these two turn maps; tool utility has its
  own field below. A report-writing turn may receive reasoning credit or blame
  only for the concrete synthesis, omission, unsupported claim, or prediction
  it commits. Protocol syntax remains a separate deterministic diagnostic.
- Never blame sound evidence gathering because later synthesis failed.
- Never blame a unique, reasonable search merely because it returned no match.
  Repeated reads, citation checks, failed commands, and report corrections are
  neutral unless the selected turn itself makes or commits a concrete answer
  error. A verification that finds and fixes a real defect can receive success
  credit.
- Attribute an omission only to a turn that caused premature stopping, ignored
  available evidence, or committed the omission in the final synthesis.
- If the submitted binary outcome is wrong, the failure map must contain at
  least one genuinely causal turn. Find the earliest decisive reasoning turn
  that formed or accepted the bad inference or outcome and the later
  report-synthesis turn that committed it, when these are distinct.
- A forecast-decision turn qualifies only when its actual assistant content
  chooses, accepts, or materially justifies the outcome or direction.
  Planning to assess the forecast, gathering or citing evidence, and discussing
  uncertainty without making the decision do not qualify. Never invent causal
  content that is absent from the selected turn.
- Report writing is not itself a failure. Blame a report turn only for the
  incorrect prediction, faulty synthesis, unsupported claim, or material
  omission it commits. Do not blame a final chat message that merely confirms
  completion.
- Evidence collected but ultimately unused is neutral in the reasoning maps.
  Routine navigation, generic planning, formatting, malformed tool syntax, and
  final confirmation receive zero unless the turn itself commits a concrete
  reasoning error.
- Use sparse attribution. Never distribute credit or blame uniformly. There is
  no target number of nonzero turns.
- Values represent causal strength and must be in (0,1]. Omit zero values.
- A channel may be empty only when genuinely no model-controlled turn passes
  its causal test. Score or correctness alone is not a valid empty reason.

Rubric coverage is not synthesis. The forecast must answer the exact event,
horizon, and threshold; weigh the strongest evidence on both sides; and choose
an outcome consistent with the evidence. Use the objective diagnoses to locate
each identified weakness in the reasoning or report turn that caused it.

TOOL EVIDENCE UTILITY:
- Audit each assistant turn that has tool_calls against all evidence available
  before that turn and the following_tool_outputs it produced.
- Positive utility means the call added new task-relevant evidence, a new
  calculation, or a verification that resolved a real uncertainty.
- Negative utility means the call was demonstrably redundant with evidence
  already available, irrelevant to the task, or avoidably failed and produced
  no usable information. A repeated call can be negative even if its syntax is
  valid.
- A unique reasonable search that returns no match is neutral. Never punish
  sound exploration merely because later synthesis ignored its evidence.
- Omit neutral calls. Omit turns that write or modify answer.md.
- utility is signed in [-1,1] excluding zero. The utility applies to every
  semantic command token in that tool-call turn; fixed wrapper syntax is masked
  separately.

Return JSON only:
{
  "schema_version": 8,
  "turn_count": N,
  "success_attributions": [
    {"turn_index": 0, "success_credit": 0.8, "reason": "specific causal contribution"}
  ],
  "success_empty_reason": null,
  "failure_attributions": [
    {"turn_index": 1, "failure_blame": 0.7, "failure_kind": "answer_error", "reason": "specific causal defect in the reasoning or final answer"}
  ],
  "failure_empty_reason": null,
  "tool_call_attributions": [
    {"turn_index": 2, "utility": -0.4, "evidence_delta": "duplicates evidence already available from turn 0", "reason": "specific utility judgment"}
  ]
}

Copy turn_count exactly. Indices must be unique within each list and between 0
and turn_count-1. The same index may occur once in each list. When a list is
nonempty, its empty_reason must be null. When empty, empty_reason must provide a
specific explanation.
\end{Verbatim}
\end{tcolorbox}

\FloatBarrier
\section{Model Behavior}
\label{app:frontier-behavior}
We examine how research coverage and forecast accuracy relate to report length, trajectory length, model family, and citation penalties. We also compare repeated reports from the same model on the same environment to understand when more complete research accompanies a correct or incorrect forecast. These analyses use the same 3,036 reports and jury assessments as the main evaluation (Section~\ref{sec:frontier}). 

\subsection{Report Length and Research Coverage}
\label{app:frontier-length}
Table~\ref{tab:frontier-behavior} compares report length with the two evaluation scores. Tokens are counted in the final report with the same \texttt{o200k\_base} tokenizer for every model, including citation text. These counts measure written output not the full trajectory length. 

Qwen3.8 Max writes the longest reports on average, at 16,777 tokens, with 53.54\% coverage. GPT-5.6 Sol reaches 59.17\% coverage with 5,713 report tokens. Gemini 3.8 Flash writes 7,887 tokens on average but covers 32.35\% of the requirements. Longer reports do not necessarily cover more of the required analysis. After subtracting each environment's three-trial mean, the within-model correlation between log report length and coverage ranges from $-0.009$ to $0.194$ across models.

\begin{table}[!ht]
\centering
\small
\caption{Report length and scores on \benchpublic{}. Each row contains 276 scored reports. Report-token means are rounded to the nearest token, and coverage and accuracy are percentages.}
\label{tab:frontier-behavior}
\begin{tabular}{lrrr}
\toprule
Model & Mean report tokens & Coverage & Accuracy \\
\midrule
GPT-5.6 Sol & 5,713 & 59.17 & 69.57 \\
Claude Opus 5 & 10,188 & 55.89 & 73.19 \\
Claude Fable 5.1 & 7,713 & 55.85 & 71.38 \\
Grok 4.6 & 9,703 & 58.25 & 64.49 \\
Qwen3.8 Max & 16,777 & 53.54 & 75.36 \\
GLM 5.3 & 3,761 & 47.90 & 67.03 \\
GPT-5.6 Terra & 4,402 & 50.86 & 64.49 \\
Muse Spark 1.3 & 5,521 & 47.00 & 61.23 \\
Kimi K3 & 4,566 & 42.92 & 67.03 \\
Gemini 3.8 Flash & 7,887 & 32.35 & 67.03 \\
DeepSeek V4 Pro 0813 & 2,627 & 37.26 & 61.96 \\
\bottomrule
\end{tabular}
\end{table}

\subsection{Same-Environment Report Comparisons}
\label{app:frontier-cases}
We compare reports on Fiserv and Flywire. For each environment, the same model produces one report that predicts correctly but covers little of the required analysis. Another report covers more than 60\% but predicts incorrectly (Table~\ref{tab:frontier-cases}). Below, we explain how their research and forecast reasoning differ.

\begin{table}[!ht]
\centering
\small
\caption{Contrasting trials on the same research assignment. Trial numbers refer to the three trials reported in the paper. Coverage counts requirements that receive final jury credit.}
\label{tab:frontier-cases}
\begin{tabular}{llrrl}
\toprule
Assignment & Model & Trial & Coverage & Prediction \\
\midrule
Fiserv: Argentina growth & Muse Spark 1.3 & 1 & 3/14 (21.43\%) & Correct (Yes) \\
 & & 3 & 10/14 (71.43\%) & Incorrect (No) \\
Flywire: valuation & Claude Fable 5.1 & 1 & 3/12 (25.00\%) & Correct (No) \\
 & & 2 & 8/12 (66.67\%) & Incorrect (Yes) \\
\bottomrule
\end{tabular}
\end{table}

The Fiserv question asks whether Q1 2026 Argentina revenue will be lower year over year in as-reported dollars, before currency adjustment. Trial 1 distinguishes ``as-reported dollars before currency adjustment, not constant-currency organic'' and predicts Yes. It receives limited research credit. For requirement 1a, the Sol juror notes that the report does not identify Argentina's historical annual growth above 50\% or connect it to the consolidated 9-12\% organic-growth target. Trial 3 documents Argentina growth of 257\% in 2023, 329\% in 2024, and 56\% year to date in 2025, with all three jurors crediting requirement 1a. Its final answer nevertheless predicts that revenue ``will NOT be lower,'' reasoning from a normalized growth comparison. The resolved as-reported revenue falls from \$194 million to \$174 million. The higher-coverage report better documents the historical growth mechanism but misses the subsequent reported-dollar outcome.

The Flywire question asks whether the next quarterly report will disclose revenue dollars or a margin for an individual vertical. Trial 1 predicts No and receives 3/12 final credits. Trial 2 predicts Yes and receives 8/12. Trial 2 earns unanimous credit for relating a low forward EBITDA multiple to double-digit growth. The Gemini juror cites roughly $7\times$ 2026 EV/adjusted EBITDA and guidance for 15-21\% revenue-less-ancillary-services growth. The forecast instead turns on whether acquisition-level Sertifi revenue counts as a vertical disclosure. Trial 1 argues that ``Sertifi is an acquired company, not the travel vertical,'' while Trial 2 treats its revenue as attributable to that vertical. The resolution records no qualifying vertical revenue-dollar or margin disclosure. In Trial 1, a majority of jurors recognize five other requirements but withhold credit because of citation objections. The forecast is correct even though parts of the report are inadequately supported.

These examples show that more complete research does not always lead to a correct forecast. We need to evaluate research coverage and forecast accuracy separately. 

\subsection{Coverage and Correctness Within the Same Environment}
\label{app:frontier-within-task}
We now test whether the relationship illustrated by the two examples in Appendix~\ref{app:frontier-cases} holds across the full evaluation. To separate differences between models from differences between environments, we analyze the three trials of each model separately. For model $m$, environment $e$, and trial $t$, let $C_{met}$ be coverage in percentage points and $Y_{met}\in\{0,1\}$ indicate a correct prediction. We subtract each model--environment pair's trial mean, giving $\widetilde C_{met}=C_{met}-\overline C_{me}$ and $\widetilde Y_{met}=Y_{met}-\overline Y_{me}$. Table~\ref{tab:frontier-within-task} reports their correlation and the within-environment slope,
\[
 b_m=1000\,\frac{\sum_{e,t}\widetilde C_{met}\widetilde Y_{met}}
 {\sum_{e,t}\widetilde C_{met}^{2}},
\]
expressed in accuracy percentage points per ten coverage points. The slope measures whether higher-coverage reports tend to predict more accurately across repeated trials.

For each model, we permute correctness labels among the three trials within each environment, keeping that environment's number of correct predictions fixed. Under the test's null hypothesis, the correctness labels can be reassigned among the three trials of the same environment. We use 100,000 permutations with seed 20260923 and a two-sided test of the within-environment covariance, then apply Holm correction across the eleven models. All 92 environments enter each calculation. The table also reports $N_{\mathrm{mixed}}$, the number of environments with both correct and incorrect predictions across trials. The within-environment correlations range from $-0.123$ to $0.119$. None are statistically significant after Holm correction. These results are consistent with the examples in Appendix~\ref{app:frontier-cases}: more complete research does not always lead to a correct forecast. They do not imply that access to research documents and tools is unhelpful. Ten of eleven models predict more accurately with the full environment than with the question alone (Figure~\ref{fig:question-only}).

\begin{table}[!ht]
\centering
\small
\caption{Coverage-correctness association within each model and environment. $N_{\mathrm{mixed}}$ counts environments with both correct and incorrect trials. $b_m$ is the accuracy-point difference per ten coverage points, and $p$ is the two-sided permutation value.}
\label{tab:frontier-within-task}
\begin{tabular}{lrrrrr}
\toprule
Model & $N_{\mathrm{mixed}}$ & $r$ & $b_m$ & $p$ & Holm $p$ \\
\midrule
GPT-5.6 Sol & 11 & $0.056$ & $1.25$ & 0.449 & 1.000 \\
Claude Opus 5 & 9 & $0.001$ & $0.01$ & 1.000 & 1.000 \\
Claude Fable 5.1 & 8 & $-0.080$ & $-1.15$ & 0.293 & 1.000 \\
Grok 4.6 & 5 & $-0.056$ & $-0.60$ & 0.431 & 1.000 \\
Qwen3.8 Max & 17 & $0.016$ & $0.31$ & 0.846 & 1.000 \\
GLM 5.3 & 14 & $0.037$ & $0.86$ & 0.605 & 1.000 \\
GPT-5.6 Terra & 14 & $-0.123$ & $-2.83$ & 0.077 & 0.846 \\
Muse Spark 1.3 & 16 & $0.017$ & $0.34$ & 0.825 & 1.000 \\
Kimi K3 & 24 & $0.119$ & $2.97$ & 0.102 & 1.000 \\
Gemini 3.8 Flash & 11 & $0.094$ & $1.06$ & 0.240 & 1.000 \\
DeepSeek V4 Pro 0813 & 21 & $0.029$ & $0.86$ & 0.686 & 1.000 \\
\bottomrule
\end{tabular}
\end{table}

\subsection{Open-Weight and Proprietary Model Families}
\label{app:frontier-access}
We group Qwen3.8, GLM~5.3, Kimi~K3, and DeepSeek~V4 Pro as families with publicly released weights, and compare them with the remaining seven proprietary families. All scores come from the evaluated endpoints and settings in Section~\ref{sec:evaluation}. Table~\ref{tab:frontier-access} compares open-weight and proprietary models.

Forecast accuracy is nearly identical between open-weight (67.84\%) and proprietary families (67.34\%). In contrast, research coverage shows a clear gap: open-weight families average 45.41\% compared with 51.34\% for proprietary families, leading to joint success rates of 12.77\% and 20.19\%, respectively. Individual models overlap substantially: Qwen3.8 Max reaches 25.72\% joint success and the highest forecast accuracy in the full comparison.

\begin{table}[!ht]
\centering
\small
\caption{Open-weight versus proprietary models on \benchpublic{}. Scores are percentages averaged equally across models, with 276 reports per model.}
\label{tab:frontier-access}
\begin{tabular}{lrrrrr}
\toprule
Model group & Models & Reports & Coverage & Accuracy & Joint \\
\midrule
Open-weight families & 4 & 1,104 & 45.41 & 67.84 & 12.77 \\
Proprietary families & 7 & 1,932 & 51.34 & 67.34 & 20.19 \\
\bottomrule
\end{tabular}
\end{table}

\subsection{Citation Penalties}
\label{app:frontier-evidence}
We measure how often the jury removes credit from a requirement that the report addresses but does not adequately support. For report $r$, let $H_r$ count majority-hit requirements and $D_r$ count those that also receive a majority citation objection. With $K_r$ rubric requirements, final coverage is $100(H_r-D_r)/K_r$. Table~\ref{tab:frontier-evidence} reports how often a report loses credit ($D_r>0$) and what fraction of addressed requirements lose credit ($\sum_r D_r/\sum_r H_r$). It also gives the mean citation penalty ($100D_r/K_r$). These counts show where the jury found missing or inadequate support, including claims it could not verify.

The final saved assessments remove 3,723 of 22,377 majority-hit requirements across 1,622 reports. Gemini 3.8 Flash has the largest mean citation penalty, at 32.74 coverage points, whereas GPT-5.6 Terra and GLM~5.3 have citation penalties of 2.12 and 1.99 points. This explains part of the difference between discussing the requested topics and earning research credit. The same-environment examples in Appendix~\ref{app:frontier-cases} show how citation objections affect particular reports.

\begin{table}[!ht]
\centering
\small
\caption{Citation penalties that remove research credit. Report rate and removed-move rate are percentages, and the final column is the mean citation penalty in percentage points.}
\label{tab:frontier-evidence}
\begin{tabular}{lrrrrr}
\toprule
Model & Reports & Rate & Moves & Rate & Penalty \\
\midrule
GPT-5.6 Sol & 99/276 & 35.87 & 137/2183 & 6.28 & 4.08 \\
Claude Opus 5 & 195/276 & 70.65 & 396/2319 & 17.08 & 11.55 \\
Claude Fable 5.1 & 169/276 & 61.23 & 328/2252 & 14.56 & 9.56 \\
Grok 4.6 & 172/276 & 62.32 & 334/2332 & 14.32 & 9.57 \\
Qwen3.8 Max & 210/276 & 76.09 & 504/2355 & 21.40 & 14.65 \\
GLM 5.3 & 52/276 & 18.84 & 68/1719 & 3.96 & 1.99 \\
GPT-5.6 Terra & 60/276 & 21.74 & 71/1825 & 3.89 & 2.12 \\
Muse Spark 1.3 & 118/276 & 42.75 & 195/1817 & 10.73 & 5.70 \\
Kimi K3 & 203/276 & 73.55 & 462/1939 & 23.83 & 13.46 \\
Gemini 3.8 Flash & 261/276 & 94.57 & 1121/2243 & 49.98 & 32.74 \\
DeepSeek V4 Pro 0813 & 83/276 & 30.07 & 107/1393 & 7.68 & 3.13 \\
\bottomrule
\end{tabular}
\end{table}

\subsection{Trajectory Length and Report Quality}
\label{app:frontier-trajectories}
We examine whether longer agent runs are associated with higher research coverage or more accurate forecasts. Here, length covers the entire run, not just the final report analyzed in Appendix~\ref{app:frontier-length}. We use all 3,036 trajectories from the 92 \benchpublic{} environments, with three trials per environment for each of the 11 models. Table~\ref{tab:frontier-trajectories} reports each model's average output tokens and assistant turns alongside its research coverage, forecast accuracy, and citation penalties.

We count assistant turns across the complete trajectory, including turns before and after any interruption. Output-token counts use the totals recorded by the evaluation harness from API usage data.  Different from final-report lengths in Table~\ref{tab:frontier-behavior}, which use the same tokenizer for every model.

\begin{table}[!ht]
\centering
\small
\caption{Average trajectory length and scores across all 92 \benchpublic{} environments, with 276 reports per model. Coverage and accuracy are percentages; citation penalties are percentage points.}
\label{tab:frontier-trajectories}
\begin{tabular}{lrrrrr}
\toprule
Model & Output tokens & Turns & Coverage & Accuracy & Penalty \\
\midrule
GPT-5.6 Sol & 63,670 & 55.9 & 59.17 & 69.57 & 4.08 \\
Claude Opus 5 & 112,683 & 73.7 & 55.89 & 73.19 & 11.55 \\
Claude Fable 5.1 & 41,705 & 40.6 & 55.85 & 71.38 & 9.56 \\
Grok 4.6 & 121,434 & 62.8 & 58.25 & 64.49 & 9.57 \\
Qwen3.8 Max & 290,237 & 111.6 & 53.54 & 75.36 & 14.65 \\
GLM 5.3 & 123,392 & 45.0 & 47.90 & 67.03 & 1.99 \\
GPT-5.6 Terra & 59,324 & 57.2 & 50.86 & 64.49 & 2.12 \\
Muse Spark 1.3 & 142,876 & 57.5 & 47.00 & 61.23 & 5.70 \\
Kimi K3 & 22,116 & 28.1 & 42.92 & 67.03 & 13.46 \\
Gemini 3.8 Flash & 90,455 & 75.2 & 32.35 & 67.03 & 32.74 \\
DeepSeek V4 Pro 0813 & 61,713 & 31.1 & 37.26 & 61.96 & 3.13 \\
\bottomrule
\end{tabular}
\end{table}

We then ask whether longer trials of the same model on the same environment tend to perform better. Table~\ref{tab:frontier-trajectory-correlations} compares output-token length with research coverage, forecast correctness, and citation penalties. For each model-environment pair, we subtract the three-trial mean from $\log(1+\text{output tokens})$ and from each outcome. We then calculate the correlations separately for each model. We use two-sided permutation tests with 100,000 draws, shuffling outcomes only among the three trials of the same environment. We apply Holm correction across 33 tests: three outcomes for each of the 11 models.

Coverage correlations range from $-0.254$ to $0.113$, and forecast-correctness correlations range from $-0.190$ to $0.065$. None is statistically significant after Holm correction. In these comparisons, longer runs do not consistently have higher coverage or more accurate forecasts.

\begin{table}[!ht]
\centering
\small
\caption{Trajectory length versus report outcomes across all 92 environments. Correlations use log-transformed output-token counts and compare trials of the same model on the same environment. A positive penalty correlation means that longer runs receive larger citation penalties. None is significant after Holm correction across the 33 comparisons.}
\label{tab:frontier-trajectory-correlations}
\begin{tabular}{lrrr}
\toprule
Model & Coverage & Correctness & Penalty \\
\midrule
GPT-5.6 Sol & $-0.254$ & $-0.017$ & $0.114$ \\
Claude Opus 5 & $0.018$ & $0.023$ & $0.121$ \\
Claude Fable 5.1 & $0.113$ & $0.061$ & $0.002$ \\
Grok 4.6 & $-0.080$ & $-0.029$ & $0.091$ \\
Qwen3.8 Max & $0.021$ & $-0.190$ & $0.047$ \\
GLM 5.3 & $0.040$ & $-0.077$ & $-0.100$ \\
GPT-5.6 Terra & $0.020$ & $0.065$ & $0.002$ \\
Muse Spark 1.3 & $0.086$ & $-0.110$ & $-0.050$ \\
Kimi K3 & $0.085$ & $0.056$ & $-0.089$ \\
Gemini 3.8 Flash & $0.040$ & $-0.092$ & $-0.069$ \\
DeepSeek V4 Pro 0813 & $0.027$ & $-0.044$ & $-0.036$ \\
\bottomrule
\end{tabular}
\end{table}

\FloatBarrier

\section{Execution Errors and Model-Level Case Studies}
\label{app:frontier-failures}

Agents can spend many turns retrieving documents and checking citations while still leaving the required analysis incomplete. To understand where they struggle, we inspect all 3,036 \benchpublic{} trajectories across 92 environments, three trials, and 11 models. We count command rejections, automatic repairs, repeated document reads, and long runs with low coverage. We then examine examples from each model, distinguishing errors corrected during the run from problems that remain in the final report. These selected cases illustrate specific model behaviors.

\subsection{Recurring execution patterns}
\label{app:frontier-failure-counts}

We examine how often agents encounter command-format problems, repeat document reads, or produce long trajectories with low coverage. Counting these behaviors across all 3,036 runs helps distinguish recurring patterns from isolated examples. Table~\ref{tab:frontier-operational} gives the counts for each model. A run can appear in more than one column.

\begin{table}[!ht]
\centering
\small
\setlength{\tabcolsep}{5pt}
\caption{Execution patterns across all 92 environments. Each entry gives the number and percentage of a model's 276 runs. Repeated reads means at least five extra identical read commands. Long/low coverage means at least 100 assistant turns and less than 20\% research coverage.}
\label{tab:frontier-operational}
\begin{tabular}{lrrrr}
\toprule
Model & Rejection & Auto-repair & Repeated reads & Long/low coverage \\
\midrule
Claude Fable 5.1 & 0 (0.0\%) & 4 (1.4\%) & 0 (0.0\%) & 0 (0.0\%) \\
Claude Opus 5 & 0 (0.0\%) & 24 (8.7\%) & 0 (0.0\%) & 0 (0.0\%) \\
DeepSeek V4 Pro & 0 (0.0\%) & 8 (2.9\%) & 72 (26.1\%) & 0 (0.0\%) \\
Gemini 3.8 Flash & 61 (22.1\%) & 215 (77.9\%) & 0 (0.0\%) & 9 (3.3\%) \\
GLM 5.3 & 1 (0.4\%) & 41 (14.9\%) & 214 (77.5\%) & 0 (0.0\%) \\
GPT-5.6 Sol & 0 (0.0\%) & 0 (0.0\%) & 71 (25.7\%) & 0 (0.0\%) \\
GPT-5.6 Terra & 0 (0.0\%) & 8 (2.9\%) & 44 (15.9\%) & 0 (0.0\%) \\
Grok 4.6 & 0 (0.0\%) & 25 (9.1\%) & 0 (0.0\%) & 0 (0.0\%) \\
Kimi K3 & 0 (0.0\%) & 14 (5.1\%) & 0 (0.0\%) & 0 (0.0\%) \\
Muse Spark 1.3 & 0 (0.0\%) & 1 (0.4\%) & 154 (55.8\%) & 0 (0.0\%) \\
Qwen3.8 Max & 0 (0.0\%) & 2 (0.7\%) & 0 (0.0\%) & 8 (2.9\%) \\
\bottomrule
\end{tabular}
\end{table}

First, we examine response-format errors. A rejection means the harness could not parse the agent's response. An automatic repair means it corrected the formatting so the response could still be executed. Gemini receives 1,961 rejections across 61 environments, accounting for 9.45\% of its 20,753 assistant turns; 59 runs contain at least five rejections. Across all three trials, 10,040 turns receive repair warnings. 

Beyond formatting, we examine repeated document reads. At least five extra identical reads occur in 214 GLM runs and 154 Muse Spark runs. We count a read as repeated when its command exactly matches an earlier document-read command. This includes shell commands such as \texttt{cat}, \texttt{sed}, and \texttt{grep}, but excludes writes and checks of \texttt{answer.md}. It also misses equivalent commands written differently and reads performed in Python. These repetitions do not necessarily indicate wasted work: in the GLM and DeepSeek examples in Table~\ref{tab:frontier-operational-cases}, agents reread passages because earlier terminal output was cut off.

Finally, we examine whether long runs cover the required analysis. Eight Qwen runs and nine Gemini runs use at least 100 turns but achieve less than 20\% coverage, spanning eight environments for each model. More broadly, 119 Qwen runs use at least 100 turns but cover no more than 60\% of the required analysis. The examples below examine what agents do during these runs and what analysis remains missing. Appendix~\ref{app:frontier-behavior} examines how trajectory length relates to coverage and forecast accuracy.

\subsection{Examples across all eleven models}
\label{app:frontier-model-cases}

We examine one run from each model to understand the behaviors counted in Appendix~\ref{app:frontier-failure-counts}. For each example, we describe what happened, how the agent responded, and the final report's research coverage and forecast correctness (Table~\ref{tab:frontier-operational-cases}). The cases include errors corrected on the next turn, repeated attempts to read source passages, and long runs that still miss required analysis. Turn numbers count assistant turns, starting from 1.

\begingroup
\small
\setlength{\tabcolsep}{4pt}
\renewcommand{\arraystretch}{1.15}
\setlength{\LTpre}{8pt}
\setlength{\LTpost}{6pt}
\begin{longtable}{@{}p{0.23\linewidth}p{0.73\linewidth}@{}}
\caption{One example run per model on \benchpublic{}. Each row gives the company, research topic, and trial, followed by the agent's actions and final results.}\label{tab:frontier-operational-cases}\\
\toprule
Model and environment & Observed behavior and final outcome \\
\midrule
\endfirsthead
\multicolumn{2}{l}{\textit{Table~\ref{tab:frontier-operational-cases}, continued.}}\\
\toprule
Model and environment & Observed behavior and final outcome \\
\midrule
\endhead
\bottomrule
\endfoot
Claude Fable 5.1\newline FND valuation (trial 1) &
The agent incorrectly treats a price-history dictionary as a list on turn 3, then checks its keys on turn 4. It corrects a separate string-formatting error from turn 7 on turn 8. The run finishes in 51 turns, with 40.00\% coverage and an incorrect forecast. \\
\addlinespace[3pt]
Claude Opus 5\newline LULU supply chain (trial 2) &
A Python command fails on turn 32 because line breaks are written as text rather than actual line breaks. The agent moves the code into a script file and prints the peer financial data on turn 33. Although the command now works, the final report earns credit for only 3/12 analytical moves after 69 turns and 183,429 output tokens. The forecast is incorrect. \\
\addlinespace[3pt]
DeepSeek V4 Pro\newline SNAP hardware (trial 1) &
The agent requests the same 10-Q passage on five consecutive turns (19--23). It notes that earlier terminal output was cut off, but continues requesting batches of passages and financial records. These repeated reads aim to recover missing source text. The run finishes in 32 turns with 25.00\% coverage and a correct forecast. \\
\addlinespace[3pt]
Gemini 3.8 Flash\newline IRDM constellation (trial 1) &
The harness rejects the response format on 36 of 134 turns. The agent later reports that its own citation-format check passes all 149 citations. The jury recognizes eight analytical moves but removes credit for six because of citation penalties, leaving 2/12 credited moves (16.67\%). The forecast is incorrect. Appendix~\ref{app:frontier-verification-cases} explains why passing a citation-format check does not establish that the sources support the analysis. \\
\addlinespace[3pt]
GLM 5.3\newline SOC regulation (trial 1) &
The agent requests one 10-K passage 11 times, including every turn from 29 through 34. It says it has identified the main argument but still needs source text that has not been cut off. It continues combining multiple reads in each request while trying to obtain those passages. The run finishes after 61 turns with 33.33\% coverage and a correct forecast. \\
\addlinespace[3pt]
GPT-5.6 Sol\newline TKO revenue synergies (trial 2) &
The report-writing script searches for exact phrases in the source text. A search fails on turn 26; by turn 28, the agent finds that the phrase spans two lines. More searches fail, and rerunning the unchanged script on turn 30 repeats the error. The agent then searches for shorter phrases and allows matches across lines. The run finishes in 53 turns with 50.00\% coverage and an incorrect forecast. \\
\addlinespace[3pt]
GPT-5.6 Terra\newline LULU niche sports (trial 3) &
After a broad search produces too much output, the agent sends the text \texttt{C-c} followed by a newline instead of a keyboard interrupt. The shell treats it as a command and returns ``command not found'' on turn 5. The agent then continues with narrower searches. The run finishes in 40 turns with 33.33\% coverage and a correct forecast. \\
\addlinespace[3pt]
Grok 4.6\newline APO private credit (trial 2) &
A multiline data-extraction command fails on turn 15. The agent switches to a script file, but a command used to shorten its output produces a broken-pipe message. It then saves the results to a file and reads selected passages. The run finishes in 77 turns with 50.00\% coverage and a correct forecast. \\
\addlinespace[3pt]
Kimi K3\newline FND competitive moat (trial 2) &
The agent finds incorrect peer figures and a wrong fiscal date while checking its draft on turns 30--31. It corrects both and later removes a trading-volume claim that refers to the wrong period. Despite these corrections, the final report earns credit for only 1/12 analytical moves (8.33\%): the jury recognizes four moves but removes credit for three because of citation penalties. The forecast is correct. \\
\addlinespace[3pt]
Muse Spark 1.3\newline LULU geographic growth (trial 2) &
The agent reads one 10-Q passage 26 times and another 15 times. Across turns 17--41, it repeatedly searches for exact source locations or additional evidence. The run takes 90 turns and 317,488 output tokens. The jury recognizes six analytical moves but removes credit for four because of citation penalties, leaving 16.67\% coverage. The forecast is incorrect. \\
\addlinespace[3pt]
Qwen3.8 Max\newline LULU supply chain (trial 1) &
The agent repeatedly checks sources before writing its report at turn 90. The full run uses 113 turns and 323,704 output tokens. The report contains 251 citations and headings for all four research themes, but earns credit for only 1/12 analytical moves (8.33\%). The forecast is incorrect. Many citations and the expected headings do not mean the required analysis is complete. \\
\end{longtable}
\endgroup

A further Gemini example shows how self-checking can lead to additional execution errors. In the IRDM-IoT run (trial 3), a quoting error leaves the shell waiting for the end of a string. The agent sends the literal characters \texttt{\textbackslash x03} instead of a keyboard interrupt. Bash treats them as a command. Later, three consecutive attempts to run or fix the citation checker fail with syntax errors (turns 75--77). The run still finishes in 101 turns with 40.00\% coverage and a correct forecast. We found this exact failed-interrupt pattern in one report; it is not one of the recurring patterns counted in Table~\ref{tab:frontier-operational}.

\subsection{Limits of Agent Self-Checks}
\label{app:frontier-verification-cases}

We examine why reports can still receive low research coverage after agents check their own work. A citation-format check confirms that a cited file exists and the line numbers are valid. It does not establish that the passage supports the claim or that the report contains the required analysis.

Gemini's IRDM constellation report shows the difference between citation format and source support. The agent reports that all 149 citations pass its own format check, but the jury credits only two analytical moves after citation penalties. A citation can point to a valid passage without supporting the analysis that cites it. Appendix~\ref{app:frontier-evidence} reports citation penalties across the full evaluation.

Qwen's TKO strategic-M\&A report (trial 3) shows that repeated checking does not necessarily resolve this problem. The agent repeatedly declares evidence collection complete, then returns to checking sources and citation locations. It first writes the report on turn 120 and continues checking during a 197-turn run, using 302,360 output tokens. Its citation checker passes all 253 citations, but the jury removes credit for three of eight recognized analytical moves, leaving 41.67\% coverage. The forecast is correct. This run has no response-format rejections, and the agent is not simply repeating the same read. Its repeated checks still leave gaps in the analysis and its support.

Missing analysis is a separate problem. In Qwen's LULU supply-chain report (trial 1), only one required analytical move is recognized, and it loses no credit to citation penalties. The report includes the requested headings and many citations but omits most of the required comparisons and reasoning. Kimi's FND report shows that self-checks can make useful corrections without solving this broader problem: the agent corrects numbers in its draft, yet its final report still misses much of the required analysis. 

\end{document}